%% file: main.tex
\documentclass[numbib]{imaiai}

\usepackage{amsmath}
\usepackage{amsthm}

\input{preamble}
\usepackage{graphicx}
\usepackage{soul}
\usepackage{rotating}
\usepackage{empheq}
\begin{document}

\title{Mad Max: Affine Spline Insights into Deep Learning
}

\shorttitle{Mad Max: Affine Spline Insights into Deep Learning} 
\shortauthorlist{Balestriero \& Baraniuk} 

\author{{
\sc Randall Balestriero}, 
{\sc Richard Baraniuk}\\[2pt]
Rice University, Houston, Texas, USA\\[2pt]
{\email{randallbalestriero@gmail.com}}, {\email {richb@rice.edu}} 
}
\maketitle

\begin{abstract}
{
We build a rigorous bridge between deep networks (DNs) and approximation theory via spline functions and operators.
Our key result is that a large class of DNs can be written as a composition of max-affine spline operators (MASOs), which provide a powerful portal through which to view and analyze their inner workings.
For instance, conditioned on the input signal, the output of a MASO DN can be written as a simple affine transformation of the input.
This implies that a DN constructs a set of signal-dependent, class-specific templates against which the signal is compared via a simple inner product; we explore the links to the classical theory of optimal classification via matched filters and the effects of data memorization.
Going further, we propose a simple penalty term that can be added to the cost function of any DN learning algorithm to force the templates to be orthogonal with each other; this leads to significantly improved classification performance and reduced overfitting with no change to the DN architecture. 
The spline partition of the input signal space that is implicitly induced by a MASO directly links DNs to the theory of vector quantization (VQ) and $K$-means clustering, which opens up new geometric avenue to study how DNs organize signals in a hierarchical fashion.
To validate the utility of the VQ interpretation, we develop and validate a new distance metric for signals and images that quantifies the difference between their VQ encodings. (This paper is a significantly expanded version of \textit{A Spline Theory of Deep Learning} from ICML 2018.)
}
%
{deep learning, neural networks, splines, approximation, vector quantization}
\end{abstract}

\input{intro}

\input{background}
\input{splines}

\input{deepMaso}
\input{template}

\input{orthoTemplate}

\input{partition}
\input{conclusions}

\input{acks}

\clearpage
\appendix

\begin{center}
{\Large\bf
~~ \\[5mm]
SUPPLEMENTARY MATERIAL
\\[5mm]} 
\end{center}

\input{APPENDIX/notation}
\input{APPENDIX/background}
\input{APPENDIX/affine}
\input{APPENDIX/proofs}

\input{APPENDIX/topology}

\clearpage
\bibliographystyle{plainnat}
\bibliography{ref}
\end{document}

%% file: preamble.tex
\usepackage{url}
\usepackage{comment}
\usepackage{float}
\usepackage{comment}

\usepackage[version=4]{mhchem}
\DeclareMathAlphabet\mathbfcal{OMS}{cmsy}{b}{n}
\DeclareMathOperator*{\argmax}{arg\,max}
\DeclareMathOperator*{\argmin}{arg\,min}

\def \bmu{\boldsymbol{\mu}}

\def \a{\alpha}
\def \b{\beta}
\def \be{\beta}
\def \bb{b}
\def \bx{\boldsymbol{x}}
\def \bW{W}

\def \by{\boldsymbol{y}}
\def \bz{\boldsymbol{z}}

\def \bu{\boldsymbol{u}}
\def \bv{\boldsymbol{v}}
\def \bb{b}

\def \be{\boldsymbol{e}}

\def \bA{A}

\def \bB{B}

\def \bC{\boldsymbol{C}}

\def\R{{\mathbb R}}

\def\Indic{\textbf{1}}

\newcommand{\D}[0] {{ \mathcal{D} }}

\newcommand{\qq}{\vspace*{-2mm}}
\newcommand{\qqq}{\vspace*{-3mm}}

\newcommand{\bigcdot}{\boldsymbol{\cdot}}

\newcommand {\richb}[1]{{\color{blue}\sf{[richb: #1]}}}
\newcommand  {\randall}[1]{{\color{red}\sf{[rb: #1]}}}

\usepackage{xcolor}

%% file: intro.tex
\section{Introduction}
\label{sec:intro}

Deep learning has significantly advanced our ability to address a wide range of difficult machine learning and signal processing problems.
Today's machine learning landscape is dominated by {\em deep (neural) networks} (DNs), which are compositions of a large number of simple parameterized linear and nonlinear operators.
An all-too-familiar story of late is that of plugging a deep network into an application as a black box, learning its parameter values using copious training data, and then significantly improving performance over classical task-specific approaches.

Despite this empirical progress, the precise mechanisms by which deep learning works so well remain relatively poorly understood, adding an air of mystery to the entire field.
Ongoing attempts to build a rigorous mathematical framework fall roughly into five camps:
(i) probing and measuring networks to visualize their inner workings 
\cite{zeiler2014visualizing};
(ii) analyzing their properties such as expressive power \cite{pmlr-v49-cohen16}, loss surface geometry \cite{lu2017depth, soudry2017exponentially}, nuisance management \cite{soatto2016visual}, 
sparsification \cite{papyan2017convolutional}, 
and generalization abilities;
(iii) new mathematical frameworks that share some (but not all) common features with DNs \cite{bruna2013invariant};
(iv) probabilistic generative models from which specific DNs can be derived \cite{arora2013provable,patel2016probabilistic}; and (v) 
information theoretic bounds \cite{tishby2015deep}.

In this paper, we build a rigorous bridge between DNs and {\em approximation theory} via {\em spline functions and operators}.
We prove that a large class of DNs --- including convolutional neural networks (CNNs) \cite{lecun1998mnist}, residual networks (ResNets) \cite{resnet-he}, skip connection networks \cite{srivastava2015training}, fully connected networks \cite{pal1992multilayer}, recurrent neural networks (RNNs) \cite{graves2013generating}, scattering networks \cite{bruna2013invariant}, inception networks \cite{szegedy2017inception}, and more --- can be written as spline operators.

Moreover, when these DNs employ current standard-practice piecewise affine and convex nonlinearities (e.g., ReLU, absolute value, max-pooling, etc.) they can be written as the composition of {\em max-affine spline operators} (MASOs), which are a new extension of max-affine splines \cite{magnani2009convex,hannah2013multivariate}.
An extremely useful feature of such a spline is that it circumvents the major  complication of spline function approximation -- the need to jointly optimize not only the spline function parameters but also the partition of the domain over which those parameters are constant (the ``knots'' of the spline).  
Instead, the partition of a max-affine spline is determined implicitly in terms of its slope and offset parameters.
%
We focus on such nonlinearities here but note that our framework applies also to non-piecewise affine nonlinearities through a standard approximation argument.

The max-affine spline connection provides a powerful portal through which to view and analyze the inner workings of a DN using tools from approximation theory and functional analysis. 
While there has been prior work relating DNs to splines  \cite{montufar2014number,rister2017piecewise,unser2018representer}, to date  theoretical studies have been focused on either specific topologies or providing theoretical guarantees and bounds on specific properties, such as a DN's approximation capacity or stability.
In contrast, we develop a theory that is agnostic to the DN architecture and supports not just analysis but also extensions.

Here is a summary of our primary contributions:
\begin{description}
\setlength{\itemsep}{0mm}

\item[{\bf[C1]}] 
From our proof that a large class of DNs can be written as a composition of MASOs, it follows immediately that, {\em conditioned on the input signal, the output of a DN is a simple affine transformation of the input.}
We illustrate in Section~\ref{sec:deepMaso} by deriving closed form expressions for the input/output mapping of CNNs and ResNets.

\item[{\bf[C2]}]
We prove that a composition of two or more MASOs is capable of {\em approximating an arbitrary operator} in Section~\ref{sec:universality}.

\item[{\bf[C3]}] 
The affine mapping formula enables us to interpret a MASO DN as constructing a set of {\em signal-dependent, class-specific templates} against which the signal is compared via a simple inner product. 
In Section \ref{sec:template} we relate DNs directly to the classical theory of optimal classification via matched filters and provide insights into the effects of {\em data memorization} \cite{zhang2016understanding}.

\item[{\bf[C4]}] 
We propose a simple penalty term that can be added to the cost function of any DN learning algorithm to force the templates to be {\em orthogonal} to each other. In Section \ref{sec:orthoTemplate}, we show that this leads to significantly improved classification performance and reduced overfitting on standard test data sets like SVHN, CIFAR10, and CIFAR100 with no change to the DN architecture. 

\item[{\bf[C5]}] 
The {\em partition of the input space} induced by a MASO links DNs to the theory of {\em vector quantization} (VQ) and {\em $K$-means clustering}, which opens up a new geometric avenue to study how DNs cluster and organize signals in a hierarchical fashion.
Section~\ref{sec:partition} studies the properties of the MASO partition.   

\item[{\bf[C6]}] 
Leveraging the fact that a DN considers two signals to be similar if they lie in the same MASO partition region, we develop a new {\em VQ-based distance} for signals and images in Section~\ref{sec:clustering} that measures the difference between their VQ encodings. 
The distance is easily computed via backpropagation on the DN.

\end{description}

The main text of this paper contains the key results, theorem statements, and experimental results. 
A number of appendices in the Supplementary Material (SM) contain the rigorous mathematical set up, proofs, additional insights, and additional examples.
A condensed version of this paper containing a subset of the results was presented at ICML 2018 \cite{report18}.

%% file: background.tex
\section{Background on Deep Networks}
\label{sec:back}

A {\em deep network} (DN) is an operator $f_\Theta: \R^D \rightarrow \R^C$ that maps an input signal
$\bx\in\R^D$ to an output prediction $\by \in \R^C$.
All current DNs can be written as a  composition of $L$ intermediate mappings called {\em layers}
\begin{equation}
f_\Theta(\bx)= \left(f^{(L)}_{\theta^{(L)}} \circ \dots \circ f^{(1)}_{\theta^{(1)}}\right)\!(\bx),
\label{eq:layers1}
\end{equation}
where $\Theta=\left\{\theta^{(1)},\dots,\theta^{(L)}\right\}$ is the 
collection of the network's parameters from each layer.
This composition of mappings is nonlinear and non-commutative, in general.

The DN {\em layer} at level $\ell$ is an operator $f^{(\ell)}_{\theta^{(\ell)}}$ that takes as input the vector-valued signal $\bz^{(\ell-1)}(\bx)\in \R^{D^{(\ell-1)}}$ and produces the vector-valued output $\bz^{(\ell)}(\bx)\in \R^{D^{(\ell)}}$.
The signals $\bz^{(\ell)}(\bx),\ell \geq 1$ are typically called {\em feature maps}; it is easy to see that 
\begin{equation}
\bz^{(\ell)}(\bx)=\left(f^{(\ell)}_{\theta^{(\ell)}} \circ \dots \circ f^{(1)}_{\theta^{(1)}}\right)\!(\bx),~~\ell \in \{1,\dots,L\}
\label{eq:layers2}
\end{equation}
with the initialization $\bz^{(0)}(\bx):=\bx$ and $D^{(0)}:=D$.
Note that $D^{(L)}=C$. 

For concreteness, we will focus here on processing multichannel images $\bx$, such as RGB color digital photographs.
However, our analysis and techniques apply to signals of any index-dimensionality, including speech and audio signals, video signals, etc., simply by adjusting the appropriate dimensionalities.
We will use two equivalent representations for the signal and feature maps, one based on tensors and one based on flattened vectors.
In the {\em tensor} representation, the input image $x$ contains $C^{(0)}$ channels of size $\left(I^{(0)} \times J^{(0)}\right)$ pixels, and the feature map $z^{(\ell)}$ contains $C^{(\ell)}$ channels of size $\left(I^{(\ell)} \times J^{(\ell)}\right)$ pixels.
In the {\em vector} representation, 
$[\bx]_k$ represents the entry of the $k^{\rm th}$ dimension of the flattened, vector version $\bx$ of $x$.
Hence, $D^{(\ell)}=C^{(\ell)}I^{(\ell)}J^{(\ell)}$, 
$C^{(L)}=C$, $I^{(L)}=1$, and $J^{(L)}=1$. Appendix \ref{sec:notation} describes our notation in detail.

\subsection{DN Operators and Layers}
\label{sec:dnop}

We briefly overview the basic DN operators and layers we consider in this paper. 
A \textbf{{\em fully connected operator}} performs an arbitrary affine transformation by multiplying its input by the dense matrix $\bW^{(\ell)} \in \mathbb{R}^{D^{(\ell)} \times D^{(\ell-1)}}$ and adding the arbitrary bias vector $\bb_\bW^{(\ell)} \in \mathbb{R}^{D^{(\ell)}}$, as in $f^{(\ell)}_\bW \!\left(\bz^{(\ell-1)}(\bx)\right):=   \bW^{(\ell)}\bz^{(\ell-1)}(\bx)+\bb_\bW^{(\ell)}$.

A \textbf{{\em convolution operator}} performs a constrained affine transformation by multiplying its input by the multichannel block-circulant convolution matrix $\bC^{(\ell)}\in \mathbb{R}^{D^{(\ell)} \times D^{(\ell-1)}}$ and adding the bias vector $\bb_{\bC}^{(\ell)} \in \mathbb{R}^{D^{(\ell)}}$, as in $f^{(\ell)}_{\bC} \!\left(\bz^{(\ell-1)}(\bx)\right):= \bC^{(\ell)}\bz^{(\ell-1)}(\bx)+\bb_{\bC}^{(\ell)}$.
The multichannel convolution matrix applies $C^{(\ell)}$ different filters of dimensions $\left(C^{(\ell-1)},I^{(\ell)}_{\bC},J^{(\ell)}_{\bC}\right)$ to the input; that is, each filter is a $C^{(\ell-1)}$-channel filter of spatial size $\left(I^{(\ell)}_{\bC},J^{(\ell)}_{\bC}\right)$.
Typically, the bias $\bb_{\bC}^{(\ell)}$ is constrained to be a (potentially different) constant at all spatial positions of each of the $C^{(\ell)}$ output channels.
The convolution and bias constraints radically reduce the number of parameters compared to a fully connected operator for similar input-output dimensions, but impose a structural bias of translation invariance \cite{lecun1998mnist}.  
Additional details are provided in Appendix \ref{sm:convop}.

An \textbf{{\em activation operator}} applies a pointwise nonlinearity $\sigma$ to its input, as in $
    \left[f^{(\ell)}_\sigma \!\left(\bz^{(\ell-1)}(\bx)\right)\right]_k=\sigma\! \left([\bz^{(\ell-1)}(\bx)]_k\right)$, $k=1,\dots,D^{(\ell)}$.
Nonlinearities are crucial to DNs, since otherwise the entire network would collapse to a single global affine transform. 
Generally, a different nonlinearity can be used for each dimension of $\bz^{(\ell-1)}(\bx)$, but standard practice employs the same $\sigma$ across all entries.
Three popular activation functions are the {\em rectified linear unit} (ReLU)
$\sigma_{\rm ReLU}(u):=\max(u,0)$, the {\em leaky ReLU}
$\sigma_{\rm LReLU}(u):= \max(\eta u,u),\;\eta>0$, and the {\em absolute value} $\sigma_{\rm abs}(u):=|u|$.
These three functions are both piecewise affine and convex.
Other popular activation functions include the {\em sigmoid} $\sigma_{\rm sig}(u):=\frac{1}{1+e^{-u}}$ and
{\em hyperbolic tangent} $\sigma_{\rm tanh}(u):=2\sigma_{\rm sig}(2u)-1$.
These two functions are neither piecewise affine nor convex.

A \textbf{{\em pooling operator}} subsamples its input to reduce its dimensionality according to a policy $\rho$ applied over $D^{(\ell)}$ pooling regions, each one defined by the collection of indices
$\left\{\mathcal{R}^{(\ell)}_k\right\}_{k=1}^{D^{(\ell)}}$ on which $\rho$ will be applied independently. 
Pooling regions typically correspond to small, non-overlapping image blocks or channel slices. 
A widely used policy $\rho$ is {\em max-pooling}: $
    \left[f^{(\ell)}_\rho \! \left(\bz^{(\ell-1)}(\bx)\right)\right]_k
    =
    \max_{d\in \mathcal{R}^{(\ell)}_k} \left[\bz^{(\ell-1)}(\bx)\right]_d ,k=1,\dots,D^{(\ell)}$. 
See Appendix \ref{sm:poolingop} for the precise definition of max-pooling plus the definitions of {\em average pooling} and {\em channel pooling}, which is used in maxout networks \cite{goodfellow2013maxout}.

More details on the above operators plus additional ones corresponding to the {\em skip connections} used in skip connection networks \cite{srivastava2015training} and residual networks (ResNets) \cite{resnet-he,targ2016resnet} and the {\em recurrent connections} used in RNNs \cite{graves2013generating} are contained in Appendix~\ref{app:operator}.

There is no consensus definition in the literature of a DN layer. 
For the purposes of this paper, we will use the following.

\begin{definition}
A DN {\em layer} $f^{(\ell)}$ comprises a single nonlinear DN operator composed with any 
preceding affine operators lying between it and the preceding nonlinear operator.
\label{def:layer}
\end{definition}

This definition yields a single, unique layer decomposition of any DN, and the complete DN is then the composition of its layers per (\ref{eq:layers1}).
For example, in a standard CNN, there are two different layers types: i)~convolution-activation and ii)~max-pooling.

\subsection{Output Operators}

The final operators in a DN typically consist of one or more fully connected operators interspersed with activation operators.  
We form the prediction by feeding the DN output $f_\Theta(\bx)$ through a final nonlinearity $g:\mathbb{R}^{C}\rightarrow \mathbb{R}^{C}$ to form $y=g(f_\Theta(\bx))$.
For classification problems, $g$ is typically the {\em softmax}, which arises naturally from posing the classification task as a multinomial logistic regression  \cite{bishop1995neural}, but other options are available, such as the {\em spherical softmax} \cite{de2015exploration}.
For regression problems, typically no final nonlinearity $g$ is applied.

\subsection{DN Learning}

We {\em learn} the DN parameters $\Theta$ for a particular prediction task in a supervised setting using a labeled data set $\mathcal{D}=(\bx_n,y_n)_{n=1}^N$,
a loss function $\mathcal{L}:\mathbb{R}^{C} \times \mathbb{R}^{C} \rightarrow \mathbb{R}^+$, 
and a learning policy to update the parameters $\Theta$ in the predictor $f_\Theta(\bx)$.
For a $C$-class classification problem posed as a multinomial logistic regression \cite{bishop1995neural}, $y_n \in \{1,\dots,C\}$, and the loss function is typically a regularized version of the negative cross-entropy 
\begin{align}
    \mathcal{L}_{\rm CE}&(y_n,g(f_\Theta(\bx_n))) = -\log \left( [g(f_\Theta(\bx_n))]_{y_n}\right)= -[f_\Theta(\bx_n)]_{y_n}+\log\left(\sum_{c=1}^C e^{[f_\Theta(\bx_n)]_{c}}\right).
    \label{eq:crossEnt}
\end{align}

For regression problems, $y_n$ becomes the vector $\by_n \in \mathbb{R}^C$, and the loss function is typically the mean squared error.
The total loss is obtained by summing over all of the $N$ samples in $\mathcal{D}$ or mini-batches sampled from it.

Since the layer-by-layer operations in a DN are differentiable almost everywhere with respect to their parameters and inputs, we can optimize the parameters $\Theta$ with respect to the total loss using some form of 
first-order optimization such as gradient descent (e.g., Adam \cite{kingma2014adam}).
Moreover, the gradients of all internal parameters can be computed efficiently using {\em backpropagation} \cite{hecht1988theory}, which follows from the chain rule of calculus. 

%% file: splines.tex
\section{Background on Spline Operators}
\label{sec:splines}

{\em Approximation theory} is the study of how and how well functions can best be approximated using simpler functions \cite{powell1981approximation}.
A classical example of a simpler function is a {\em spline} $s: \R^D\rightarrow \R$ \cite{de1978practical}.
{\em For concreteness, we will focus exclusively on affine splines in this paper (a.k.a. ``linear splines''), but our ideas generalize naturally to higher-order splines.}

\subsection{Multivariate Affine Splines}

Consider a {\em partition} of a domain $\R^D$ into a set of regions $\Omega=\{\omega_1,\dots,\omega_R\}$ 
and a set of local mappings $\Phi=\{\phi_{1},\dots,\phi_{R}\}$ that map each region in the partition to $\R$ via 
$\phi_{r}(\bx) := \langle [\a]_{r,\bigcdot},\bx\rangle +[\b]_{r}$ for $\bx\in\omega_r$.\footnote
{
To make the connection between splines and DNs more immediately obvious, here $\bx$ is interpreted as a point in $\R^D$, which plays the r\^{o}le of the space of signals in the other sections.
}
The parameters are:
$\a \in\R^{R\times D}$, a matrix of hyperplane ``slopes,'' and $\b \in \R^{R}$, a vector of hyperplane ``offsets'' or ``biases''.
We will use the terms offset and bias interchangeably in the sequel.
The notation $[\a]_{r,\bigcdot}$ denotes the column vector formed from the $r^{\rm th}$ row of $\a$.

Then the {\em multivariate affine spline} is defined as 
\begin{align}
    s[\a,\b,\Omega](\bx)= \sum_{r=1}^R 
    \left(\langle [\a]_{r,\bigcdot},\bx\rangle 
    + [\b]_{r} \right)\Indic(\bx \in \omega_r) =: \langle \a[\bx], \bx \rangle + \b[\bx],
\label{eq:lmas}
\end{align}
where $\Indic(\bx \in \omega_r)$ denotes the indicator function that equals $1$ when $\bx \in \omega_r$ and $0$ otherwise.
This equation also introduces the streamlined notation $\a[\bx]=[\a]_{r,\bigcdot}$ when $\bx \in \omega_r$;
the definition for $\b[\bx]$ is similar. 
Such a spline is piecewise affine and hence piecewise convex.
However, in general, it is neither globally affine nor globally convex unless $R=1$, a case we denote as a \textit{degenerate spline}, since it corresponds simply to an affine mapping.

\subsection{Max-Affine Spline Functions}

A major complication of function approximation with splines in general is the need to jointly optimize both the spline parameters $\a, \b$ and the input domain partition $\Omega$ (the ``knots'' for a $1$D spline) \cite{bennett1985structural}.

However, if a multivariate affine spline 
is constrained to be {\em globally convex}, then it can be rewritten as a {\em max-affine spline} \cite{magnani2009convex,hannah2013multivariate}
\begin{align}
    s[\a,\b,\Omega](\bx)=\max_{r=1,\dots,R} \langle [\a]_{r,\bigcdot} , \bx\rangle +[\b]_r\;. 
    \label{eq:maffine}
\end{align}
An extremely useful feature of such a spline is that {\em it is completely determined by its parameters $\a$ and $\b$ without needing to specify the partition $\Omega$}.
As such, we denote a max-affine spline simply as $s[\a,\b]$.
Changes in the parameters $\a,\b$ of a max-affine spline automatically induce changes in the partition $\Omega$, meaning that they are {\em adaptive partitioning splines} \cite{magnani2009convex}. 

A max-affine spline is always a piecewise affine and globally convex (and hence also continuous) function.
Conversely, any piecewise affine and globally convex function can be written as a max-affine spline.

\begin{proposition}
\label{thm:masoequiv}
For any (continuous) $h\in \mathcal{C}^{0}(\mathbb{R}^D)$ that is piecewise affine and globally convex, there exist $\a,\b$ such that $h(\bx)=s[\a,\b](\bx),\forall \bx$.
\end{proposition}

This result follows from the fact that the pointwise maximum of a collection of convex functions is convex \cite{roberts1993convex}.
Using standard approximation arguments, it is easy to show that a max-affine spline can approximate an arbitrary (nonlinear) convex function arbitrarily closely \cite{nishikawa1998accurate}.

\subsection{Max-Affine Spline Operators}

A natural extension of an affine spline function is an {\em affine spline operator} (ASO) $S\big[A,B,\Omega^S\big]$ that produces a multivariate output. 
It is obtained simply by concatenating $K$ affine spline functions from (\ref{eq:lmas}). 
The details and a more general development are provided in Appendix~\ref{sec:sm_spline_op} 

We are particularly interested in the {\em max-affine spline operator} (MASO) $S[A,B]:\mathbb{R}^D \rightarrow \mathbb{R}^K$ 
formed by concatenating $K$ independent max-affine spline functions from (\ref{eq:maffine}). 
A MASO with slope parameters $A \in \mathbb{R}^{K \times R \times D}$ and offset parameters $B\in \mathbb{R}^{K \times R}$ is defined as
%
%
\begin{align}
    S\!\left[A,B\right]\!(\bx)
    & = \left[ 
    \begin{matrix}
    \max_{r=1,\dots,R}\langle [A]_{1,r,\bigcdot}, \bx\rangle+[B]_{1,r}\\
    \vdots \\
    \max_{r=1,\dots,R}\langle [A]_{K,r,\bigcdot}, \bx\rangle+[B]_{K,r}
    \end{matrix}
    \right]
     =: A[\bx]\,\bx+B[\bx].
        \label{eq:MASO}
\end{align}
The three subscripts of the slopes tensor $[A]_{k,r,d}$ correspond to output $k$, partition region $r$, and input signal index $d$.
The two subscripts of the bias tensor $[B]_{k,r}$ correspond to output $k$ and partition region $r$.
This equation also introduces the streamlined notation in terms of the signal-dependent matrix $A[\bx]$ and signal-dependent vector $B[\bx]$, where 
$[A[\bx]]_{k,\bigcdot}:=[A]_{k,r_k(\bx),\bigcdot}$ and $[B[\bx]]_{k}:=[B]_{k,r_k(\bx)}$ 
with $r_k(\bx)=\argmax_r \langle [A]_{k,r,\bigcdot}, \bx\rangle+[B]_{k,r}$.
Since a MASO is built from $K$ independent max-affine spline funcitons, it has a property analogous to Proposition~\ref{thm:masoequiv}.

\begin{proposition}
\label{thm:MASOequiv}
For any operator $H(\bx)=[h_1(\bx),\dots,h_K(\bx)]^T$ with $h_k\in \mathcal{C}^{0}(\mathbb{R}^D)~\forall k$ that are each piecewise affine and globally convex, there exist $A,B$ such that $H(\bx)=S[A,B](\bx)~\forall \bx$.
\end{proposition}

As above, using standard approximation arguments, it is easy to show that a MASO can approximate arbitrarily closely any (nonlinear) operator that is convex in each output dimension.

\subsection{Simplified Max-Affine Spline Operators}

A streamlined MAS formulation that we will not emphasize in this paper but that is sufficient in the DN context simplifies (\ref{eq:maffine}) to
\begin{align}
    s'[\a,\b',\Omega](\bx)
 =\max_{r=1,\dots,R} \big\langle [\a]_{r,\bigcdot} \,, \big( \bx +\b' \big) \big\rangle,
\label{eq:simpleMAS}
\end{align}
where the simplification is that the offset/bias $\b'$ is now constant vector with same dimension as the input $\bx$, and no longer a scalar depending on $r$.
The equivalence of (\ref{eq:simpleMAS}) to (\ref{eq:maffine}) under certain conditions is established in Appendix \ref{app:SMAS}.
A similar streamlining leads to the simplified MASO 
\begin{align}
    S'\!\left[A,\b'\right]\!(\bx)
    & = \left[ 
    \begin{matrix}
    \max_{r=1,\dots,R}\left\langle [A]_{1,r,\bigcdot}, \big(\bx+\b'\big)\right\rangle\\
    \vdots \\
    \max_{r=1,\dots,R}\left\langle [A]_{K,r,\bigcdot}, \big(\bx+\b'\big)\right\rangle
    \end{matrix}
    \right]
\label{eq:simpleMASO}
\end{align}
where $\b'$ is shared across multiple MAS.
The shared bias parameters $\b'$ restrict the MAS/MASO's modeling capacity, but they remain sufficient for DNs with activation functions like ReLU, leaky-ReLU, and absolute value and with linearly independent filters.

%% file: deepMaso.tex
\section{Deep Networks are Compositions of Spline Operators}
\label{sec:deepMaso}


The key enabling result of this paper is that 
a large class of DNs can be written as a composition of MASOs, one for each layer.
This includes many standard DNs topologies, including CNNs, ResNets, inception networks, maxout networks, network-in-networks, scattering networks, RNNs, and their variants, provided they use piecewise-affine and convex operators.
%
%

\subsection{DN Operators and Layers are MASOs}

We begin by showing that the DN operators defined in Section \ref{sec:dnop} are MASOs.
The proofs of these results are contained in Appendix~\ref{sec:proofs}. 

\begin{proposition}
An arbitrary fully connected operator $f^{(\ell)}_\bW$ is an affine mapping and hence a degenerate MASO $S\!\left[A_\bW^{(\ell)},B_\bW^{(\ell)}\right]$, with $R=1$, $[A^{(\ell)}_{\bW}]_{k,1,\bigcdot}=\left[\bW^{(\ell)}\right]_{k,\bigcdot}$ and $[B^{(\ell)}_{\bW}]_{k,1}=\left[\bb_{\bW}^{(\ell)}\right]_k$, leading to 
\begin{align}
& \bW^{(\ell)}\bz^{(\ell-1)}(\bx)+b_\bW^{(\ell)} = A_\bW^{(\ell)}[\bx]\bz^{(\ell-1)}(\bx)+B_\bW^{(\ell)}[\bx].
\end{align}
The same is true of a convolution operator with $\bW^{(\ell)},\bb_\bW^{(\ell)}$ replaced by $\bC^{(\ell)},\bb_{\bC}^{(\ell)}$.
\label{thm:conv}
\end{proposition}

\begin{proposition}
Any activation operator $f_\sigma^{(\ell)}$ using a piecewise affine and convex activation function is a MASO  
$S\!\left[\bA_\sigma^{(\ell)},\bB_\sigma^{(\ell)}\right]$ with 
$R=2$,  $\left[\bB^{(\ell)}_\sigma\right]_{k,1}=\left[\bB^{(\ell)}_\sigma\right]_{k,2}=0~\forall k$, and 
for ReLU
\begin{equation}
    \left[\bA^{(\ell)}_\sigma\right]_{k,1,\bigcdot}=0 ~~~ \left[\bA^{(\ell)}_\sigma\right]_{k,2,\bigcdot}=\be_k~~\forall k, 
\end{equation}
for leaky ReLU
\begin{equation}
    \left[\bA^{(\ell)}_\sigma\right]_{k,1,\bigcdot}=\nu \be_k, 
    ~~~\left[\bA^{(\ell)}_\sigma\right]_{k,2,\bigcdot}=\be_k~~\forall k,\nu>0, 
\end{equation}
and for absolute value
\begin{equation}
    \left[\bA^{(\ell)}_\sigma\right]_{k,1,\bigcdot}=-\be_k, 
    ~~~\left[\bA^{(\ell)}_\sigma\right]_{k,2,\bigcdot}=\be_k~~\forall k,
\end{equation}
where $\be_k$ represents the $k^{\rm th}$ canonical basis element of $\mathbb{R}^{D^{(\ell)}}$.
%



    
\label{thm:act}
\end{proposition}

To illustrate more specifically, here are the ReLU and absolute value activation operators written as MASOs:
\begin{align}
    S\big[A_{\rm ReLU},B_{\rm ReLU}\big](\bx)=\left[ 
    \begin{matrix}
    \max (0,[\bx]_1) \\
    \vdots\\
    \max (0,[\bx]_K)
    \end{matrix}
    \right],
    \quad\quad
    S\big[A_{\rm abs},B_{\rm abs}\big](\bx)=\left[ 
    \begin{matrix}
    \max (-[\bx]_1,[\bx]_1)\\
    \vdots\\
    \max (-[\bx]_K,[\bx]_K)
    \end{matrix}
    \right].
\end{align}

\begin{proposition}
Any pooling operator $f_\rho^{(\ell)}$ that is piecewise-affine and convex is a MASO  $S\big[\bA^{(\ell)}_\rho,\bB^{(\ell)}_\rho\big]$.\footnote
{
This result is agnostic of the type of pooling (spatial or channel).
The details are provided in Figure~\ref{fig:pooling} of Appendix~\ref{sm:poolingop}.
}
Max-pooling has $R=\#\mathcal{R}_k$ (typically a constant over all output dimensions $k$),
$\big[\bA^{(\ell)}_\rho\big]_{k,\bigcdot,\bigcdot}=\{ \be_{i},i \in \mathcal{R}_k\}$, 
and $\big[\bB^{(\ell)}_\rho\big]_{k,r}=0~\forall k,r$.
Average-pooling is a degenerate MASO with $R=1$, 
$\big[\bA^{(\ell)}_\rho\big]_{k,1,\bigcdot}=\frac{1}{\#(\mathcal{R}_k)}\sum_{i \in \mathcal{R}_k} \be_i$, and 
$\big[\bB^{(\ell)}_\rho\big]_{k,1}=0~\forall k$. 
\label{thm:pool}
\end{proposition}


Combinations of DN operators that result in a layer (recall Definition~\ref{def:layer}) are also MASOs.

\begin{proposition}
\label{thm:layerMASO}
A DN layer 
constructed from an arbitrary composition of fully connected/convolution operators 
followed by one activation or pooling operator 
is a MASO $S\big[A^{(\ell)},B^{(\ell)}\big]$ such that
\begin{equation}
    f^{(\ell)}(\bz^{(\ell-1)})=A^{(\ell)}[\bx]\bz^{(\ell-1)}(\bx)+B^{(\ell)}[\bx].
    \label{eq:masopixel}
\end{equation}
\end{proposition}
%

For example, a layer composed of 
an affine transform $S\!\big[\bA^{(\ell)}_\bW,\bB^{(\ell)}_\bW \big]$
followed by an activation operator $S\big[\bA^{(\ell)}_\sigma,\bB^{(\ell)}_\sigma \big]$ is a MASO  $S\big[\bA^{(\ell)},\bB^{(\ell)}\big]$
with parameters $\big[\bA^{(\ell)}\big]_{k,r,\bigcdot}=\bW^{(\ell)T}\big[\bA^{(\ell)}_\sigma\big]_{k,r,\bigcdot}$ and 
$\big[\bB^{(\ell)}\big]_{k,r}=\big[\bB^{(\ell)}_\sigma\big]_{k,r}+
\bb_{\bW}^{(\ell)T} \big[\bA^{(\ell)}_\sigma\big]_{k,r,\bigcdot}$.
%
%
When $D^{(\ell)}>D^{(\ell-1)}$, such a composition is an {\em expansion layer} that up-samples its input to increase its dimensionality. Such layers are common in autoencoders \cite{vincent2008extracting} and in the generator of a generative adversarial network (GAN) \cite{goodfellow2014generative}.
Appendix~\ref{sec:proof_rewritting} contains several additional examples of rewriting a composition of multiple DN operators as a MASO layer mapping.


\subsection{Compositions of MASOs}

The development of the previous section enables us to state a general result for DNs.

\begin{theorem}
A DN constructed from an arbitrary composition
of fully connected/convolution, activation, and pooling operators of the types covered in Propositions~\ref{thm:conv}--\ref{thm:pool} or layers from Proposition \ref{thm:layerMASO} is a composition of MASOs.
Moreover, the overall composition is itself an ASO.
\label{thm:big2}
\end{theorem}

\sloppy
DNs covered by Theorem~\ref{thm:big2} include CNNs, ResNets, inception networks, maxout networks, network-in-networks, scattering
networks, and their variants using connected/convolution operators, (leaky) ReLU or absolute value activations,
and max/mean pooling.

Note carefully that DNs of the form stated in Theorem~\ref{thm:big2} are not MASOs, in general, since the composition of two or more MASOs is not necessarily a MASO (it is, at least, an ASO).
Indeed, a composition of MASOs remains a MASO if and only if all of the intermediate operators are non-decreasing with respect to each of their output dimensions \cite{boyd2004convex}.
Interestingly, ReLU, max-pooling, and average pooling are all non-decreasing, while leaky ReLU is strictly increasing. 
The culprits of the non-convexity of the composition of operators are negative entries in the fully connected and convolution operators.
A DN where these culprits are thwarted is an interesting special case, because it is convex with respect to its input \cite{amos2016input} and multiconvex \cite{xu2013block} with respect to its parameters (i.e., convex with respect to each operator's parameters while the other operators' parameters are held constant).

\begin{proposition}
\label{thm:increasing}
A DN layer is nondecreasing with respect to each of its output dimensions if and only if its slope parameters are nonnegative
\begin{align}
\left[A^{(\ell)}\right]_{k,r,d} \geq  0.
\label{eq:nondec}
\end{align}
\end{proposition}

The condition (\ref{eq:nondec}) holds if only if the layer's activation or pooling operators are nondecreasing and the entries of any fully connected or convolution matrices satisfy  
$\big[\bW^{(\ell)}\big]_{k,d},
\big[\bC^{(\ell)}\big]_{k,d} \geq 0$.\footnote{Alternatively, the 
activation or pooling operators can be nonincreasing and the entries of any fully connected or convolution matrices can satisfy  
$\big[\bW^{(\ell)}\big]_{k,d},
\big[\bC^{(\ell)}\big]_{k,d} \leq 0$.}

\begin{theorem}
An $L$-layer DN whose layers $\ell=2,\dots , L$ are not only piecewise-affine and convex but also nondecreasing with respect to each of their output dimensions is globally a MASO and thus also globally convex with respect
to each of its output dimensions.
\label{thm:big3}
\end{theorem}

Note that Theorem \ref{thm:big3} remains true even if the DN's first layer ($\ell=1$) is {\em not} non-decreasing.
For example, it could include fully connected/convolution matrices with negative values and/or the absolute value activation function.

\subsection{DNs are Signal-Dependent Affine Transformations}
\label{sec:dnaffine}

The second conclusion in Theorem~\ref{thm:big2} is that the mapping from the input $\bx$ to the layer output $\bz^{(\ell)}(\bx)$ is an ASO.
Hence, $\bz^{(\ell)}(\bx)$ is a {\em signal-dependent, piecewise affine transformation} of $\bx$.
(Much more on this below in Section~\ref{sec:template}.)
The particular affine mapping applied to $\bx$ depends on which partition of the spline it falls in $\R^D$.
(Much more on this in Section~\ref{sec:partition} below.)
As such, a more precise but unwieldy terminology that we will not emphasize would be that $\bz^{(\ell)}(\bx)$ is a {\em partition-region-dependent, piecewise affine transformation} of $\bx$ (see Remark \ref{rem:sig1} below).

The above results pertain to DNs using convex, piecewise affine operators.
DNs using operators that are convex but not piecewise affine can be approximated arbitrarily closely by a MASO. 
DNs using operators that are non-convex (e.g., the sigmoid and arctan activation functions) can be approximated by an ASO (that itself is a composition of MASOs).
We investigate this approximation in more detail below in Section~\ref{sec:universality}.

For any DN covered by Theorem \ref{thm:big2}, we can substitute (\ref{eq:MASO}) into (\ref{eq:layers2}) to obtain an explicit formula for the layer output $\bz^{(\ell)}(\bx)$ in terms of the input $\bx$.
We now illustrate with two examples, namely CNNs and ResNets.

\subsection{Example: CNN Affine Mapping Formula}
\label{sec:formula}


The explicit input/output formula for a standard CNN (using convolution, ReLU activation, max-pooling, and a final fully connected operator) is given by
\begin{multline}
    \bz_{\rm CNN}^{(L)}(\bx) = 
    \bW^{(L)}
    \underbrace{\left(
    \prod_{\ell=L-1}^1 A_{\rho}^{(\ell)}[\bx] A^{(\ell)}_{\sigma}[\bx] \bC^{(\ell)} \right)}_{A_{\rm CNN}[\bx]}
    \bx
    \\
    + ~ \bW^{(L)}\underbrace{
    \sum_{\ell=1}^{L-1}
    \left(
    \prod_{j=L-1}^{\ell+1}A_{\rho}^{(j)}[\bx]
    A^{(j)}_{\sigma}[\bx] \bC^{(j)}
    \right)\!
    \left(A_{\rho}^{(\ell)}[\bx]
    A^{(\ell)}_{\sigma}[\bx] \bb_{\bC}^{(\ell)}
    \right)}_{B_{\rm CNN}[\bx]}
    +~ \bb_{\bW}^{(L)}.
    \label{eq:CNNaffine}
\end{multline}
Here, the
$A^{(\ell)}_{\sigma}[\bx]$ are signal-dependent matrices corresponding to the ReLU activation functions, 
$A_{\rho}^{(\ell)}[\bx]$ are signal-dependent matrices corresponding to the max-pooling operators, and the biases $\bb_{\bW}^{(L)},\bb_{\bC}^{(\ell)}$ arise directly from the fully connected and convolution operators. The absence of $B^{(\ell)}_\sigma[\bx]$ and $B_\rho^{(\ell)}[\bx]$ is due to the absence of bias in the ReLU (recall (\ref{thm:act})) and max-pooling operators (recall (\ref{thm:pool})). 

\begin{remark}
\label{rem:sig1}
The affine transformation parameters $A[\bx]$, $B[\bx]$ just above and throughout the development below are dependent not only on the signal $\bx$ but also more generally on the spline partition region containing $\bx$ (see Remark \ref{rem:partsig1} below in Section \ref{sec:vq}).
To streamline our notation, however, we will note the dependence simply by $[\bx]$ and refer to it as ``signal-dependent'' rather than than the more precise ``partition-region-of-$\bx$-dependent.''
\end{remark}

Inspection of (\ref{eq:CNNaffine}) reveals the exact form of the signal-dependent, piecewise affine mapping linking $\bx$ to $\bz_{\rm CNN}^{(L)}(\bx)$.
Moreover, this formula can be collapsed into 
\begin{equation}
\bz_{\rm CNN}^{(L)}(\bx) = 
    \bW^{(L)} \big( A_{\rm CNN}[\bx]\,\bx +B_{\rm CNN}[\bx]\big) +     \bb_{\bW}^{(L)}
    \label{eq:CNNaffine1}
\end{equation}
from which we can recognize 
\begin{equation}
\bz_{\rm CNN}^{(L-1)}(\bx)=A_{\rm CNN}[\bx]\,\bx + B_{\rm CNN}[\bx]
\label{eq:CNNaffine1a}
\end{equation}
as an explicit, signal-dependent, affine formula for the {\em featurization process} that aims to convert $\bx$ into a set of (hopefully) linearly separable features that are then input to the linear classifier in layer $\ell=L$ with parameters $\bW^{(L)}$ and $\bb_{\bW}^{(L)}$. 
More generally, we will use the notation 
\begin{equation}
    \bz_{\rm CNN}^{(\ell)}(\bx)=A^{(\ell)}_{\rm CNN}[\bx]\,\bx + B^{(\ell)}_{\rm CNN}[\bx]
    \label{eq:CNNaffine1b}
\end{equation}
but will omit the superscript when $\ell=L$ (as in (\ref{eq:CNNaffine1a})).
When a DN is terminated using more than one fully connected layer, then all but the last layer can be incorporated into $A_{\rm CNN}[\bx]$ and $B_{\rm CNN}[\bx]$ as the class-agnostic representation of $\bx$; only the last fully connected layer is required in the linear classifier.

Of course, the final prediction $y$ is formed by running $\bz_{\rm CNN}^{(L)}(\bx)$ through a softmax nonlinearity $g$, to create a probability distribution.
The same argument applies in the regression context if we suppress the softmax nonlinearity.

Some interesting properties of CNNs can be inferred from (\ref{eq:CNNaffine}) thanks to its explicit nature.
For example, what amounts to the matrix
\begin{equation}
    A_{\rm CNN}[\bx]=\prod_{\ell=L-1}^1 A_{\rho}^{(\ell)}[\bx] A^{(\ell)}_{\sigma}[\bx] \bC^{(\ell)}
    \label{eq:CNNmat}
\end{equation}
has been probed and empirically shown to converge to the zero matrix as $L$ grows.
This observation has motivated skip-connections and ResNets \cite{targ2016resnet}, to which we now turn.

\subsection{Example: ResNet Affine Mapping Formula}
\label{sec:resnet}

A ResNet layer leverages a skip-connection that ``short-circuits'' a layer from its input to its output \cite{targ2016resnet}. 
We study here the simplest form of skip-connection that comprises a convolution operator followed by an activation operator in parallel with a second convolution operator (the so-called linear skip-connection) as in
\begin{align}
\bz^{(\ell)}(\bx)&= \bC^{(\ell)}_{\rm skip}\bz^{(\ell-1)}(\bx) +A^{(\ell)}_{\sigma}[\bx]\left(\bC^{(\ell)}\bz^{(\ell-1)}(\bx)+\bb_{\bC}^{(\ell)}\right)+\bb^{(\ell)}_{\rm skip}.   \label{eq:resnet}
\end{align}
The  notation ``skip'' distinguishes the parameters for the skip-connection.
Such a topology results in state-of-the-art performance in standard image classification tasks \cite{targ2016resnet}.
Skip-connections that skip over more than one layer can be formulated simply by adding additional terms of the form $\bC^{(\ell)}_{\rm skip-2}\bz^{(\ell-2)}(\bx)+\bC^{(\ell)}_{\rm skip-3}\bz^{(\ell-3)}(\bx)+\dots$, to (\ref{eq:resnet}).
Pooling in a standard ResNet is accomplished via dyadic subsampling over pixel space and thus corresponds to a simple subsampling of the rows of $\bC_{\rm skip}^{(\ell)}, A^{(\ell)}[\bx]$ and $\bb_{\rm skip}^{(\ell)}$ in (\ref{eq:resnet});  the addition of more complex layers is straightforward.

The input/output formula for a standard ResNet (using convolution, ReLU activation, max-pooling, and a final fully connected operator) is given by
\begin{multline}
\bz_{\rm RES}^{(L)}(\bx)  = \small
\bW^{(L)}\underbrace{\left( \prod_{\ell=L-1}^1\left(A_{\sigma}^{(\ell)}[\bx]\bC^{(\ell)}+ \bC^{(\ell)}_{\rm skip} \right) \right)}_{A_{\rm RES}[\bx]}\bx
\\
+~
\bW^{(L)}\underbrace{\sum_{\ell=L-1}^1\left(\prod_{i=L-1}^{\ell+1}( A_{\sigma}^{(\ell)}[\bx]\bC^{(\ell)}+ \bC^{(\ell)}_{\rm skip})\right)\left( A_{\sigma}^{(\ell)}[\bx]\bb_{\bC}^{(\ell)}+\bb_{\rm skip}^{(\ell)} \right)}_{B_{\rm RES}[\bx]} +~\bb_{\bW}^{(L)}.
\normalsize
\label{eq:CNNresnet}
\end{multline}
Comparison of (\ref{eq:CNNaffine}) and (\ref{eq:CNNresnet}) reveals that CNNs and ResNets share the same overall functional form (in fact, any composition of MASOs will have this form) of a featurizer followed by a linear classifier (recall (\ref{eq:CNNaffine1}) and (\ref{eq:CNNaffine1a})).
However, in a ResNet, there exists a linear path from the input to the output of any layer.
We can distribute this linear path based on (\ref{eq:CNNresnet}) to highlight that a ResNet can be interpreted as an ensemble of CNNs of all depths $\leq L$ \cite{veit2016residual} with tied weights (same weights shared across models)
\begin{align}
\bz^{(L)}_{\rm RES}(\bx)
=&~
\bb_{\bW}^{(L)}+\bW^{(L)}\Big(\bb_{\rm RES}[\bx]\nonumber \\
&\left.
\begin{array}{l}
    +~\prod_{\ell=L-1}^1 \bC^{(\ell)}_{\rm skip}\bx\\
    +~\prod_{\ell=L-1}^2 \bC^{(\ell)}_{\rm skip}A_{\sigma}^{(1)}[\bx]\bC^{(1)}\bx\\
    +~\prod_{\ell=L-1}^3 \bC^{(\ell)}_{\rm skip}A_{\sigma}^{(2)}[\bx]\bC^{(2)}A_{\sigma}^{(1)}[\bx]\bC^{(1)}\bx\\
    +~ \dots\\
    +~\prod_{\ell=L-1}^1 A_{\sigma}^{(\ell)}[\bx]\bW^{(\ell)}  \bx \Big).
    \end{array} \right\}\parbox{3em}{\scriptsize CNN \\ Ensemble}
    \label{eq:CNNresnet2}
\end{align}

Skip-connections and the resulting ensemble-of-CNN's structure prevent some of the detrimental stability behavior of very deep CNNs (large $L$) that was elucidated by \cite{targ2016resnet}.
Indeed, the skip connections modify the matrix (\ref{eq:CNNmat}) that converges to zero with $L$ to 
$\prod_{\ell=L-1}^1\left(A_{\sigma}^{(\ell)}[\bx]\bC^{(\ell)}+ \bC^{(\ell)}_{\rm skip} \right)$, which is stabilized by the presence of $\bC^{(\ell)}_{\rm skip}$.

Formulas analogous to (\ref{eq:CNNaffine})--(\ref{eq:CNNresnet}) can be derived for a large class of DNs; for exmaple, we include a formula for RNNs in Appendix~\ref{sec:sm_RNN}.

\subsection{Universality of MASO DNs}
\label{sec:universality}

The ability of certain DNs to approximate an arbitrary functional/operator mapping has been well established \cite{cybenko1989approximation} using squashing activation functions ($\sigma:\mathbb{R} \rightarrow [0,1]$). 
While it is clear that a convex operator can be approximated arbitrarily closely by a single MASO \cite{magnani2009convex}, it is not clear a priori whether a composition of multiple MASOs is capable of approximating an arbitrary operator. 
Leveraging a result from \cite{breiman1993hinging}, we prove in Appendix~\ref{sec:proof_universality} that the composition of just two MASOs has a universal approximation property.

\begin{theorem}
\label{thm:universality}
Let $f(\bx)$ represent an arbitrary continuous operator $f:\mathbb{R}^{D}\rightarrow \mathbb{R}^C$,
and let $f_\Theta(\bx)$ represent the output of an arbitrary DN with parameters $\Theta$ constructed from a composition of $L=2$ MASOs with $R^{(1)}=2$ and $R^{(2)}=1$ (hence linear second layer).
Then there exist (optimal) parameters $\Theta$ such that
\begin{equation}
    \left\| f(\bx) - f_\Theta(\bx) \right\|^2_2 \leq \frac{D^{(2)}\sum_{k=1}^C \eta_k}{D^{(1)}},
\end{equation}
where $\eta_k$ is a constant characterizing the regularity of the $k^{\rm th}$ output of $f$.
\end{theorem}

The capability of a shallow DN ($L=2$) with many partition regions \big(large $R^{(1)}$\big) to approximate a complicated operator has been studied numerically in \cite{zhang2016understanding}, where both two-layer perceptrons and CNNs were trained to memorize a training set with randomly permuted labels. 
Understanding the approximation capabilities of DNs for $L>2$ remains open and should be studied not only from an approximation theory perspective (since such a DN can memorize an arbitrary finite dataset) but also from a generalization perspective.

\begin{corollary}
A DN containing non-convex and/or non-piecewise affine operators can be approximated arbitrarily closely by a MASO-based DN. 
\label{cor:approxMaso}
\end{corollary}

Corollary \ref{cor:approxMaso} expands the scope of our theory to include, for example, the sigmoid-based activations that are common in RNNs such as LSTMs and GRUs \cite{graves2013generating,cho2014learning,chung2014empirical}.
In Appendix~\ref{sec:sm_RNN} we work out the details of a piecewise affine spline approximation to a sigmoid based RNN.


In addition of being universal approximators, MASO-based DNs also appear naturally as solutions of an infinite-dimensional optimization for the optimal DN nonlinearities \cite{unser2018representer} (see Appendix \ref{sec:proof_unser}).
The approximation properties of DNs have been further studied in the ReLU case, providing additional insights \cite{hanin2017approximating,koehler2018representational}.

%% file: template.tex
\section{DNs are Template Matching Machines}
\label{sec:template}

We now take a deeper look into the featurization/prediction process of (\ref{eq:CNNaffine1}) in order to bridge DNs and classical optimal classification theory via matched filters (aka ``template matching'').
We will focus on classification for concreteness.
Our analysis holds
for any DN meeting the conditions of Theorem \ref{thm:big2}, for example, the CNN in (\ref{eq:CNNaffine}) or the ResNet in (\ref{eq:CNNresnet}). 

\subsection{Matched Filter Template Matching}
\label{sec:tm}

Rewriting (\ref{eq:CNNaffine1}) (and removing the CNN demarcation, because the result holds in more generality) as
\begin{equation}
\bz^{(L)}(\bx) = 
    \left(\bW^{(L)} A[\bx]\right) \bx 
+ \left(\bW^{(L)} B[\bx] + \bb_{\bW}^{(L)}\right)
    \label{eq:CNNaffine1c}
\end{equation}
provides the alternate interpretation that $\bz^{(L)}(\bx)$ is the output of a bank of {\em linear matched filters} (plus a set of biases). That is, the $c^{\rm th}$ element of $\bz^{(L)}(\bx)$ equals the inner product between the signal $\bx$ and the matched filter for the $c^{\rm th}$ class, which is contained in the $c^{\rm th}$ row of the matrix $\bW^{(L)} A[\bx]$.
The bias $\bW^{(L)}B[\bx] + \bb_{\bW}^{(L)}$ can be used to account for the fact that some classes might be more likely than others (i.e., the prior probability over the classes). 
It is well-known that a matched filterbank is the optimal classifier for deterministic signals in additive white Gaussian noise \cite{rabiner1975theory}.
Given an input $\bx$, the class decision is simply the index of the largest element of $\bz^{(L)}(\bx)$.\footnote
{Again, since the softmax merely rescales the entries of $\bz^{(L)}(\bx)$ into a probability distribution, it does not affect the location of its largest element.}

\subsection{Hierarchical Template Matching}
\label{sec:htm}

Yet another interpretation of (\ref{eq:CNNaffine1}) is that $\bz^{(L)}(\bx)$ is computed not in a single matched filter calculation but hierarchically as the signal propagates through the DN layers. 
Abstracting (\ref{eq:MASO}) to write the per-layer maximization process as $\bz^{(\ell)}(\bx)=\max_{r^{(\ell)}} A^{(\ell)}_{r^{(\ell)}}\bz^{(\ell-1)}(\bx)+B^{(\ell)}_{r^{(\ell)}}$ and cascading, obtain a formula for the end-to-end DN mapping
\begin{align}
&    \bz^{(L)}(\bx) = W^{(L)}\max_{r^{(L-1)}}
\left( \bA_{r^{(L-1)}}^{(L-1)} \max_{r^{(2)}} \left( \bA_{r^{(2)}}^{(2)} \; \dots  \max_{r^{(1)}}  \left(\bA_{r^{(1)}}^{(1)}\bx+\bB^{(1)}_{r{(1)}}\right)+\bB^{(2)}_{r^{(2)}} \right)\dots +\bB^{(L-1)}_{r^{(L-1)}} \right)+b_{W}^{(L)}.
    \label{eq:maxDN1}
\end{align}
This formula elucidates that a DN performs a {\em hierarchical, greedy template matching} on its input, a computationally efficient yet sub-optimal template matching technique. 
Such a procedure is globally optimal when the DN is globally convex.


\begin{corollary}
\label{cor:gopt}
For a DN abiding by the requirements of Theorem \ref{thm:big3}, the computation (\ref{eq:maxDN1}) collapses to the following globally optimal template matching 
\begin{align}
\bz^{(L-1)}(\bx) = W^{(L)}\max_{r^{(L-1)}, r^{(2)}, \dots, r^{(1)} }
\big( \bA_{r^{(L-1)}}^{(L-1)}  \big( \bA_{r^{(2)}}^{(2)}  \dots \big(\bA_{r^{(1)}}^{(1)}\bx+\bB^{(1)}_{r^{(1)}}\big)+\bB^{(2)}_{r^{(2)}} \big)\dots +\bB^{(L-1}_{r^{(L-1)}} \big)+b_W^{(L)}.
    \label{eq:maxDN2}
\end{align}
\end{corollary}

The global optimality of  hierarchical template matching \label{eq:maxDN1} with positive filter weights has been shown for the special case of CNNs in 
 \cite{patel2016probabilistic,amos2016input}.
Corollary \ref{cor:gopt} extends this result to arbitrary MASO DNs.


\subsection{Template Visualization Examples}
\label{sec:vizua}

Since the complete DN mapping (up to the final softmax) can be expressed as in (\ref{eq:CNNaffine1}),\footnote
{Recall that this formula is not restricted to CNNs.} 
given a signal $\bx$, the signal-dependent template for class $c$ is given by
\begin{equation}
    A[\bx]_c=\frac{d [\bz^{(L)}(\bx)]_{c}}{d \bx}, 
    \label{eq:saliency}
\end{equation}
which can be efficiently computed via backpropagation \cite{hecht1988theory}.\footnote
{In fact, we can use the same backpropagation procedure used for computing the gradient with respect to a fully connected or convolution weight but instead with the input $\bx$. This procedure is becoming increasingly popular in the study of adversarial examples \cite{szegedy2013intriguing}.}
Once the template $A[\bx]_c$ has been computed, the bias term $b[\bx]_c$ can be computed as $b[\bx]_c=\bz^{(L)}(\bx)_c-\langle A[\bx]_{c,\bigcdot},\bx\rangle$. 
This method of computation is typically much more efficient than using the explicit formulas (\ref{eq:CNNaffine}), (\ref{eq:CNNresnet}).

\begin{figure}[t]
\centering
\includegraphics[width=0.6\textwidth]{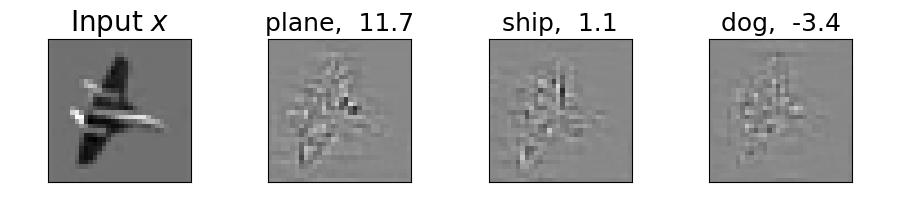}
\\[-0mm]
(a) {\em largeCNN}, ReLU activation, no BN
\\
\includegraphics[width=0.6\textwidth]{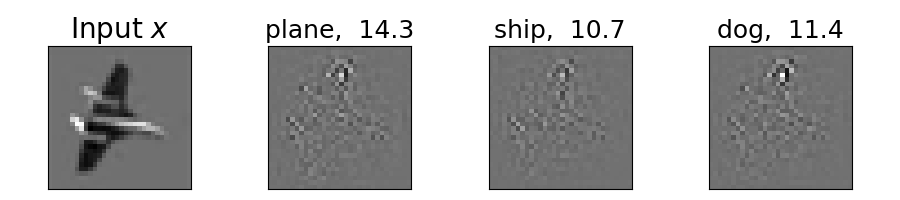}
\\[-0mm]
(b) {\em largeCNN}, absolute value activation, no BN
\\
\includegraphics[width=0.6\textwidth]{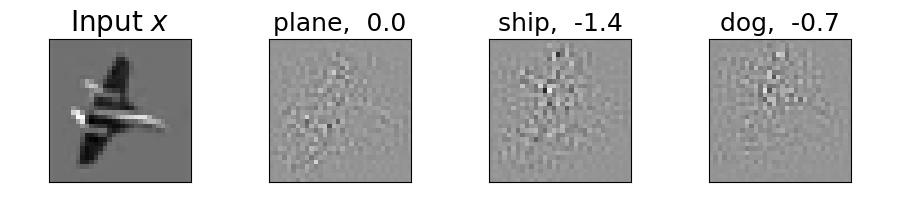}
\\[-0mm]
(c) {\em largeCNN}, ReLU activation, BN
\\
\includegraphics[width=0.6\textwidth]{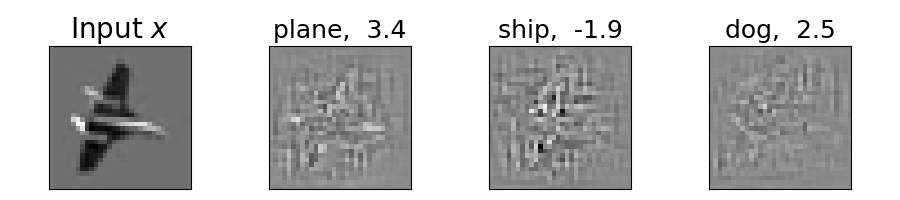}
\\[-0mm]
(d) {\em largeCNN}, absolute value activation, BN
\caption{
Visualization of the signal-dependent, matched filter templates generated by the {\em largeCNN} architecture employing max-pooling, either ReLU or absolute value activation function, and batch normalization (BN) or not (see Section~\ref{sec:templateBias} for more details) in a classification task with the CIFAR10 dataset.
For ease of visualization only, the images were converted to gray scale by averaging their RGB components.
To the right of the input image $\bx$ of class `airplane' we depict the three induced templates for the classes
`plane,' `ship,' and `dog.'
The color map for the templates maps black to large negative values, grey to zero, and white to large positive values.
Above each of the templates, we indicate the inner product between $\bx$ and the corresponding template.
(The final classification will
involve not only these inner products but also the relevant biases and softmax transformation.)
}
\label{fig:tempVisA}
\end{figure}
\begin{figure}[t]
\centering
\begin{minipage}{0.48\linewidth}
\includegraphics[width=1\textwidth]{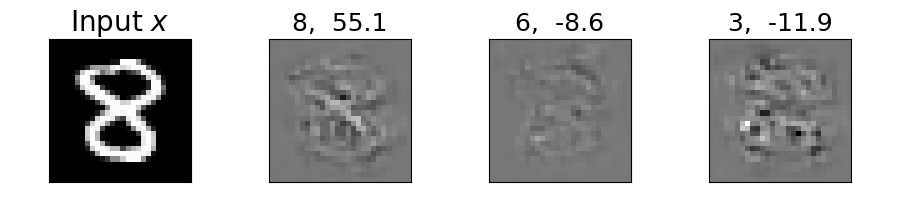}\\
   \centering
   (a)  {\em smallResNet}, ReLU activation, no BN
  \\
\includegraphics[width=1\textwidth]{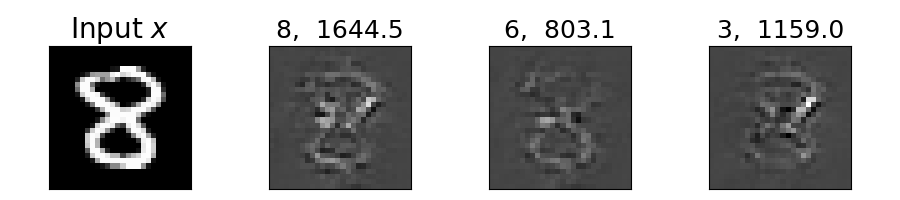}\\
   \centering
(c)  {\em smallResNet}, absolute value activation, no  BN
\end{minipage}
\begin{minipage}{0.48\linewidth}
   \includegraphics[width=1\textwidth]{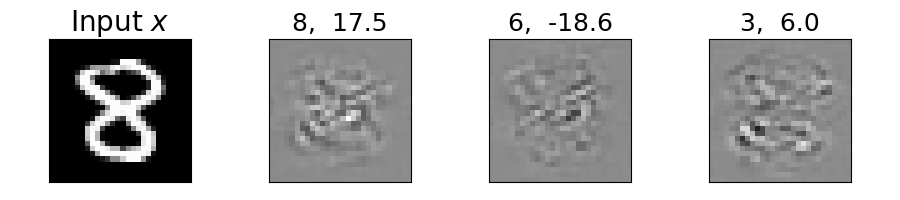}
   \\
   \centering
(b) {\em smallResNet}, ReLU activation, BN \\
   \includegraphics[width=1\textwidth]{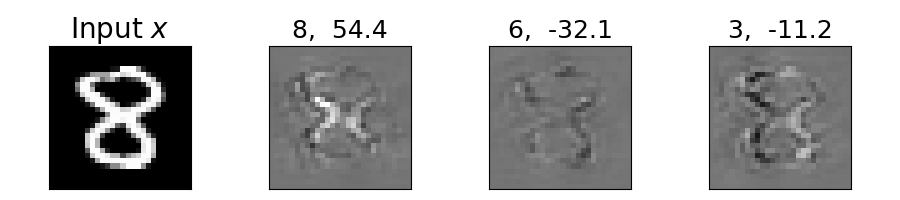}
   \\
   \centering
(d) {\em smallResNet}, absolute value activation, BN
\end{minipage}
\caption{
Visualization of the signal-dependent, matched filter templates generated by the {\em smallResNet} architecture employing max-pooling, either ReLU or absolute value activation function, and batch normalization (BN) or not in a classification task with the MNIST dataset.
To the right of the input image $\bx$ of class `8' we depict the three induced templates for the classes
`8,' `6,' and `3.'
}
\label{fig:tempVisC}
\end{figure}

Figures \ref{fig:tempVisA}--\ref{fig:tempVisC} plot the signal-dependent, matched filter templates for two experiments with the CIFAR10 and MNIST datasets.
We show results for two different DNs -- {\em largeCNN} and {\em smallResNet} (see Appendix \ref{ap:dntopology} for the definitions) -- using two different activation functions -- ReLU and absolute value.
We trained for $75$ epochs on CIFAR10 and $30$ epochs on MNIST using a batch-size of $50$ and the Adam optimizer with a constant learning rate of $0.0005$. 


The visual resemblance of the templates to an `airplane'/`8' or an `anti-airplane'/`anti-8' is intriguing.
The inner products between the input image and the templates also conforms to our intuition regarding the behavior of a matched filter bank.
That is, while the templates might seem rather incoherent on visual inspection, they are in fact tuned to the input in order to produce a high inner product for the correct class.

What we call ``templates'' have been probed in the deep learning community and called {\em saliency maps} \cite{simonyan2013deep,zeiler2014visualizing}.
In a result of independent interest, the above development proves that the saliency map technique is exact for MASO DNs. 
Indeed, a first-order Taylor expansion provides an exact representation of a piecewise affine mapping but only an approximation of a non-piecewise mapping like a sigmoid.


\begin{proposition}
The saliency map technique based on 
(\ref{eq:saliency})
results in {\em exact} saliency maps for a DN with MASO layers and otherwise results in a first-order approximation.
\label{thm:saliency}
\end{proposition}

\subsection{Template Matching vs.\ Bias}
\label{sec:templateBias}

In emphasizing the visual appearance of the signal-dependent templates $A[\bx]_{c,\bigcdot}$ in Figures~\ref{fig:tempVisA}--\ref{fig:tempVisC}, we have neglected 
two additional important factors that can play major roles in determining the outcome of the classification process.

The first factor is the {\em bias}
$\bW^{(L)} B[\bx] + \bb_{\bW}^{(L)}$ from (\ref{eq:CNNaffine1a}).
Indeed, a large positive or negative bias can swamp the inner product between the input image and the template and throw the election for the candidate class. 
Note that this bias includes not only the final linear classifier bias (which can be interpreted as a prior over the classes) but also all of the per-layer biases (mapped to the output).

The second factor is whether {\em batch normalization} \cite{ioffe2015batch} is applied during learning and inference.
Batch normalization (BN) is a specific kind of ``data centering'' process that is often applied between the fully connected/convolution and activation operators.
It replaces $\bz^{(\ell)}(\bx)$ with 
\begin{align}
    \frac{\bz^{(\ell)}(\bx)-{\bf m}}
    {\sqrt{\nu^2+\epsilon}}\: \gamma + \zeta
\label{eq:BN}
\end{align}
where ${\bf m}$ and $\nu$ are computed by averaging over a batch of images/feature maps and $\gamma$ and $\zeta$ are learned ($\epsilon$ is a small constant to prevent instability). During inference, the parameters ${\bf m}$ and $\nu$ are replaced by the averages of the values computed during training. 
This centering introduces a bias into output of the DN that is distinct from that in (\ref{eq:CNNaffine1a}).

To explore the performance impacts of these two factors, we expanded on the experiments that generated Figures~\ref{fig:tempVisA}--\ref{fig:tempVisC} by testing and training the {\em largeCNN} with/without the bias term $\bW^{(L)} B[\bx] + \bb_{\bW}^{(L)}$ and with/without batch normalization.
(We trained without the bias term simply by initializing all of the biases in the network to zero as usual and not training them.)
For each of the four training scenarios, we classified both the training and test datasets using two predictors:
(\ref{eq:CNNaffine1a}) for the ``bias'' scenario and $y=W^{(L)}A[\bx]$ for the ``\st{bias}'' scenario.
The results for each of the four experiments are displayed in Tables~\ref{tab:biasnobiasMNIST} and~\ref{tab:biasnobiasCIFAR10} for the MNIST and CIFAR10 datasets, respectively.  

\begin{table}[t]
    \centering
    \small
    MNIST \\[2mm]
    \begin{tabular}{l|cccc} 
    DN & \multicolumn{2}{c}{Train} & \multicolumn{2}{c}{Test} \\
    Configuration & (bias) & (\st{bias}) & (bias) & (\st{bias}) \\ \hline
    \st{bias},\st{BN}    
    & 99.9 & 99.9 & 99.4 & 99.4 \\ \hline
    bias,\st{BN}       
    & 100 & 100 & 99.4 & 99.4 \\ \hline
    \st{bias},BN     
    & 68.1 & 68.1 & 67.9 & 67.9 \\ \hline
     bias,BN    
    & 100 & 72 & 99.5 & 81.9 \\ \hline
    \end{tabular}
    \caption{
    Classification results for the {\em largeCNN} trained and tested on the MNIST dataset with/without bias and with/without batch normalization (BN).
    (In the two \st{bias} rows, some entries are repeated, because performance is unchanged by including bias in the classifier.)
    }
    \label{tab:biasnobiasMNIST}
\end{table}

\begin{table}[t]
    \centering
    \small
    CIFAR10 \\[2mm]
    \begin{tabular}{l|cccc} 
    DN& \multicolumn{2}{c}{Train} & \multicolumn{2}{c}{Test} \\
    Configuration& (bias) & (\st{bias}) & (bias) & (\st{bias}) \\ \hline
    \st{bias},\st{BN}   
    & 98.8 & 98.8 & 76.0 & 76.0 \\ \hline
    bias,\st{BN}       
    & 99.4 & 81.4 & 78.5 & 66.5 \\ \hline
    \st{bias},BN    
    & 28.0 & 28.0 & 26.9 & 26.9 \\ \hline
    bias,BN    
    & 98.2 & 38.2 & 80.6 & 34.6 \\ \hline
    \end{tabular}
    \caption{
    Classification results for the {\em largeCNN} trained on the CIFAR10 dataset. (See Table~\ref{tab:biasnobiasMNIST} for more details.)
    }
    \label{tab:biasnobiasCIFAR10}
\end{table}

Tables~\ref{tab:biasnobiasMNIST} and~\ref{tab:biasnobiasCIFAR10} provide several useful insights.
Row 1: 
Training with neither bias nor batch normalization and predicting based only on the template matching results in a high-performance and competitive classifier. 
This both confirms our matched filterbank interpretation and indicates that the templates are very informative for the task.
Row 2:
Training with bias but not batch normalization (standard practice as of a few years ago) is stable (from training to testing) for MNIST but not CIFAR10, most likely thanks to the lack of nuisance variation in MNIST (same color for all digits, no background, occlusions, nor out-of-plane rotations) as compared to CIFAR10.
Rows 3 and 4:
While batch normalization 
has been shown to accelerate training and improve accuracy, comparison of Rows 3 and 4 reveals that it is crucial to include the bias term.

\subsection{Stability and Lipschitz Constant}

The MASO formulation of DNs enables a straightforward analysis of its {\em Lipschitz constant}. 
Recall that 
the Lipschitz constant $\kappa$ of a continuous and differentiable mapping equals the infinity norm of its total derivative. 
As such, it quantifies its regularity, or how fast the mapping can change.

There are several immediate applications of such a regularity analysis to DNs.
The Lipschitz constant plays a key role in analyzing the generalization capacity of a DN based on flat minima \cite{hochreiter1997flat}. 
The Lipschitz constant also quantifies the volatile behavior of DNs that makes them succeptible to adversarial attacks \cite{kurakin2016adversarial}.
Adversarial attacks leverage the non-contractivity of a DN (large Lipschitz constant) to highly transform an output prediction given only a slightly perturbed input.

\begin{proposition}
For a MASO DN layer $f_\Theta^{(\ell)}$, we have
\begin{equation}
    \left\| f_\Theta^{(\ell)}(\bz_1)-f_\Theta^{(\ell)}(\bz_2) \right\| ^2
    \leq 
    \sum_{k=1}^{D^{(\ell)}}\max_{r=1,\dots,R^{(\ell)}}
    \left\|\big[A^{(\ell)}\big]_{k,r,\bigcdot}\right\|^2 
    \left\| \bz_1 - \bz_2\right\|^2
\end{equation}
and thus its Lipschitz constant \begin{equation}
\kappa^{(\ell)}=\sum_{k=1}^{D^{(\ell)}}\max_{r=1,\dots,R}\left\|\big[A^{(\ell)}\big]_{k,r,\bigcdot}\right\|^2.
\end{equation}
\end{proposition}

In Appendix~\ref{proof_lipschi} we calculate the Lipschitz constants of the standard DN operators and a complete DN constructed from them.

\begin{proposition}
\label{thm:lipschi}
The Lipschitz constants for the standard DN operators are given by:  
fully connected $\kappa_{\bW}^{(\ell)}\leq\|\bW^{(\ell)}\|^2$;
convolution
$\kappa_{\bC}^{(\ell)}\leq\|\bC^{(\ell)} \|^2$;
ReLU, leaky-ReLU, absolute value
$\kappa_{\sigma}^{(\ell)}\leq(D^{(\ell)})^2$;
max-pooling
$\kappa_{\rho}^{(\ell)}\leq(D^{(\ell)})^2$;
softmax 
$\kappa\leq \frac{C-1}{C^2}$.
\end{proposition}

\begin{proposition}
For a DN $f_\Theta$ constructed by composing $L$ MASOs and any two input signals $\bx_1, \bx_2$, we have  
\begin{equation}
    \left\|f_\Theta(\bx_1)-f_\Theta(\bx_2) \right\|^2 \leq 
    \left(\prod_{\ell=1}^L\kappa^{(\ell)}\right) \left\| \bx_1-\bx_2 \right\|^2
\end{equation}
and thus its Lipschitz constant
\begin{equation}
    \kappa = \left(\prod_{\ell=1}^L\kappa^{(\ell)}\right), 
    \label{eq:lcdn}
\end{equation}
where the $\kappa^{(\ell)}$ are obtained from Proposition \ref{thm:lipschi} according to the DN's topology.
\end{proposition}

From Proposition \ref{thm:lipschi}, note the potentially large Lipschitz constants for all operators but the softmax; indeed it is the only contraction, in general.
These large constants can reinforce to produce a potentially very large overall Lipschitz constant for the complete DN in (\ref{eq:lcdn}).
There are several potential remedies: i) reduce the norms of the fully connected and convolution operators, ii) make the DN narrower by reducing $\D^{(\ell)}$, and iii) invent new, more stable activation and pooling operators.

\subsection{Collinear Templates and Data Set Memorization}\label{sec:colinear}

Under the matched filterbank interpretation of a DN developed in Section \ref{sec:tm}, the optimal template for an image $\bx$ of class $c$ is a scaled version of $\bx$ itself.
But what are the optimal templates for the other (incorrect) classes?  
In an idealized setting, we can answer this question; the proof is provided in Appendix \ref{sec:proof_colinear}. 

\begin{proposition}
\label{thm:colinear}
Consider an idealized DN consisting of a composition of MASOs that has sufficient approximation power to span arbitrary MASO matrices 
$A[\bx_n]$ from (\ref{eq:CNNaffine1a}) for any input $\bx_n$ from the training set.
Train the DN to classify among $C$ classes using the training data $\mathcal{D}=(\bx_n,y_n)_{n=1}^N$ with normalized inputs $\| \bx_n \|_2=1~\forall n$ and the cross-entropy loss $\mathcal{L}_{\rm CE}(y_n,f_\Theta(\bx_n))$ with the addition of the regularization constraint that $\sum_c\|A[\bx_n]_{c,\bigcdot }\|_2<\alpha$ with $\alpha>0$. 
%
At the global minimum of this constrained optimization problem, the rows of $A^\star[\bx_n]$ (the optimal templates) have the following form:
 \begin{equation}
     [A^\star[\bx_n]]_{c,\bigcdot}=\left\{ \begin{array}{ll}
         +\sqrt{\frac{(C-1)\alpha}{C}}\; \bx_n, \quad &c=y_n \\[1mm]
          -\sqrt{\frac{\alpha}{C(C-1)}}\; \bx_n,\quad &c\neq y_n
     \end{array}\right.
     \label{eq:collinear}
 \end{equation}
 \label{thm:optTemplate}
 \end{proposition}
\qq

In short, the idealized DN will {\em memorize a set of collinear templates}.

\begin{figure}[t]
\centering
\includegraphics[width=0.22\textwidth]{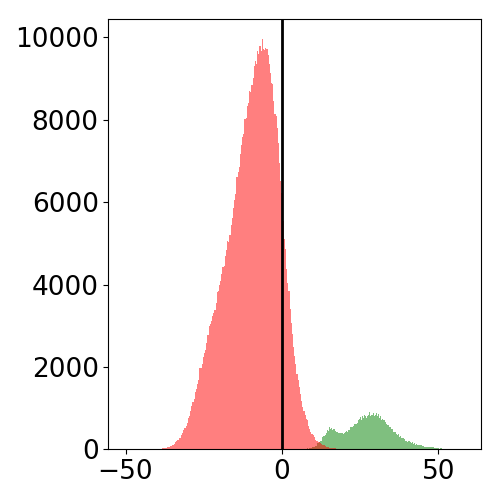}
\hspace{5mm}
\includegraphics[width=0.22\textwidth]{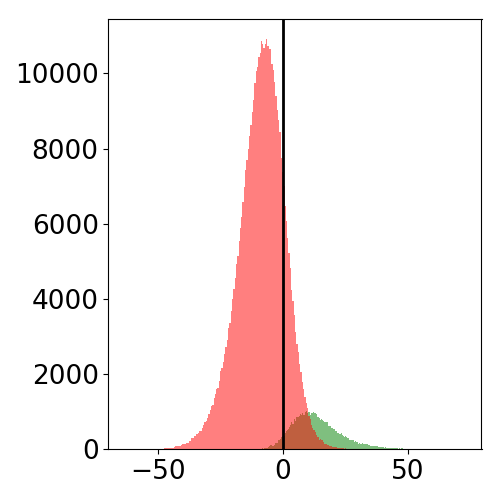}
\\
\hspace{5mm}
{\small (a) MNIST}
\hspace{25mm} 
{\small (b) CIFAR10} 
\\[-1mm]
\caption{
Empirical study of the implications of Proposition \ref{thm:optTemplate}, Part 1.  
We illustrate the bimodality of 
$\langle [A^\star[\bx_n]]_{c,\bigcdot},\bx_n \rangle$
for the {\em largeCNN} matched filterbank trained on (a) MNIST and (b) CIFAR10.
Training used batch normalization and bias units. Results are similar when trained with bias and no batch-normalization.
The green histogram summarizes the output values for the correct class (top half of (\ref{eq:collinear})), while the red histogram summarizes the output values for the incorrect classes (bottom half of (\ref{eq:collinear})). 
The easier the classification problem (MNIST), the more bimodal the distribution.
}
\label{fig:collinear}
\end{figure}

\begin{figure}[t]
\centering
\includegraphics[width=0.22\textwidth]{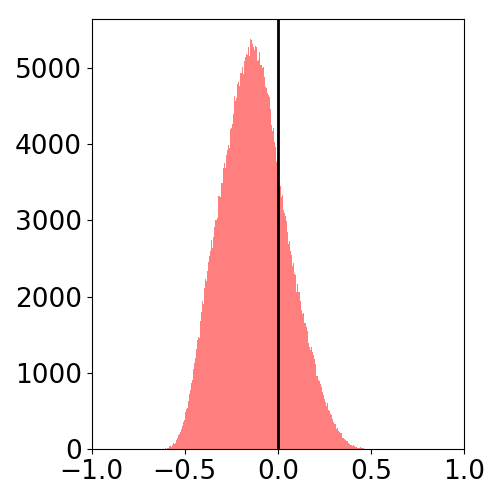}
\hspace{5mm}
\includegraphics[width=0.22\textwidth]{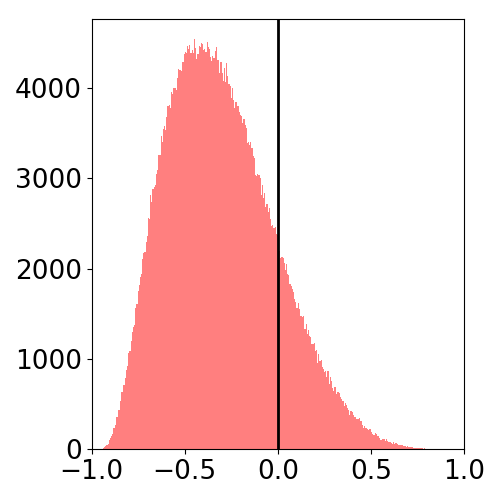}
\\
{\small (a) {\em smallCNN}} \hspace{23mm} 
{\small (b) {\em largeCNN}}
\\[-1mm]
\caption{
Empirical study of the implications of Proposition \ref{thm:optTemplate}, Part 2.  Training used batch normalization and bias units. Results are similar when trained with bias and no batch-normalization.
We plot the normalized inner products (cosine similarities) between each set of correct and incorrect class templates for (a)~{\em smallCNN} and (b)~{\em largeCNN} trained on MNIST.
Increasing the DN's capacity (from {\em smallCNN} to {\em largeCNN}) leads to templates that conform to (\ref{eq:collinear}) more closely. 
}
\label{fig:cosine}
\end{figure}

We report on an empirical study of the implications of this result in Figures~\ref{fig:collinear} and~\ref{fig:cosine}.
Figure~\ref{fig:collinear} confirms the bimodality of the inner products between the signal and the correct/incorrect templates on the MNIST and CIFAR10 datasets. 
The distance between the green and red histograms is directly related to the difficulty of the classification task.
Figure~\ref{fig:cosine} plots histograms of the normalized inner products (cosine similarities) between each set of correct and incorrect class templates
${\rm hist}\left\{ d\left(
[A[\bx_n]]_{y_n,\bigcdot},[A[\bx_n]]_{c,\bigcdot} \right)~\forall c \neq y_n~ \forall n\right\}$ 
with $d(u,v)=\frac{\langle u,v\rangle }{\|u \| \|v \|}$ for {\em smallCNN} and {\em largeCNN} on the MNIST dataset.
We clearly see that increasing the DN's capacity leads to templates that are more collinear and more the negative of the correct class template, again confirming (\ref{eq:collinear}).
In the limit with the idealized DN, (\ref{eq:collinear}) suggests that these histograms will concentrate around $-1$.

%% file: orthoTemplate.tex
\subsection{Boosting DN Performance via Orthogonal Templates}
\label{sec:orthoTemplate}

\begin{figure}[t]
\centering
{\small (a)} \includegraphics[width=0.45\textwidth]{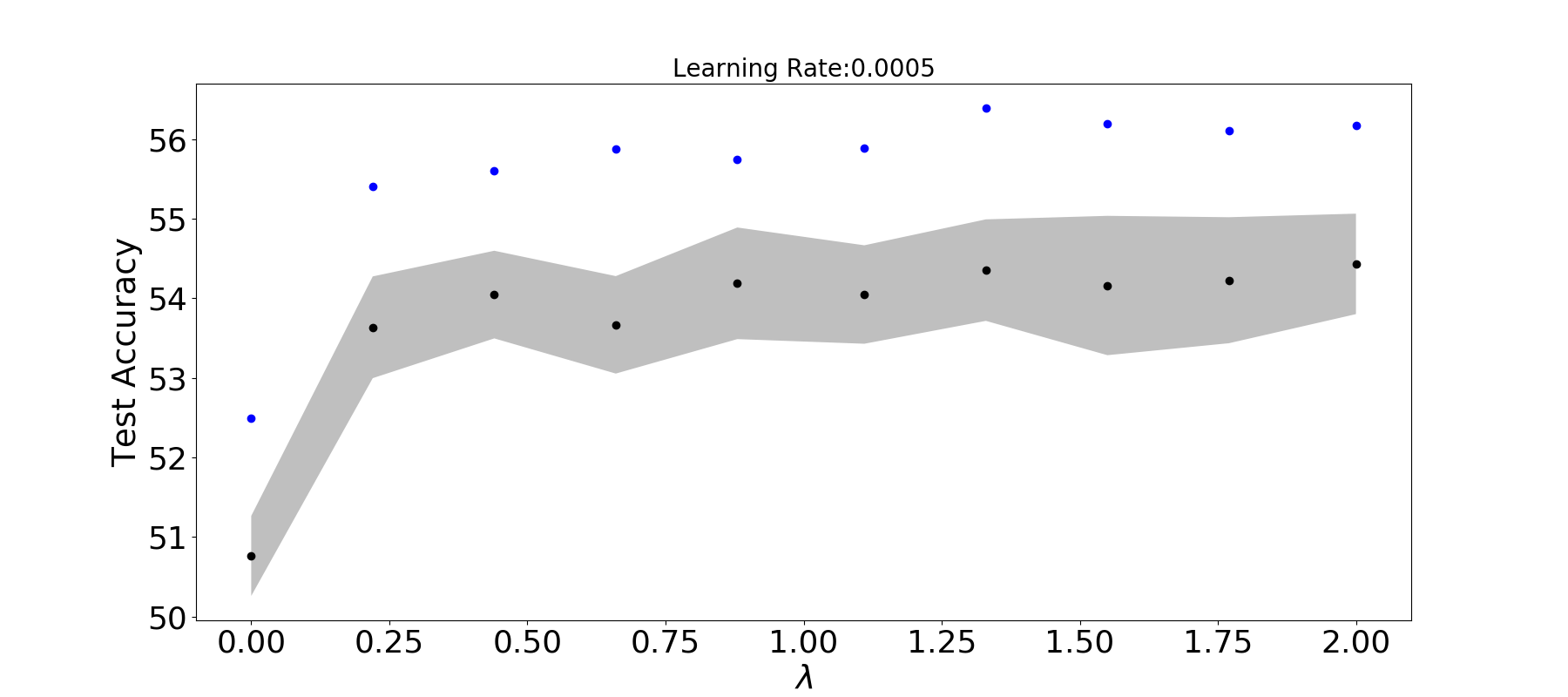}
{\small (b)} \includegraphics[width=0.45\textwidth]{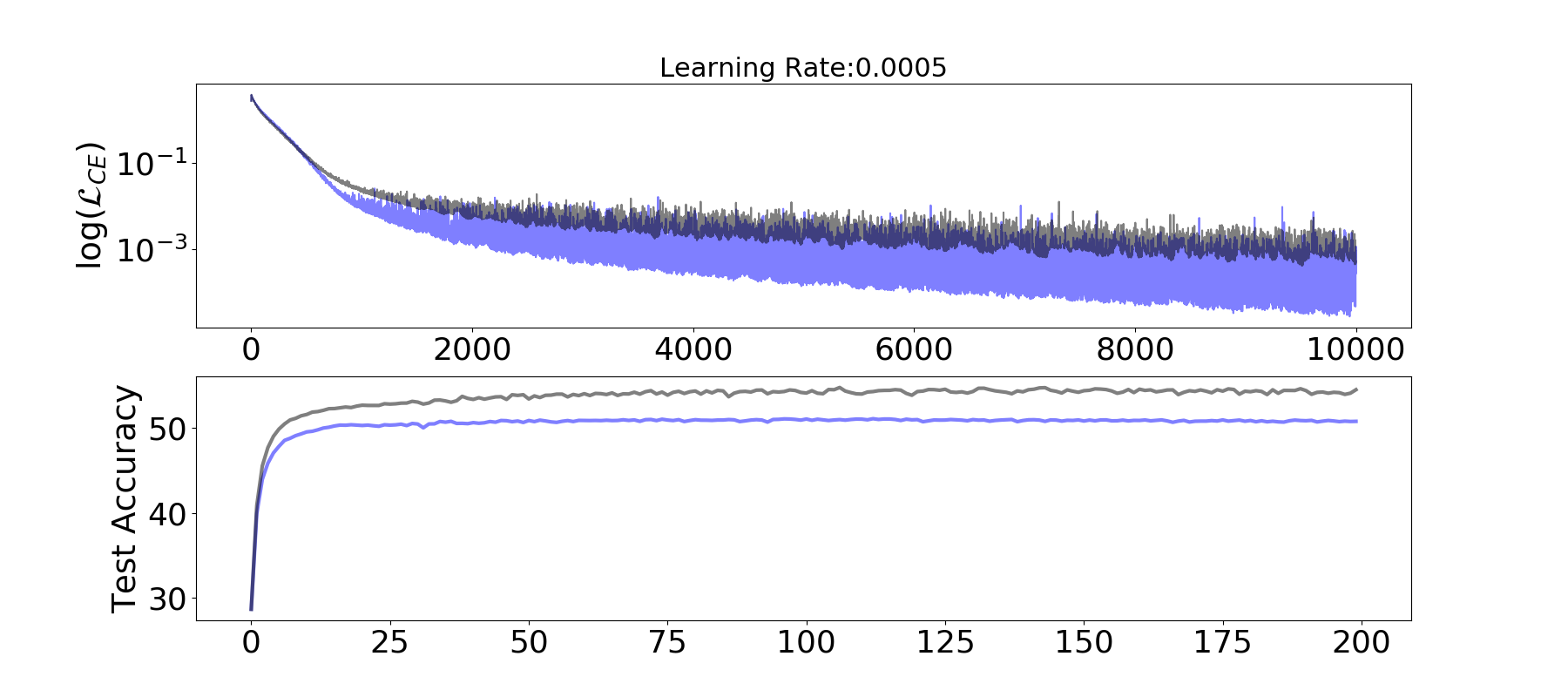}
\caption{
Orthogonal templates significantly boost DN performance with {\em no change to the architecture}.
(a) Classification performance of the {\em largeCNN} trained on CIFAR100 for different values of the orthogonality penalty $\lambda$ in (\ref{eq:orthog}).
We plot the average (back dots), standard deviation (gray shade), and maximum (blue dots) of the test set accuracy over 15 runs.
(b, top) Training set error. The blue/black curves corresponds to $\lambda=0/1$.
(b, bottom) Test set accuracy over the course of the learning.
}
\label{fig:cifar100}
\end{figure}

\begin{figure}[t]
\centering
{\small (a)}
\includegraphics[width=0.45\textwidth]{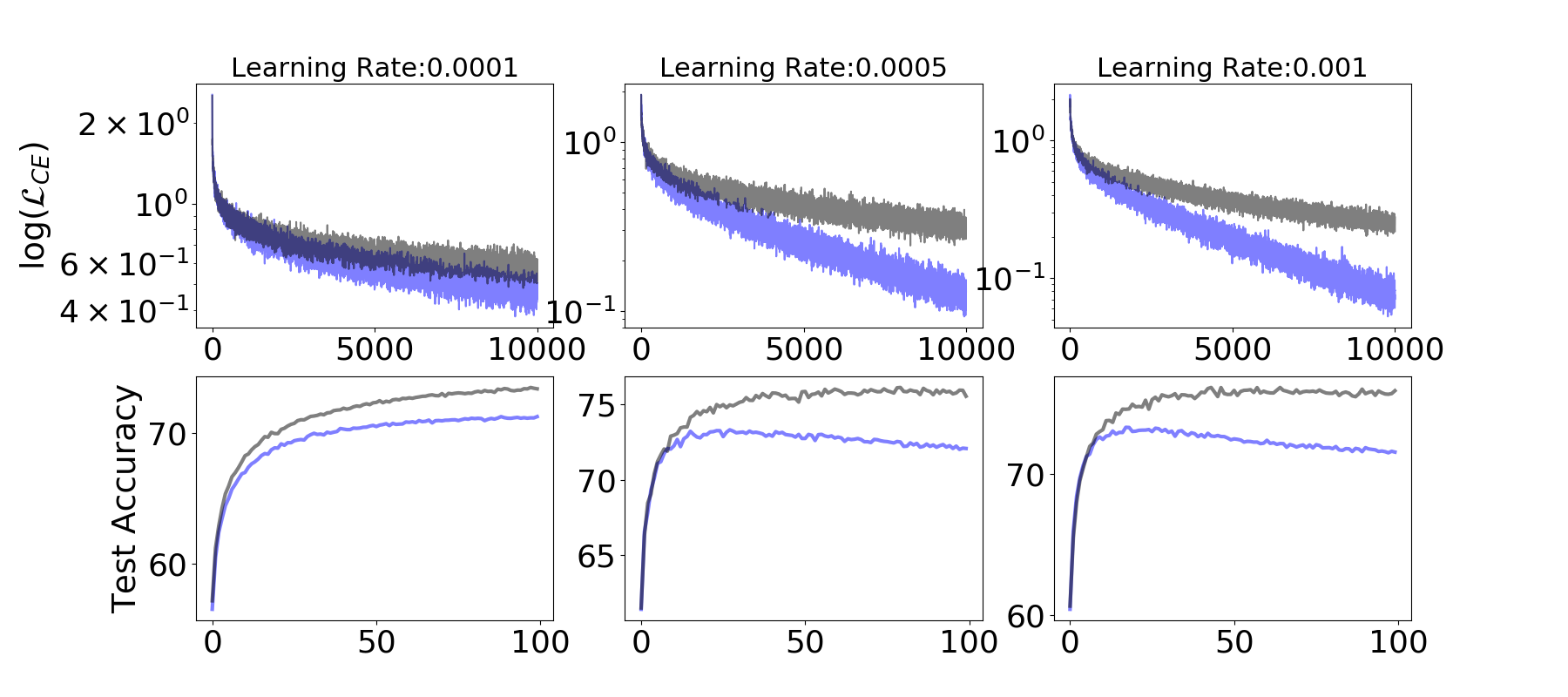}
{\small (b)}
\includegraphics[width=0.45\textwidth]{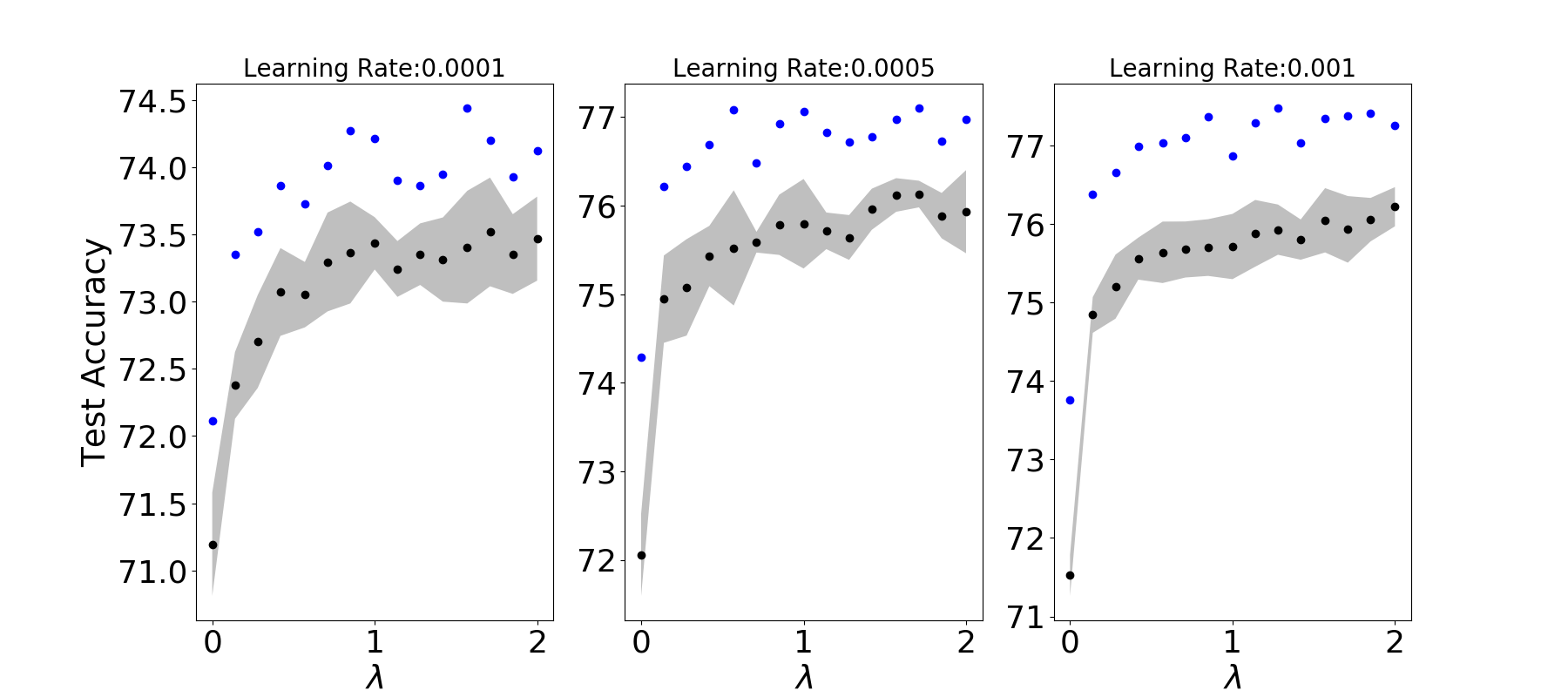}
\caption{
Orthogonal templates help to prevent overfitting with {\em smallCNN} trained on CIFAR10.
(a) In black, standard training and in blue training using the orthogonality penalty $\lambda=1$ in (\ref{eq:orthog}) averaged over 15 runs.
(b) Test set accuracy of the templates as we vary the orthogonality penalty $\lambda$ in (\ref{eq:orthog}). We plot the average, standard deviation, and maximum test set accuracy over 15 runs.  
}
\label{fig:cifarhisto}
\end{figure}

While a DN's signal-dependent matched filterbank (\ref{eq:CNNaffine1a}) is optimized for classifying signals immersed in additive white Gaussian noise, such a statistical model is overly simplistic for most machine learning problems of interest. 
In practice, errors will arise not just from noise but also from nuisance variations in the inputs, such as arbitrary object rotations and positions, occlusions, lighting conditions, etc.
The effects of these nuisances are only poorly approximated as Gaussian random noise.
Limited work has been done on filterbanks for classification in non-Gaussian noise; one promising direction involves using not matched but rather {\em orthogonal} templates \cite{eldar2001orthogonal}.

For a MASO DN's templates to be orthogonal for all inputs, it is necessary that the rows of the matrix $W^{(L)}$ in the final linear classifer layer be orthogonal.
This weak constraint on the DN still enables the earlier layers to create a high-performance, class-agnostic, featurized representation (recall the discussion just below (\ref{eq:CNNaffine1a})).
To create orthogonal templates during learning, we simply add to the standard (potentially regularized) cross-entropy loss function $\mathcal{L}_{\rm CE}$ from (\ref{eq:crossEnt}) a term that penalizes non-zero off-diagonal entries in the matrix $W^{(L)}(W^{(L)})^T$ leading to the new loss with extra penalty
\begin{align}
    \mathcal{L}_{\rm CE}+\lambda \sum_{c_1 \not = c_2}\left|\left\langle \big[W^{(L)}\big]_{c_1,\bigcdot}, \big[W^{(L)}\big]_{c_2,\bigcdot}\right\rangle \right|^2.
    \label{eq:orthog}
\end{align}
The parameter $\lambda$ controls the tradeoff between cross-entropy minimization and orthogonality preservation.
Note that there is no change to the DN's architecture, only to its cost function for learning.
Conveniently, when minimizing (\ref{eq:orthog}) via backpropagation, the orthogonal rows of $W^{(L)}$ induce orthogonal backpropagation updates for the various classes.

We now empirically demonstrate that orthogonal templates lead to significantly improved classification performance.
We conducted a range of experiments with three conventional DN architectures -- {\em smallCNN}, {\em largeCNN}, and {\em ResNet4-4} -- trained on CIFAR10 and CIFAR100.
See Appendix \ref{ap:dntopology} for the network architecture details plus additional results for the SVHN dataset.
Each DN employed ReLU activations, max-pooling, bias, and batch normalization prior to each ReLU. 
For learning, we used the Adam optimizer with an exponential learning rate decay. 
All inputs were centered to zero mean and scaled to a maximum value of one. 
No further pre-processing was performed, such as ZCA whitening \cite{nam2014local}.
We assessed how the classification performance of the DN changed as we varied the orthogonality penalty $\lambda$ in (\ref{eq:orthog}).
For each configuration of DN architecture, training dataset, learning rate, and penalty $\lambda$, we averaged over $15$ runs to estimate the average performance and standard deviation.

The results for the {\em largeCNN} on CIFAR100 in Figure~\ref{fig:cifar100} indicate that the benefits of the orthogonality penalty emerge distinctly as soon as 
$\lambda>0$.

We delve deeper into the performance of the {\em smallCNN} on CIFAR10 in Figure~\ref{fig:cifarhisto}. 
In addition to improved final accuracy and generalization performance, we see that template orthogonality reduces the tendency of the DN to overfit. 
One explanation is that the orthogonal weights $W^{(L)}$ positively impact not only the prediction but also the backpropagation via orthogonal gradient updates.
A range of additional experiments are provided in Appendix \ref{ap:moreortho}.

%% file: partition.tex
\section{
Multiscale Spline Partition Induced by a DN}
\label{sec:partition}

Like any spline, it is the interplay between the (affine) spline mappings and the input space partition that work the magic in a MASO DN.
Recall from Section \ref{sec:splines} that a MASO has the attractive property that it implicitly partitions the input signal space $\R^D$ as a function of its slope and offset parameters.
The induced partition $\Omega$ opens up a new geometric avenue to study how a MASO-based DN clusters and organizes signals in a hierarchical fashion.
In particular, we will see that there are direct links with {\em vector quantization} (VQ) and {\em $K$-means clustering}.\footnote{Do not confuse the $K$ of $K$-means with the MASO dimension.}

\subsection{Local MASO Partition}
\label{sec:MASOpartition}

Level $\ell$ of a MASO DN directly partitions its input space $\R^{D^{(\ell-1)}}$ and thus indirectly influences the partitioning of the overall input signal space $\R^D$ (recall that $D^{(0)}=D$).
We begin by developing a formula for the input space partition $\omega^{(\ell)}[k,r]$ in $\R^{D^{(\ell-1)}}$ that is effected by the $k^{\rm th}$ of the $K=D^{(\ell)}$ affine spline functions comprising the MASO.
Partitions at each level $\ell$ are hierarchically chained together to effect the global partition; we discuss this in more detail below.

We illustrate first with the partitions effected by the ReLU activation and max-pooling operators.
A ReLU operator at level $\ell$ and input/output $k$ splits its layer input space into two half-planes depending on the sign of the input.
We denote by $\omega^{(\ell)}[k,r]$ the $r^{\rm th}$ region at output $k$ and level $\ell$; since $R=2$ for a ReLU spline, we have
\begin{align}
\omega^{(\ell)}[k,1]
&=\left\{\bz^{(\ell-1)}(\bx) \in \mathbb{R}^{D^{(\ell)}}: \big[\bz^{(\ell-1)}(\bx)\big]_k<0\right\}, 
\\[2mm]
\omega^{(\ell)}[k,2]
&=\left\{\bz^{(\ell-1)}(\bx) \in \mathbb{R}^{D^{(\ell)}}: \big[\bz^{(\ell-1)}(\bx)\big]_k\geq 0\right\}.
\label{eq:regionrelu}
\end{align}
As detailed in Appendix~\ref{sec:sm_spline_op}, the complete MASO partition of the ReLU activation operator's input space $\R^{D^{(\ell-1)}}$ is the intersection of the half-spaces of each of the $K$ input/output pairs.
Preceding an activation operator like ReLU with a fully connected or convolution operator simply rotates the half-space partition in $\R^{D^{(\ell-1)}}$; addition of a bias simply translates the origin.
The number of partition regions is combinatorially large -- up to $2^{D^{(\ell)}}$ for ReLU and up to $R^{D^{(\ell)}}$ for an activation function with $R\geq 2$.
These partition regions can be of infinite volume, but they are always convex.

Consider next the partition effected by max-pooling.
A max-pooling operator at layer $\ell$ also partitions its layer input space $\R^{D^{(\ell-1)}}$ into a combinatorially large number of regions (up to $(\#R^{(\ell)})^{D^{(\ell)}}$, where $\#R^{(\ell)}$ is the size of the pooling region).
Each partition region will depend on the location of the $\argmax$ entry for each output
%
%
\begin{equation}
\omega^{(\ell)}[k,r]=\Big\{\bz^{(\ell-1)}(\bx) \in \R^{D^{(\ell)}} : 
r=\argmax_{d\in \R_k^{(\ell)}}
\big[\bz^{(\ell-1)}(\bx)\big]_{d}\Big\}
\label{eq:regionmax}
\end{equation}
for $r=1,\dots,\#R^{(\ell)}$.

More generally, to find the partition of the layer input space effected by a layer-$\ell$ MASO constructed by concatenating $K=D^{(\ell)}$ identical max-affine spline functions,\footnote
{
See Appendix \ref{sec:sm_spline_op} for the general case of concatenating $K$ non-identical splines.
}
we first rewrite the layer calculation (\ref{eq:MASO}) as 
\begin{align}
\big[\bz^{(\ell)} (\bx)\big]_k
&= \max_{r=1,\dots,R^{(\ell)}} \left\langle \big[\bA^{(\ell)}\big]_{k,r,\bigcdot},\bz^{(\ell-1)}(\bx)\right\rangle +\big[\bB^{(\ell)}\big]_{k,r}
\label{eq:originalmaso} 
\\
& =\sum_{r=1}^R \big[T^{(\ell)}
(\bz^{(\ell)} (\bx))\big]_{k,r} 
\left( \left\langle \big[\bA^{(\ell)}\big]_{k,r,\bigcdot},\bz^{(\ell-1)}(\bx)\right\rangle +\big[\bB^{(\ell)}\big]_{k,r} \right).
    \label{eq:tmaso}
\end{align}
Here,
\begin{align}
&\big[T^{(\ell)}
(\bz^{(\ell-1)} (\bx))\big]_{k,r} 
=\Indic
\Big(r=\argmax_{r=1,\dots,R}  \left\langle [\bA^{(\ell)}]_{k,r,\bigcdot},\bz^{(\ell-1)}(\bx)\right\rangle
+[\bB^{(\ell)}]_{k,r}\Big)
\label{eq:tmaso2}
\end{align}
is a {\em selection matrix} whose $k^{\rm th}$ row comprises a 1-hot selection vector (Kronecker delta function) that selects the partition region $r$ that maximizes (\ref{eq:tmaso}) in the $k^{\rm th}$ output dimension. 
We denote the corresponding index by $\big[t^{(\ell)}\big]_k$.
Then, we can write
\begin{equation}
\omega^{(\ell)}[k,r]= 
\Big\{ \bz^{(\ell-1)}(\bx) \in \mathbb{R}^{D^{(\ell)}} : 
\big[T^{(\ell)}
(\bz^{(\ell-1)} (\bx))\big]_{k,r}=1 \Big\}.
\label{eq:regiongeneral}
\end{equation}

When computing the MASO output via the max operator as in (\ref{eq:originalmaso}), we do not have explicit access to the selection matrix $T^{(\ell)}$ in (\ref{eq:tmaso2}).
However, there is a simple strategy to compute it a posteriori.

\begin{proposition}
\label{thm:findt}
For any MASO, the selection matrix $T^{(\ell)}$ in (\ref{eq:tmaso2}) can be computed via
\begin{equation}
    \left[T^{(\ell)}
    (\bz^{(\ell-1)}(\bx))\right]_{k,r}
    =
    \frac{\partial \max_r \left\langle \big[\bA^{(\ell)}\big]_{k,r,\bigcdot},\bx \right\rangle
    +
    \big[\bB^{(\ell)}\big]_{k,r}}{\partial \left(\left\langle \big[\bA^{(\ell)}\big]_{k,r,\bigcdot},\bx \right\rangle
    +
    \big[\bB^{(\ell)}\big]_{k,r}\right)}.
    \label{eq:tmasocomputation}
\end{equation}
\end{proposition}

The derivative in (\ref{eq:tmasocomputation}) can be simply and efficiently computed via backpropagation on the DN.


It is easy to show (see Appendix \ref{sec:sm_spline_op}) that the overall partitioning of the layer-$\ell$ MASO input space $\R^{D^{(\ell-1)}}$ is simply the intersection of the partitions effected by each of the $k=1,\dots,D^{(\ell)}$ output dimensions
\begin{equation}
    \omega^{(\ell)}[r_1,\dots,r_K] = \bigcap_{k=1}^{D^{(\ell)}}\omega^{(\ell)}[k,r_k], 
\label{eq:factorial}
\end{equation}
where 
$r_k \in \left\{1,\dots,R^{(\ell)}\right\}$.
The final partition contains up to $(R^{(\ell)})^{D^{(\ell)}}$ convex and connected regions; 
in practice, however, many of them could have zero volume.
(Our simple statistical study in Section~\ref{sec:simplestats} bears out this conclusion.)
We leave it as an exercise for the reader to show that the intersection (\ref{eq:factorial}) tessellates the entire layer input space $\R^{D^{(\ell)}}$.

\subsection{Local Partition in the Input Signal Space}
\label{sec:MASOpartitionIn}

We can map the partition of the layer-$\ell$ MASO's input space $\R^{D^{(\ell-1)}}$ back to a corresponding partition of the input signal space $\R^D$. 
We do this by pulling the partition (\ref{eq:factorial}) back from layer $\ell$ to the input via 
%
%
\begin{equation}
    \omega_{\rm in}^{(\ell)}[r_1,\dots,r_{D^{(\ell)}}] = \bigcap_{k=1}^{D^{(\ell)}}\omega_{\rm in}^{(\ell)}[k,r_k], 
\label{eq:factorialG}
\end{equation}
where
\begin{align}
\omega_{\rm in}^{(\ell)}[k,r]=\left\{\bx \in \mathbb{R}^{D}: \left[T^{(\ell)}(\bz^{(\ell-1)}(\bx))\right]_{k,r}=1\right\},
\label{eq:globalregiongeneral}
\end{align}
and, as in (\ref{eq:regiongeneral}) and (\ref{eq:factorial}), $k$ indexes the $k^{\rm th}$ output of the $\ell^{\rm th}$ layer
and $r, r_k \in \left\{
1,\dots,R^{D^{\ell}} \right\}$.
This partition can be interpreted as a VQ of the training data that has been featurized by 
layers 1 through $\ell-1$.
Interestingly, the total number of possible regions remains $R^{D^{\ell}}$.
However, the regions are no longer necessarily convex nor connected, due to the nonlinear transformations effected by layers 1 through $\ell-1$. 
Moreover, the spline function defined in each region is no longer a fixed affine function; rather it is a piecewise affine function, in general. 
(We rectify this latter situation in the next subsection.)

We have so far been stymied in our quest for a closed-form formula for the regions (\ref{eq:factorialG}) for an arbitrary composition of MASOs. 
However, (\ref{eq:globalregiongeneral}) enables a simple strategy to estimate the partition induced by the DN layers $1,\dots,\ell$.
The algorithm runs as follows:
First, sample the overall input signal space $\R^D$ to create a collection of inputs $\left\{\bx'\right\}$.
Second, pass each $\bx'$ through the DN until the desired level $\ell$ to compute the layer outputs 
$\bz^{(\ell')}(\bx')$
as well as the selection matrices $T^{(\ell')}(\bz^{(\ell'-1)}(\bx')$ via (\ref{eq:tmasocomputation}) for $\ell'=1,\dots,\ell$.
Third, label each $\bx'$ by the region that is selected by  $\big[T^{(\ell)}(\bz^{(\ell-1)}(\bx'))\big]_{k,r}$ based on (\ref{eq:globalregiongeneral}).
The finer the sampling of the input space, the more accurate the estimate of the partition region boundaries.

\subsection{Global DN Partition}
\label{sec:DNpartition}

In the previous subsection, we found the partition of the input signal space $\R^D$ effected by (only) the layer-$\ell$ MASO. 
Intersecting the partitions of the input signal space induced by all $\ell'$ layers between the input and the output of layer $\ell$ yields the global partition generated by the DN up to layer $\ell$
\begin{equation}
    \omega_{\rm g}^{(\ell)}\!
    \left[
    \big\{ 
    r^{(\ell')}_1,\dots,
    r^{(\ell')}_{D^{(\ell')}} 
    \big\}_{\ell'=1}^{\ell}
    \right] =
    \bigcap_{\ell'=1}^{\ell}
    \omega_{\rm in}^{(\ell')}\!
    \left[
    r^{(\ell')}_1,\dots,r^{(\ell')}_{D^{(\ell')}}
    \right].
\label{eq:factorialG1}
\end{equation}
This global partition is precisely the partition $\Omega^S$ of the ASO referred to in Theorem~\ref{thm:big2}; it takes into account the joint configuration of the selection matrices $T^{(\ell')}$ for $\ell'=1,\dots,\ell$.


\subsection{Example: Partition Visualization and Statistics}
\label{sec:simplestats}

While visualizing the partition of a high-dimensional DN is impossible, a miniature case study provides some useful intuition.
Consider an old-school neural network with input signal dimension $D=2$ and $L=3$ layers comprised of the composition of three fully connected operators interspersed with two activation operators plus a final softmax operator.
We consider three different activation functions:  ReLu, leaky ReLU, and absolute value.
The dimensionalities of the layers are:  $D^{(0)}=2$, $D^{(1)}=45$, $D^{(2)}=3$, $D^{(3)}=4$.
We seek to solve a $C=D^{(3)}=4$-class classification problem.
See Appendix \ref{ap:dntopology} for a diagram.
We trained the DN using biases and batch normalization on a training set of 5000 data points from each of the four classes using the Adam optimizer \cite{kingma2014adam} and reached 98\% accuracy on the training set. 
The use of batch normalization in this simple setting does not impact the results substantively other than to shift the region boundaries to be more centered around the clusters of training data points.

\begin{figure}[!t]
    \centering
\begin{minipage}[c]{5mm}
\begin{turn}{90}
{\small $\omega_{\rm g}^{(2)}[k,r]$} \hspace*{23mm}
{\small $\omega_{\rm in}^{(2)}[k,r]$} \hspace*{23mm}
{\small $\omega_{\rm in}^{(1)}[k,r]$}
\hspace*{3mm}
\end{turn}
\end{minipage}    
~~
\begin{minipage}[c]{0.78\textwidth}
    \centering
    {\small Training time} $\longrightarrow$ \\
\includegraphics[width=\textwidth]{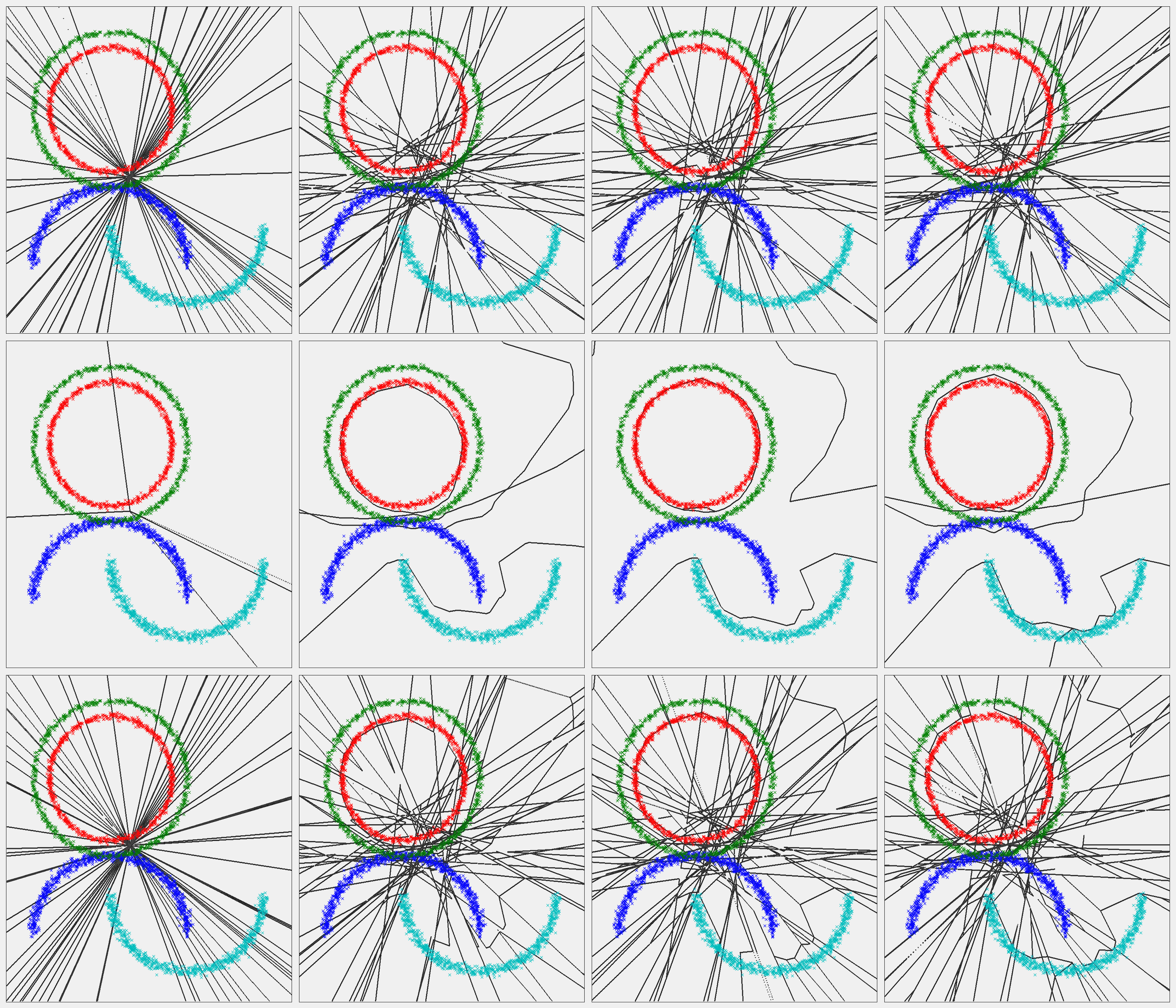}
\end{minipage}   
    \caption{
Illustration of the input signal space partition of a simple three-layer DN using ReLU activation and batch normalization.  
All biases were initialized to zero.
(See the text and Appendix \ref{ap:dntopology} for the network details.)
The top row displays the input signal space $\R^2$ with the training data points from the $C=4$ classes (colored green, red, blue, and cyan) overlaid with the partition 
$\omega^{(1)}[k,r]=
\omega_{\rm in}^{(1)}[k,r]=
\omega_{\rm g}^{(1)}[k,r]$ 
effected by the $D^{(1)}=45$ outputs of the first layer at four different time epochs in the training (initialization, 40, 80, 120 epochs).
The middle row displays the local partition in the input space 
$\omega_{\rm in}^{(2)}[k,r]$ 
effected by the $D^{(2)}=3$ outputs of the second layer.
It is interesting to note the region at the bottom of the green and red circles that contains points from both classes.
Correct classification remains possible in this region, however, thanks to the affine function applied in that region.
The bottom row displays the global partition in the input space 
$\omega_{\rm g}^{(2)}[k,r]$ 
effected by the first and second layers.
%
Note the nonconvexity of the regions
$\omega_{\rm in}^{(2)}[k,r]$ and $\omega_{\rm g}^{(2)}[k,r]$. 
}
\label{fig:simplepartitions}
\end{figure}

\begin{figure}[!t]
    \centering
\includegraphics[width=0.60\textwidth]{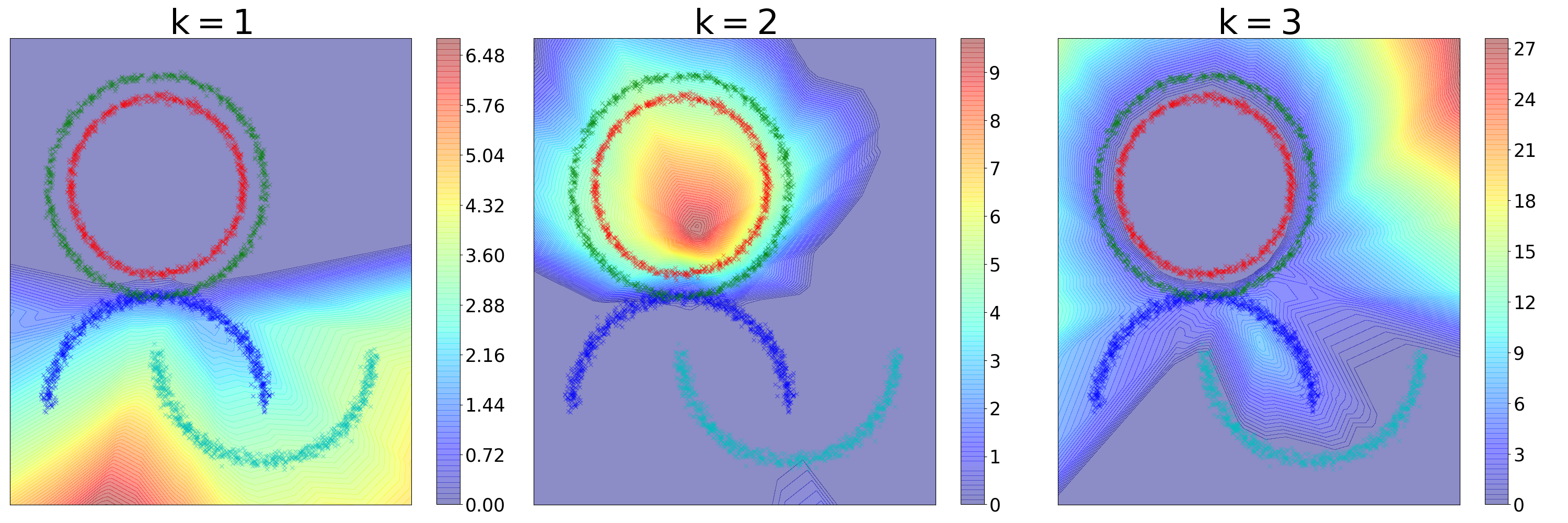}
\\
{\small (a) ReLU}
\\[1mm]
\includegraphics[width=0.60\textwidth]{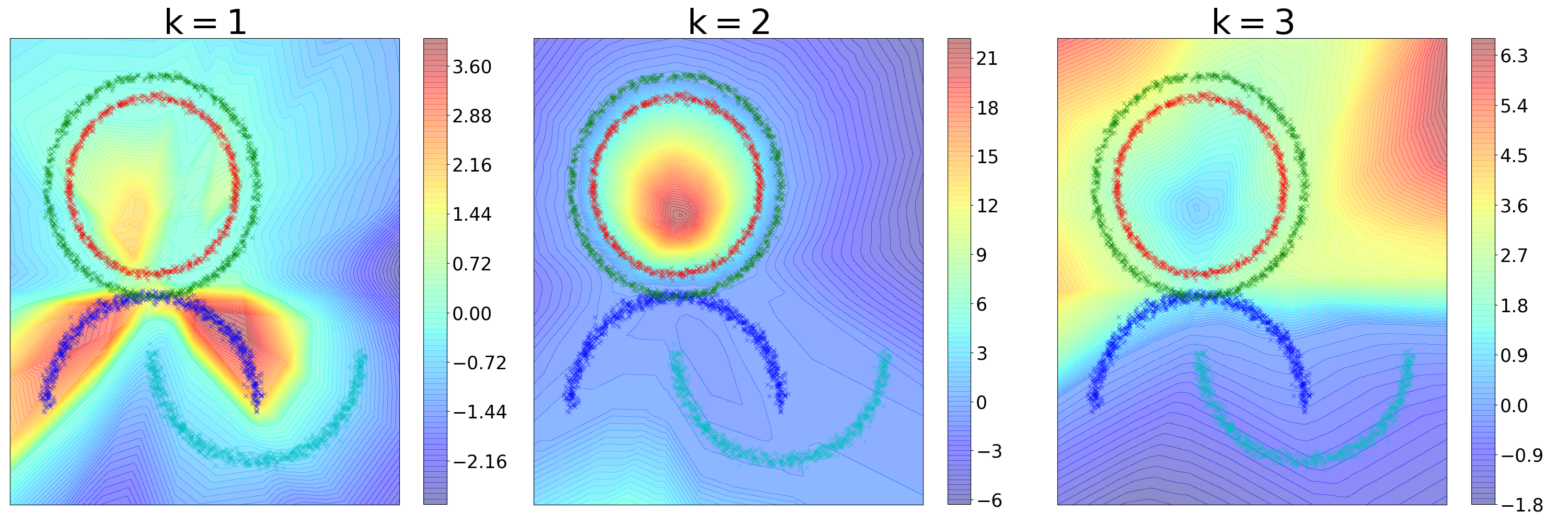}
\\
{\small (b) Leaky ReLU}
\\[1mm]
\includegraphics[width=0.60\textwidth]{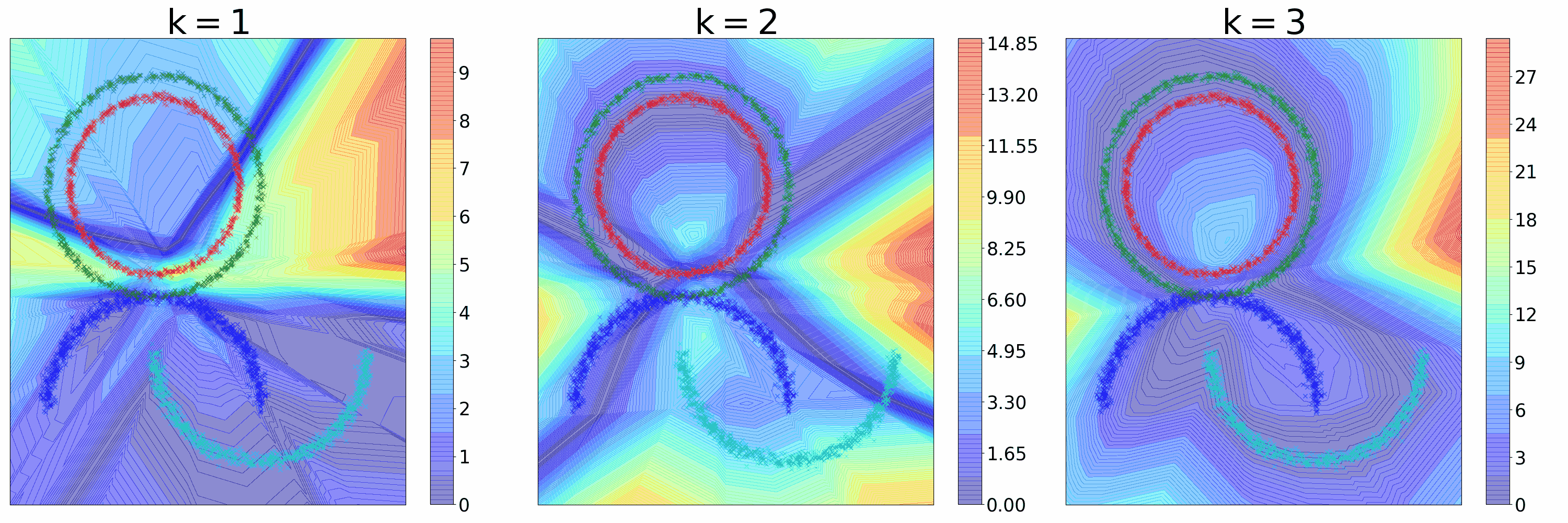}
\\
{\small (c) Absolute value}
%
    \caption{
Illustration of the learned piecewise-affine function mapping of the DN from Figure~\ref{fig:simplepartitions} from $\bx$ in the input signal space $\R^2$ to $[\bz^{(2)}(\bx)]_k$ at the output of the second layer $D^{(2)}=3$ overlaid on the training data points from the $C=4$ classes.
Each row corresponds to a different activation function.
The three columns correspond to the output dimensions $k=1,2,3$.
It is interesting to confirm visually how the function mappings in (a) are a fixed affine function on each partition region in the bottom right image in Figure~\ref{fig:simplepartitions}. Due to the high number of global DN regions (onto which the DN mapping is affine) the overall mapping displayed here can not be visually seen as being piecewise affine.
%
}
\label{fig:insimplepartitions}
\end{figure}

Figure~\ref{fig:simplepartitions} displays the input signal space $\R^2$ with the training data points overlaid with the partitions 
$\omega_{\rm in}^{(1)}[k,r]$,\footnote{
Recall that in the first layer 
$\omega_{\rm g}^{(1)} = \omega_{\rm in}^{(1)} = 
\omega^{(1)}$.}
$\omega_{\rm in}^{(2)}[k,r]$,
and 
$\omega_{\rm g}^{(2)}[k,r]$
as defined in
(\ref{eq:factorial}),
(\ref{eq:factorialG}),
and 
(\ref{eq:factorialG1}), respectively.
Figure~\ref{fig:insimplepartitions} displays the input signal space $\R^2$ with the training data points overlaid by the DN's function approximation for the three different activation functions.
We can make some interesting observations.
First, even for such a simple network, the learned partition regions are quite complex.
For instance, the regions induced by the second layer are neither convex nor connected.
Second, there exist partition regions that contain training data points from more than one class. 
Correct classification remains possible, however, thanks to the affine function 
applied within each region.
Third, thanks to the complexity of the partition regions, even this simple affine spline can realize quite complicated functional mappings.
Fourth, the partitions and functional mappings learned for each different activation function are very different.

We now provide some statistics regarding the partition regions induced by the first two layers of the toy DN.
Table~\ref{tab:null} 
tabulates the number of non-trivial partition regions (those with nonzero volume) of the toy DN averaged over 10 runs. 
We see that most of exponentially many possible regions have zero volume and are trivial. We also provide the same statistics with the leaky ReLU and absolute value nonlinearities in Appendix \ref{sec:extratoy}. 

\begin{figure}[!tb]
    \centering
\begin{minipage}[c]{2mm}
\begin{turn}{90}
{\footnotesize num.\ of points}
\end{turn}
\end{minipage}    
\begin{minipage}[c]{0.60\textwidth}
        \includegraphics[width=\linewidth]{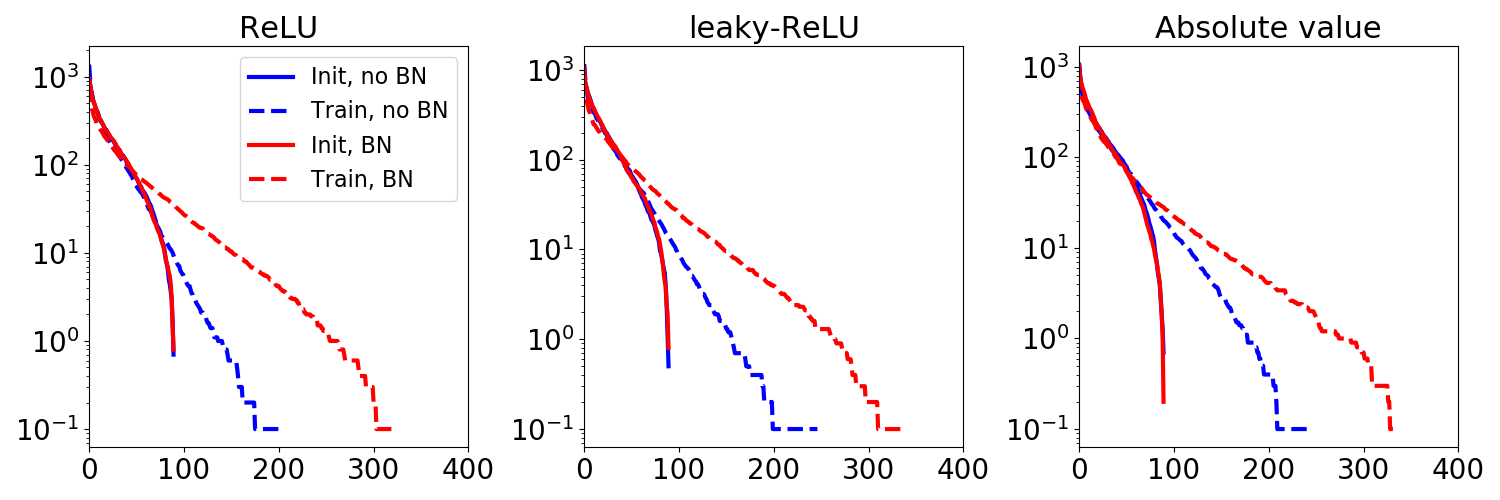}
\end{minipage}
        \\
 {\footnotesize sorted regions} 
 \\[1mm]
{\small (a) Occupancy of 
$\omega_{\rm in}^{(1)}[k,r]$}
        \\[2mm]
\begin{minipage}[c]{2mm}
\begin{turn}{90}
{\footnotesize num.\ of points}
\end{turn}
\end{minipage}    
\begin{minipage}[c]{0.60\textwidth}
        \includegraphics[width=\linewidth]{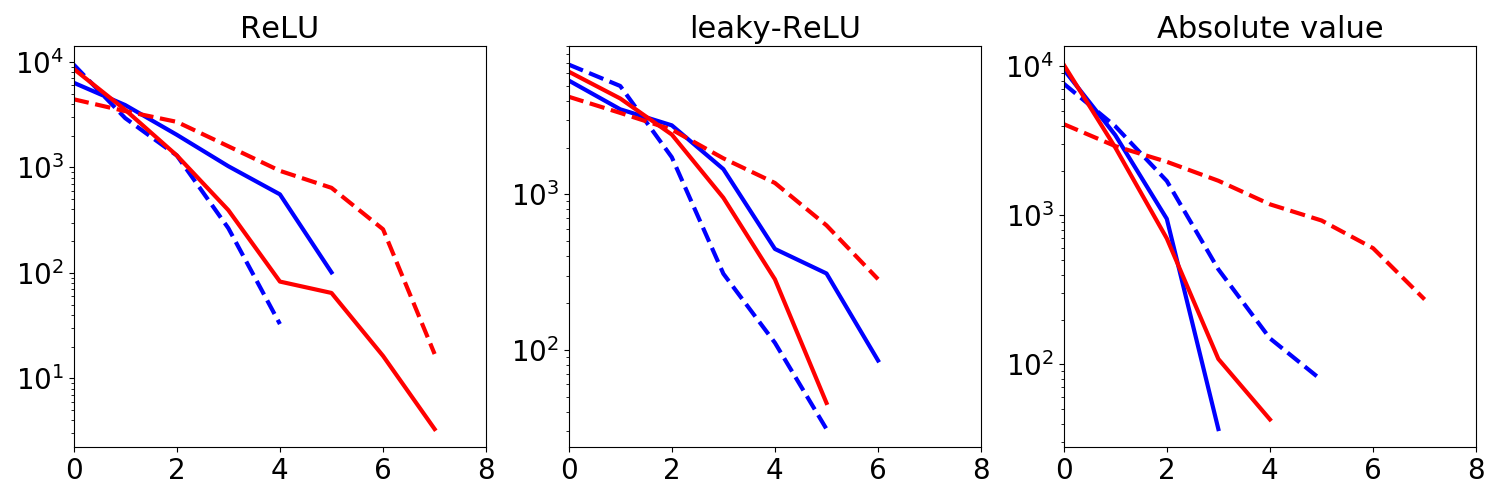}
\end{minipage}
        \\
 {\footnotesize sorted regions} 
 \\[1mm]
{\small (b) Occupancy of 
$\omega_{\rm in}^{(2)}[k,r]$}
\caption{
Occupancy of the partition regions induced in the input signal space by (a) $\omega_{\rm in}^{(1)}$ from layer 1 and 
(b) $\omega_{\rm in}^{(2)}$ from layers 1 and 2 of the DN from Figure~\ref{fig:simplepartitions}.
We sorted the regions according to how many test data points each contains and plot on a log scale the results for the ReLU, leaky ReLU, and absolute value activation functions.
The four curves on each plot correspond to the occupancy at (random) initialization and after training and with and without batch normalization.
We averaged the results over 10 runs.
A curve touching the horizontal axis indicates that all regions with higher sorted index are empty.
}
\label{fig:simplestatistics}
\end{figure}

Interestingly, many of the non-trivial partition regions are actually empty (i.e., devoid of any training data points).  
Table \ref{tab:occupancy} reports the statistics, and
Figure~\ref{fig:simplestatistics} plots 
the number of training data points per region in descending order, where only the nonempty regions are displayed for clarity.
We see that batch normalization yields a partition where the data points are more evenly distributed across the regions.

Keep in mind that the above results are only for a very small, toy DN.
An interesting research direction is extending this analysis to the kinds of large DNs used in practice.

\begin{table}
    \centering
    \small
    Number of Nonzero Volume Regions \\[2mm]
    \begin{tabular}{c|cccc} 
    Partition & \multicolumn{2}{c}{Initialization} & \multicolumn{2}{c}{Trained} \\
     & (\st{BN}) & (BN) & (\st{BN}) & (BN) \\ \hline
    $\omega_{\rm in}^{(1)}$    
    & 90 & 90 & 573 & 808 \\ \hline
    $\omega_{\rm in}^{(2)}$       
    & 6 & 7 & 6 & 8\\ \hline
    $\omega_{\rm g}^{(2)}$ &99&100&442&1032
    \end{tabular}
    \caption{
    Number of nontrivial partition regions having nonzero volume for the input signal space partition induced by 
    $\omega_{\rm in}^{(1)}$ from layer 1 and 
    $\omega_{\rm in}^{(2)}$ from  2 of the toy DN from Figure~\ref{fig:simplepartitions}.
    The results with and without batch normalization are similar: most of the exponentially many possible regions are trivial when compared to the number of possible regions. However, the use of BN allow more region to be effectively nontrivial for both layers.
    }
    \label{tab:null}
    \end{table}
    
    \begin{table}[t]
    \centering
    \small
    Number of Nonempty Regions \\[2mm]
    \begin{tabular}{c|cccc} 
     & \multicolumn{2}{c}{Initialization} & \multicolumn{2}{c}{Trained} \\
    Partition & (\st{BN}) & (BN) & (\st{BN}) & (BN) \\ \hline
    $\omega_{\rm in}^{(1)}$    
    & 90 & 90 & 205 & 305 \\ \hline
    $\omega_{\rm in}^{(2)}$
    & 6 & 7 & 5 & 8 \\ \hline
    $\omega_{\rm g}^{(2)}$ &98&98&237&389
    \end{tabular}
    \caption{
        Number of partition regions that are occupied by at least one training data point for the toy DN from Figure~\ref{fig:simplepartitions}. As for the number of nontrivial region, the use of BN seems to allow more regions to be non empty for both layers.
    }
    \label{tab:occupancy}
\end{table}

\subsection{Connection to Vector Quantization and $K$-Means}
\label{sec:vq}

\begin{figure}[t]
    \centering
    {\small (a)} \includegraphics[width=0.22\linewidth]{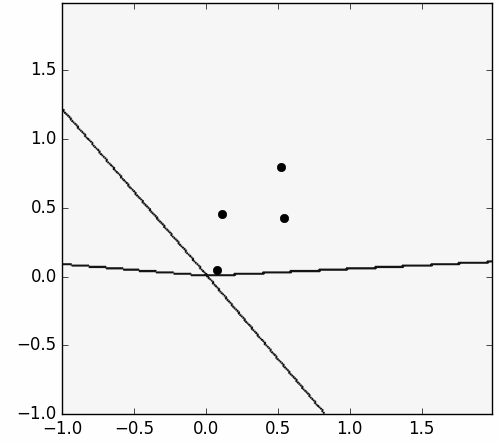}
    \hspace*{10mm}
    {\small (b)} \includegraphics[width=0.22\linewidth]{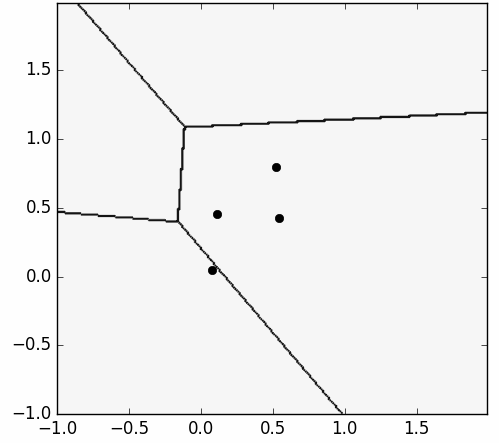}
    \hspace*{10mm}
    {\small (c)} \includegraphics[width=0.22\linewidth]{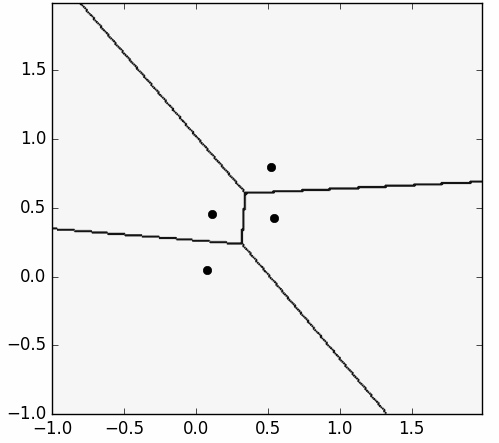}
    \caption{
    Simple example of the correspondence between VQ/$R$-means and a MASO partition at level $\ell$.
    We illustrate three different partitions of the $D^{(\ell-1)}=2$-dimensional input space via a MASO with $R^{(\ell)}=4$ regions induced by the same slope parameters
    $\big[A^{(\ell)}\big]_{k,r,\bigcdot}$ but three different biases:
    (a) 
    $\big[B^{(\ell)}\big]_{k,r}=0$, 
    (b) 
    random 
    $\big[B^{(\ell)}\big]_{k,r}$,
    and 
    (c) 
    $\big[B^{(\ell)}\big]_{k,r}=-\frac{1}{2} \big\|\big[A^{(\ell)}\big]_{k,r,\bigcdot}\big\|^2$.
    In (c) the slope parameters correspond to $K$-means centroids
    as in (\ref{eq:kmeans1}), and the partition is the Voronoi tiling based on (\ref{eq:kmeans2}).
    }
    \label{fig:VQbias}
\end{figure}

The MASO partition is equivalent to a {\em vector quantization (VQ)} of its input space \cite{gersho2012vector}.
Indeed, the connection between max-affine splines and $K$-means (a particular flavor of VQ) has been pointed out in \cite{magnani2009convex} in the least-squares regression setting.
VQ is a classical technique for lossy compression and clustering that has also been productively applied to image encoding \cite{nasrabadi1988image}, texture synthesis \cite{wei2000fast}, and beyond.

The partition (\ref{eq:regiongeneral}) induced by the $k^{\rm th}$ of the $K=D^{(\ell)}$ max-affine splines in one MASO layer can be interpreted as approximating/encoding the input vector $\bz^{(\ell-1)}(\bx)$ by a ``partition centroid'' determined by the 1-hot vector $\left[T^{(\ell)}(\bz^{(\ell-1)}(\bx))\right]_{k,\bigcdot}$
and layer parameters $A^{(\ell)}[\bx],B^{(\ell)}[\bx]$ via (\ref{eq:tmaso}).
The partition (\ref{eq:factorial}) induced by the entire MASO layer can be interpreted as a {\em collaborative VQ} \cite{zemel1994minimum,hinton1994autoencoders}, which is particularly adept at representing complicated datasets with multiple object/features present simultaneously. 
Such techniques are also related to {\em factorial hidden Markov models} and other factorial models \cite{ghahramani1996factorial}, which have been used to represent the interactions among multiple loosely coupled processes.
We study the collaborative partitioning aspects of MASOs in more detail in Appendix \ref{sec:sm_spline_op}.

\begin{remark}
At this point, we can make Remark \ref{rem:sig1} more concrete by clarifying that same affine mapping parameters $A^{(\ell)}[\bx]$, $B^{(\ell)}[\bx]$ are applied not just to the signal $\bx$ but to all signals in the same partition region containing $\bx$, as in 
$A^{(\ell)}[T^{(\ell)}(\bz^{(\ell-1)}(\bx))]$, 
$B^{(\ell)}[T^{(\ell)}(\bz^{(\ell-1)}(\bx))]$.
\label{rem:partsig1}
\end{remark}

In keeping with the VQ interpretation of the MASO partition, we can also decode a signal approximation $\mu_r$ for all signals in region $r$.
The standard approach is via $K$-means \cite{jain2010data}. 
(Note that in this case, the $K$ of $K$-means equals $R$, the number of partition regions; we will use $R$ from now on and refer to the approach as $R$-means.)
For a set of (labelled or unlabelled) training signals $(\bx_n)_{n=1}^N$, the goal of $R$-means is to learn $R$ centroids $\bmu_r$, $r=1,\dots,R$ that solve
\begin{align}
    \min_{r=1,\dots,R} 
    \left\| \sum_{n=1}^N \bmu_r - \bx_n \right\|^2.
    \label{eq:kmeans1}
\end{align}
It then assigns signal $\bx$ to the closest centroid via
\begin{align}
    t(\bx_n)=\argmin_{r=1,\dots,R} 
    \| \bmu_r -\bx_n \|^2.
    \label{eq:kmeans2}
\end{align}

While solving (\ref{eq:kmeans1}) is computationally intractable, a number of suboptimal approximations have been developed, the most celebrated being the Lloyd algorithm \cite{jain2010data}.

Using a special parameterization of the DN, the MASO encoding/decoding interpretation conforms precisely with a $R$-means-based VQ system.
By massaging the formula for the 1-hot index of a single MASO layer output (recall (\ref{eq:tmaso2})) 
%
\begin{align}
    \big[t^{(\ell)}(\bx_n)\big]_k
    &= \argmax_{r=1,\dots,R} \left(
    \left\langle \big[A^{(\ell)}\big]_{k,r,\bigcdot},\bz^{(\ell-1)(\bx_n)}
    \right\rangle+\big[B^{(\ell)}\big]_{k,r}\right)
    \nonumber\\
    &= \argmin_{r=1,\dots,R} \left( \big\|\big[A^{(\ell)}\big]_{k,r,\bigcdot}-\bz^{(\ell-1)(\bx_n)} \big\|^2
    -\big\|\big[A^{(\ell)}\big]_{k,r,\bigcdot}\big\|^2
    -2\,\big[B^{(\ell)}\big]_{k,r}\right),
    \label{eq:masoKmeans}
\end{align}
we see that (\ref{eq:masoKmeans}) corresponds precisely to  (\ref{eq:kmeans1}) with 
$\bmu_r = \big[A^{(\ell)}\big]_{k,r,\bigcdot}$ provided
$\big\|\big[A^{(\ell)}\big]_{k,r,\bigcdot}\big\|^2 =-2\,\big[B^{(\ell)}\big]_{k,r}$.
In this case, the partition region is the {\em Voronoi tiling} of the input space according to (\ref{eq:kmeans2}).
Figure \ref{fig:VQbias} plots the partition regions and centroids $\bmu_r=\big[A^{(\ell)}\big]_{k,r,\bigcdot}$ for such a constrained max-affine spline.
%
%
We memorialize this result in the following proposition.

\begin{proposition}
\label{prop:kmean}
%
The input space partition of the $k^{\rm th}$ output of a MASO at level $\ell$ with parameters $\big[B^{(\ell)}\big]_{k,r}=-\frac{1}{2} \big\|\big[A^{(\ell)}\big]_{k,r,\bigcdot}\big\|^2$ 
is equivalent to an $R^{(\ell)}$-means partition using  $\bmu_r=[A^{(\ell)}]_{k,r,\bigcdot}$ for the centroids.
The partition is a Voronoi tiling with respect to the centroids.
\end{proposition}



\subsection{A New Image Distance Based on DN VQ}
\label{sec:clustering}

\begin{figure*}[!tb]
    \centering
\includegraphics[width=.99\textwidth]{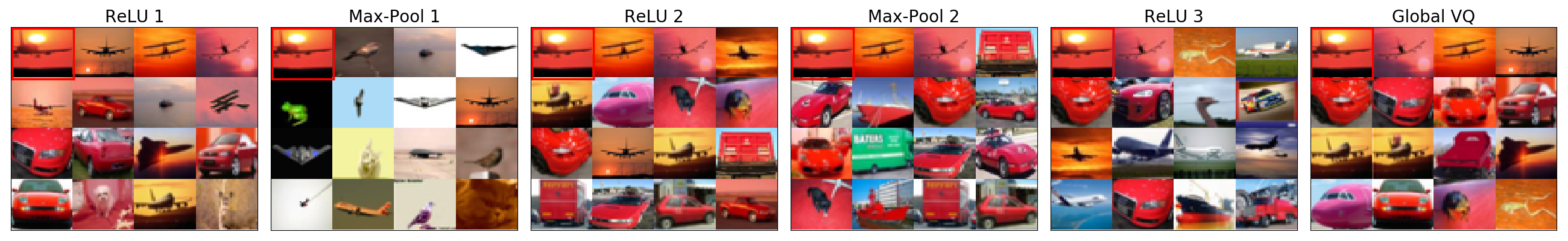}
\\[0mm]
{\small (a) Training with correct labels}
\\
\includegraphics[width=.99\textwidth]{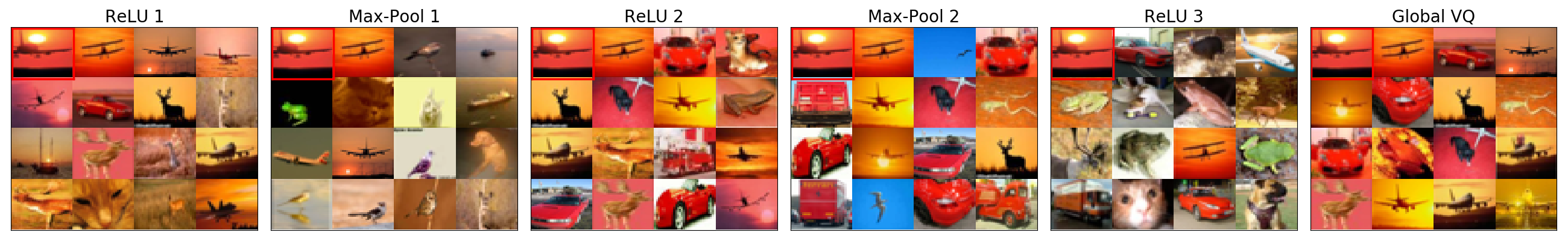}
\\[0mm]
{\small (b) Training with random labels}
\\
\includegraphics[width=.99\textwidth]{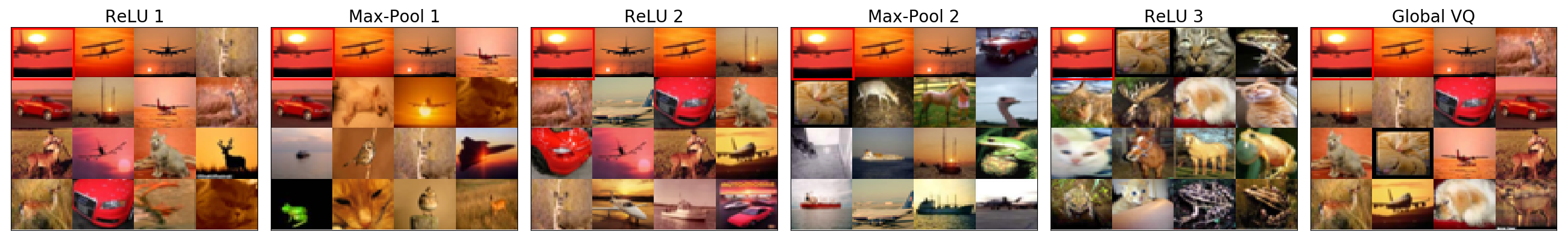}
\\[0mm]
{\small (c) No Training}
    \caption{
    (a) Illustration of the nearest neighbors of a test image in terms of the VQ-based distance (\ref{eq:distance1}), 
    (\ref{eq:distance2}) 
    for various layers of the {\em smallCNN} trained on the CIFAR10 dataset.
    In each panel, we plot the target image $\bx$ at top-left and its 15 nearest neighbors in the VQ distance at layer $\ell$ (see Appendix \ref{ap:dntopology} for the experimental details).
    From left to right, the images display layers $\ell=1,\dots, 6$, and we see that the nearest neighbors become more similar visually even as they become further in Euclidean distance.
    In particular, the first layer neighbors share similar colors and shapes (and thus are closer in Euclidean distance). 
    Later layer neighbors mainly come from the same class independently of their color or shape.
    (b) The same experiment but with the DN trained on data with randomly shuffled labels.
    (c) The same experiment but with the DN initialized with random weights and zero biases and not trained.   
     The similarity of the first 3 panels in (a)--(c) indicates that the early convolution layers of the DN are partitioning the training images based more on their visual features than their class membership.
     In (b), only after the fully connected layer (ReLU 3) is the VQ partition influenced by the labels.
}
\label{fig:partitions_cifar}
\end{figure*}

\begin{figure*}[!tb]
    \centering
\includegraphics[width=.99\textwidth]{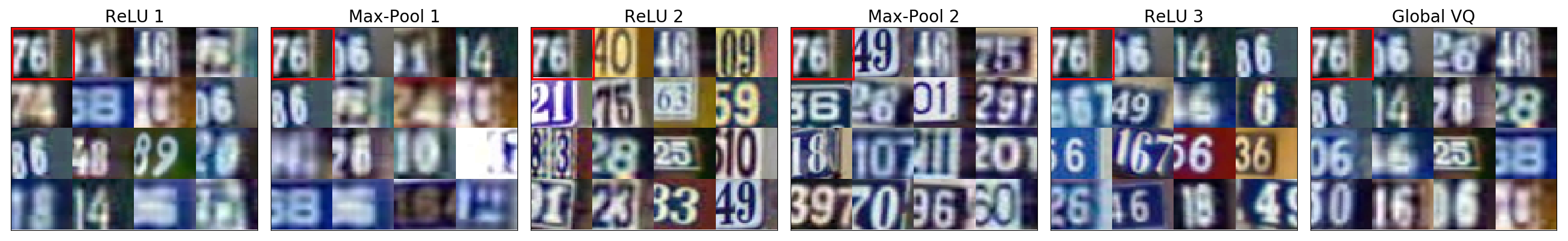}
\\[0mm]
{\small (a) Training with correct labels}
\\
\includegraphics[width=.99\textwidth]{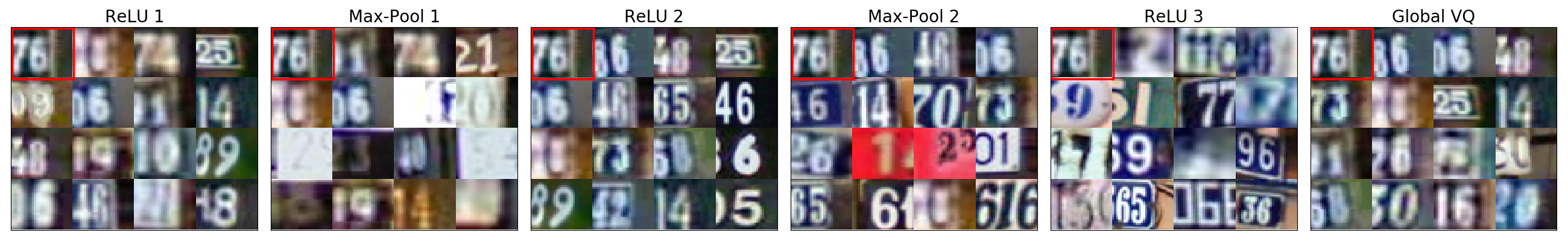}
\\[0mm]
{\small (b) Training with random labels}
\\
\includegraphics[width=.99\textwidth]{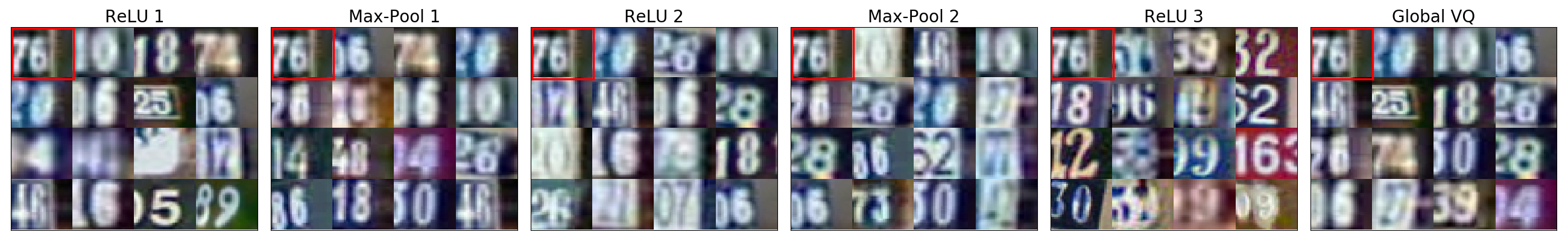}
\\[0mm]
{\small (c) No training}
    \caption{
    Reprise of the experiment of Figure~\ref{fig:partitions_cifar} for the SVHN dataset.
}
\label{fig:partitions_svhn}
\end{figure*}

\begin{figure*}[!tb]
    \centering
\includegraphics[width=.99\textwidth]{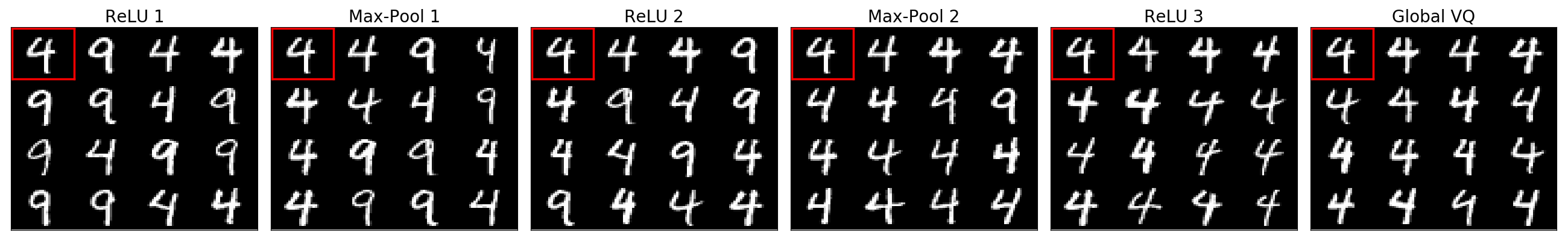}
\\[0mm]
{\small (a) Training with correct labels}
\\
\includegraphics[width=.99\textwidth]{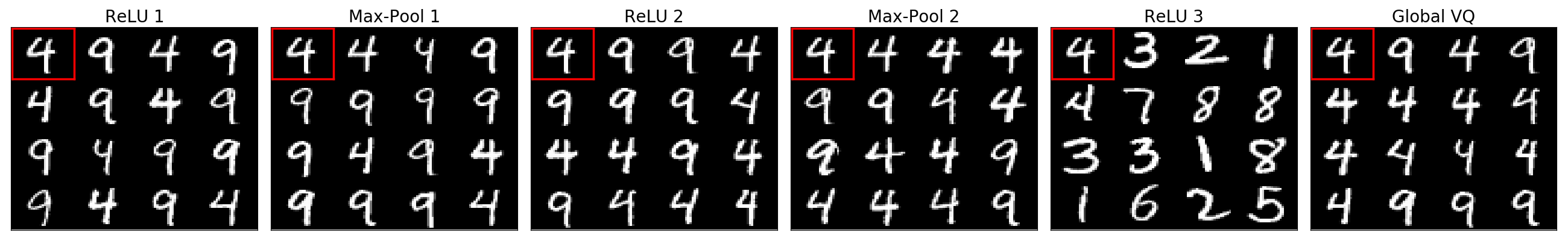}
\\[0mm]
{\small (b) Training with random labels}
\\
\includegraphics[width=.99\textwidth]{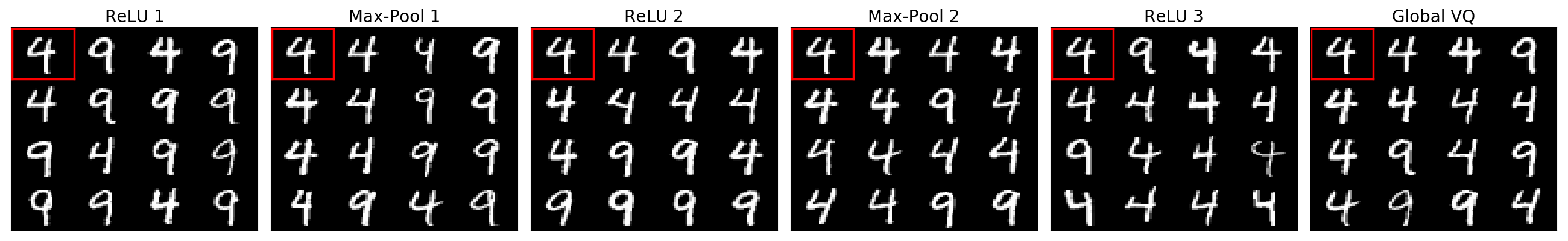}
\\[0mm]
{\small (c) No training}
    \caption{
    Reprise of the experiment of Figure~\ref{fig:partitions_cifar} for the MNIST dataset.
}
\label{fig:partitions_mnist}
\end{figure*}


The structure of the DN partition regions in the input signal space provide new tools to study and exploit the geometrical properties of the training and test data.
While, in general, it is impractical to instantiate the regions, we can access their membership via the selection matrix $T^{(\ell)}(\bz^{(\ell)}(\bx))$
from (\ref{eq:tmaso2}). 
Since two signals sharing the same selection matrices inhabit the same partition region, the concept of ``nearest neighbors'' inspires us to create a new signal distance that measures the distance between the partitions each signal inhabits.
Given signals $\bx_1$ and $\bx_2$, we define the distance as 
\begin{equation}
    d\left(T^{(\ell)}(\bx_1),T^{(\ell)}(\bx_2)\right)= 
    \frac{1}{D^{(\ell)}}\sum_{k=1}^{D^{(\ell)}} d'
    \left([T^{(\ell)}(\bx_1)]_{k,\bigcdot},[T^{(\ell)}(\bx_2)]_{k,\bigcdot}
    \right),
    \label{eq:distance1}
\end{equation}
with $d'$ a distance function of your choice.
This distance has the interpretation in terms of VQ of measuring how many of the $D^{(\ell)}$ outputs of layer $\ell$ share the same encoding of the input.
The fact that $[T^{(\ell)}(\bx)]_{k,\bigcdot}$ is a one-hot vector makes any $p$-norm based distance degenerate to
\begin{align}
 d'(\bu,\bv)=
 \left\{ \begin{array}{l} 0~~~~{\rm if}~ \bu=\bv\\ 1~~~~{\rm otherwise.} \end{array} \right.  
 \label{eq:distance2}
\end{align}
For any two signals, this distance takes a value $\in [0,1]$; it equals 0 if $\bx_1$ and $\bx_2$ share all of the same partitions and 1 if they share none of the same partitions.
For a given MASO, the distance can be computed easily via  (\ref{eq:tmasocomputation}).
Development of more sophisticated measures that would take into account the similarity between regions based on the parameters $\bA^{(\ell)},\bB^{(\ell)}$, is left for future work. 

For a ReLU MASO, the distance amounts to simply counting how many entries of the layer $\ell$ output are both positive or $0$ at the same indices $k=1,\dots,D^{(\ell)}$ for $\bx_1$ and $\bx_2$. 
For a max-pooling MASO, the distance amounts to simply counting how many $\argmax$ positions are the same for each of the max-pooling applications
for $\bx_1$ and $\bx_2$.


Figures 
\ref{fig:partitions_cifar}--\ref{fig:partitions_mnist}
provide visualizations of the nearest neighbors of a test image using our new VQ-based distance for the first five layers of the smallCNN topology trained for classification on the CIFAR10, SVHN, and MNIST datasets.  
As defined above, the distance can take only $D^{(\ell)}$ different values ($0$,$1/D^{(\ell)},2/D^{(\ell)},\dots,1$), meaning that, given an image, there are $D^{(\ell)}$ equivalence classes of neighbors based on each of those possible values. 
Visual inspection of the figures highlights that, as we progress through the layers of the DN, similar images become closer in VQ distance but further in Euclidean distance.
In particular, the similarity of the first 3 panels in (a)--(c) of the figures, which replicate the same experiment but using a DN trained with correct labels, a DN trained with incorrect labels, and a DN that is not trained at all, indicates that the early convolution layers of the DN are partitioning the training images based more on their visual features than their class membership.

The distance (\ref{eq:distance1}) is defined only with respect to the MASO at level $\ell$.
However, it is easily extended to take into account the composition of multiple MASO layers by composing their respective selection matrices.






%% file: conclusions.tex
\section{Conclusions}
\label{sec:conc}

We have applied the theory of max-affine spline operators (MASOs) to make new connections between deep networks (DNs) and approximation theory.
We group our key findings into two classes.
First, we have shown that, conditioned on the input signal, the output of a large class of DNs can be written as a simple affine transformation of the input.
This enabled us to link DNs directly to the classical theory of optimal classification via matched filters, provided insights into the effects of data memorization, and inspired a template orthogonality penalty that leads to significantly improved performance and reduced overfitting with no change to the DN architecture. 

Second, we have shown how the partition of the input signal space that is automatically induced by a MASO directly links DNs to the theory of vector quantization (VQ) and $K$-means clustering, which opens up a new geometric avenue to study how DNs organize signals.
This viewpoint inspired our new VQ-based distance for signals and images that is not only useful for probing the inner workings of DNs but also could be of independent utility.

There are many avenues for future work; here is a brief sampler.
New constraints on the templates beyond orthogonality could yield learning algorithms that boost performance even further.
A deeper study of the spline approximation of non-convex activation functions like the sigmoid could lead to new insights into why they are preferable in certain network topologies (e.g., recursive networks).
Extension of our analysis of the number of non-trivial and non-empty partition regions to the practical, large DNs could yield surprises.  
Since $K$-means is closely related to Gaussian mixture models, there are likely interesting connections with probabilistic models for deep learning \cite{patel2016probabilistic} and probabilistic trees \cite{romberg1999optimal}.
We can also study some of the popular DN tricks using MASOs.
For example, {\em dropout} \cite{gal2016dropout} can be interpreted as randomly perturbing the VQ partition regions puncturing the selection matrices $T^\ell$.
This maps training inputs into different (incorrect) neighboring regions where a different (partially wrong) affine mapping is applied. 
Judicious use of this technique will result in more robust clustering/partitioning of the training data that ensures that neighboring regions have close mappings.

%% file: acks.tex
\subsection*{\bf Acknowledgements}

Thanks to Romain Cosentino, Zichao (Django) Wang, and Daniel LeJeune for discussions and constructive criticism of the manuscript, as well as Daniel for drawing Figure~\ref{fig:toydn}.
This work was partially supported by 
ARO	grant W911NF-15-1-0316, 
AFOSR grant FA9550-14-1-0088, 
ONR grants N00014-17-1-2551 and N00014-18-12571,
DARPA grant G001534-7500,
and a DOD Vannevar Bush Faculty Fellowship (NSSEFF) grant N00014-18-1-2047.

%% file: APPENDIX/notation.tex
\section{Notation}
\label{sec:notation}
\small
\begin{center}
\begin{tabular}{|l|l|}
\hline 
$x$&Input/observation tensor of shape $(K,I,J)$\\[2mm]
$\bx$&Vectorized input of length $D$\\[2mm]
$\widehat{y}(x)$& Output/prediction associated  with input $x$\\[2mm]
$x_n$&Observation $n$\\[2mm] 
$y_n$&Label (target variable) associated with $x_n$, for classification \\
&$y_n \in \{1,\dots,C\},\;C>1$; for regression $y_n \in \mathbb{R}^C,C\geq 1$\\[2mm]
$\mathcal{D}$ (resp.\ $\mathcal{D}_s$)&Labeled training set with $N$ (resp. $N_s$) samples $\mathcal{D}=(X_n,Y_n)_{n=1}^Z$\\[2mm]
$\mathcal{D}_u$&Unlabeled training set with $N_u$ samples $\mathcal{D}_u=(X_n)_{n=1}^{N_u}$\\[2mm]
$f^{(\ell)}_{\theta^{(\ell)}}$&Mapping of deep net (DN) layer at level $\ell$ with internal \\
&parameters $\theta^{(\ell)},\ell=1,\dots,L$\\[2mm]
$\Theta$&Collection of all DN parameters $\Theta=\{\theta^{(\ell)},\ell=1,\dots,L\}$\\[2mm]
$f_{\Theta}$&DN mapping with $f_\Theta:\mathbb{R}^D\rightarrow \mathbb{R}^C$\\[2mm]
$(C^{(\ell)},I^{(\ell)},J^{(\ell)})$&Shape of the representation at layer $\ell$\\
& with $(C^{(0)},I^{(0)},J^{(0)})=(K,I,J)$, and
$(C^{(L)},I^{(L)},J^{(L)})=(C,1,1)$\\[2mm]
$D^{(\ell)}$&Dimension of the flattened representation at layer $\ell$\\
& with $D^{(\ell)}=C^{(\ell)}I^{(\ell)}J^{(\ell)}$, $D^{(0)}=D$, and $D^{(L)}=C$\\[2mm]
$z^{(\ell)}(x)$&Representation of $x$ (feature map) at layer $\ell$ in an unflattened\\
& format of shape $(C^{(\ell)},I^{(\ell)},J^{(\ell)})$, with $z^{(0)}(x)=x$\\[2mm]
$[z^{(\ell)}(x)]_{k,i,j}$&Value of the representation of $x$ at layer $\ell$, channel $c$, and spatial position $(i,j)$\\[2mm]
$\bz^{(\ell)}(x)$&Representation of $x$ at layer $\ell$ in a flattened format of dimension $D^{(\ell)}$\\[2mm]
$[\bz^{(\ell)}(x)]_k$&Value of $\bz^{(\ell)}(x)$ at dimension $k$. One has $[z^{(\ell)}]_{c,i,j}=[\bz^{(\ell)}]_{k}$\\
& with $k=c\times I^{(\ell)}\times J^{(\ell)}+i\times J^{(\ell)}+j$\\
\hline 
\end{tabular}
\end{center}
\normalsize

%% file: APPENDIX/background.tex
\section{Detailed Background on Deep Networks}
\label{app:operator}

We now describe a typical deep (neural) network (DN) topology, the deep convolutional neural network (CNN), to highlight the way the previously described operators can be combined to provide powerful predictors. Its main development goes back at least as far as  \cite{lecun1995learning} for digit classification.
The astonishing results that a CNN can achieve come from the ability of the blocks to convolve the learned filter-banks with their input, separating the underlying features present relative to the task at hand. 
This is followed by a scalar activation nonlinearity and a spatial sub-sampling to select, compress, and reduce the redundant representation while highlighting task dependent features. Finally, a multilayer perceptron (MLP) simply acts as a nonlinear classifier. 

In order to optimize the parameters $\Theta$ leading to the predicted output $\hat{y}(\bx)$, one makes use of (i) a labeled dataset $\mathcal{D}=\{(\bx_n,y_n),n=1,\dots,N\}$, (ii) a loss function $\mathcal{L}:\mathbb{R}^C \times \mathbb{R}^C \rightarrow \mathbb{R}$, (iii) a learning policy to update the parameters $\Theta$.
In the context of classification, the target variable $\bx_n$ associated to an input $\bx_n$ is  categorical, with $y_n \in \{1,\dots,C\}$. 
In order to predict such target, the output of the last operator of a network $\bz^{(L)}(\bx_n)$ is transformed via a softmax nonlinearity \cite{de2015exploration}. It is used to transform $\bz^{(L)}(\bx_n)$ into a probability distribution.
The used loss function quantifying the distance between $\hat{y}(\bx_n)$ and $Y_n$ is the cross-entropy (CE).
For regression problems, the target $y_n$ is continuous and thus the final DNN output is taken as the prediction $\hat{y}(\bx_n)=\bz^{(L)}(\bx_n)$. The loss function $\mathcal{L}$ is usually the ordinary squared error (SE).

Since all of the operations introduced above in standard DNNs are differentiable almost everywhere with respect to their parameters and inputs, given a training set and a loss function, one defines an update strategy for the weights $\Theta$. This takes the form of an iterative scheme based on a first order iterative optimization procedure. Updates for the weights are computed on each input and usually averaged over {\em mini-batches} containing $B$ exemplars with $B\ll N$. This produces an estimate of the correct update for $\Theta$ and is applied after each mini-batch. Once all the training instances of $\mathcal{D}$ have been seen, after $N/B$ mini-batches, this terminates an {\em epoch}. The dataset is then shuffled and this procedure is performed again. Usually a network needs hundreds of epochs to converge.
For any given iterative procedure, the updates are computed for all the network parameters by {\em backpropagation} \cite{hecht1988theory}, which follows from applying the chain rule of calculus. Common policies are Gradient Descent (GD) \cite{rumelhart1988learning} being the simplest application of backpropagation, Nesterov Momentum \cite{bengio2013advances} that uses the last performed updates in order to accelerate convergence and finally more complex adaptive methods with internal hyper-parameters updated based on the weights/updates statistics such as  Adam \cite{kingma2014adam}, Adadelta \cite{zeiler2012adadelta}, Adagrad \cite{duchi2011adaptive}, RMSprop \cite{tieleman2012lecture}, etc.

We now precisely describe the convolution, pooling, skip-connection, and recurrent operators one can use to create the mapping $f_\Theta$.
For the activation operator, there are no more details to add beyond the description in the main text.

\begin{figure}[t!]
    \centering
    \includegraphics[width=3in]{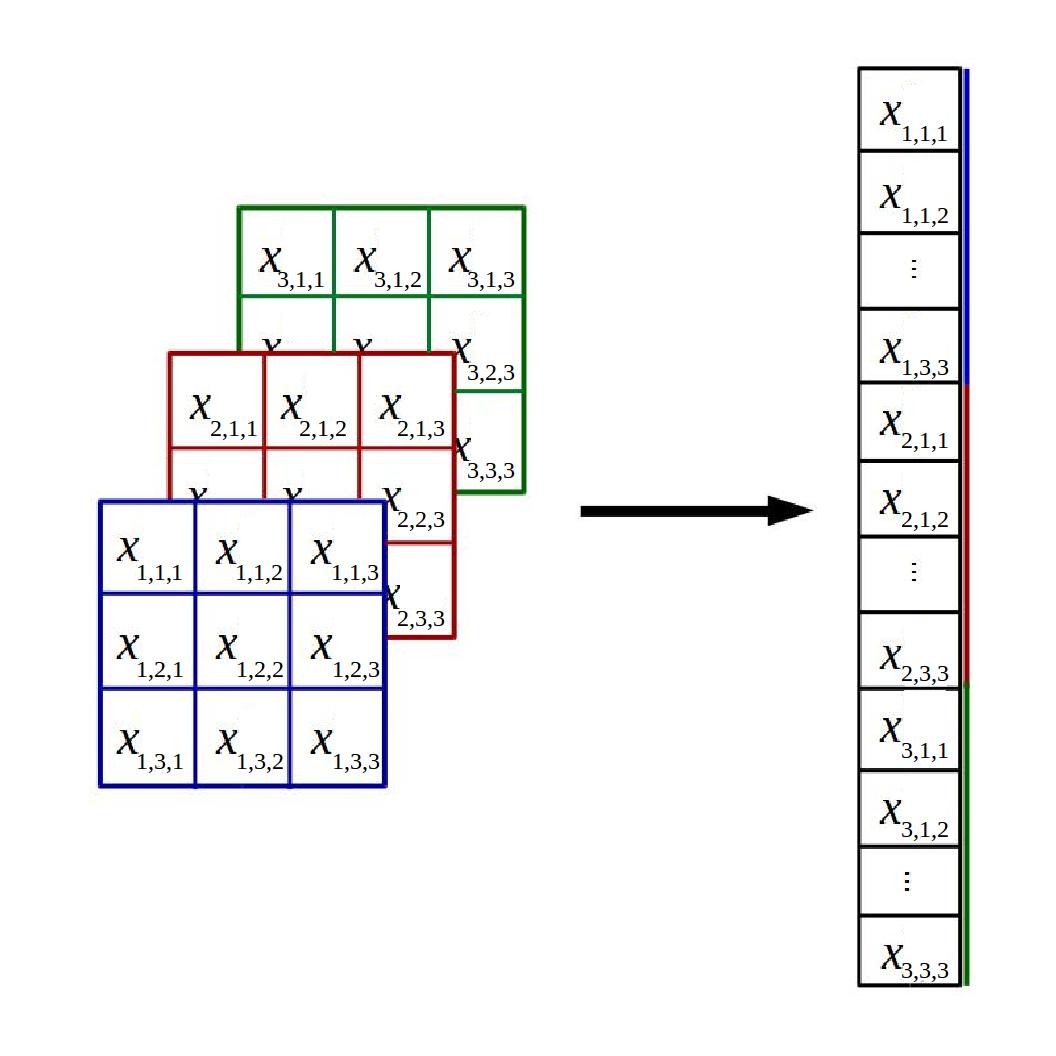}
    \caption{Reshaping of the multi-channel signal $z^{(\ell)}$ of shape $(3,3,3)$ to form the vector $\bz^{(\ell)}$ of dimension $27$.}
    \label{fig:reshape}
\end{figure}

\subsection{Convolution operator}
\label{sm:convop}

A {\em convolution operator} is defined as
\begin{align}
f^{(\ell)}_C \! \left(\bz^{(\ell-1)}(\bx)\right)=\bC^{(\ell)}\bz^{(\ell-1)}(\bx)+b^{(\ell)}_{\bC}
\end{align}
where a special structure is defined on $\bC^{(\ell)}$ so that it performs multi-channel convolutions on the vector $\bz^{(\ell-1)}(\bx)$. 

\begin{figure}[!t]
\centering
{\small (a)}
        \includegraphics[width=0.40\linewidth]{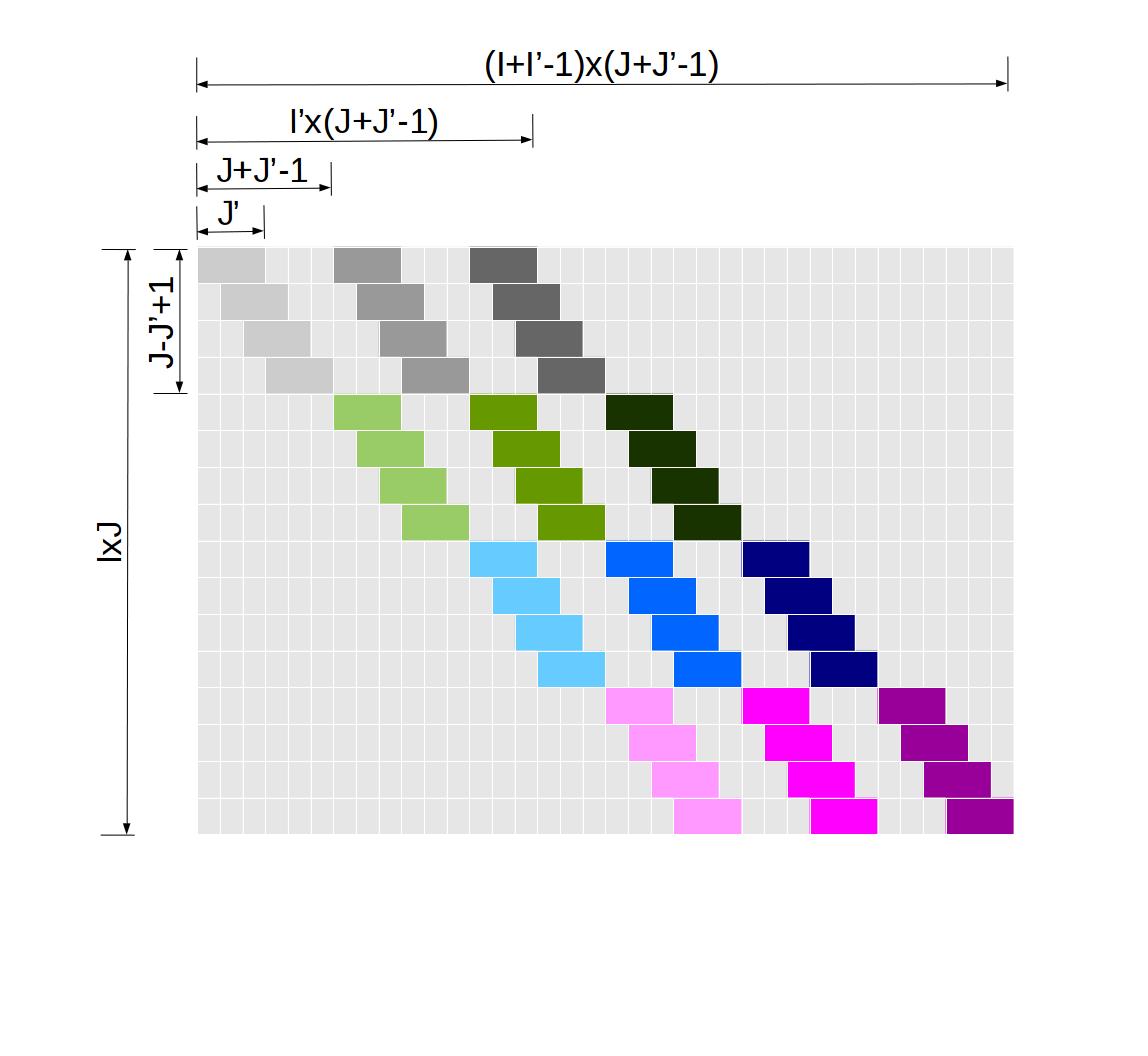}
{\small (b)}
        \includegraphics[width=0.40\linewidth]{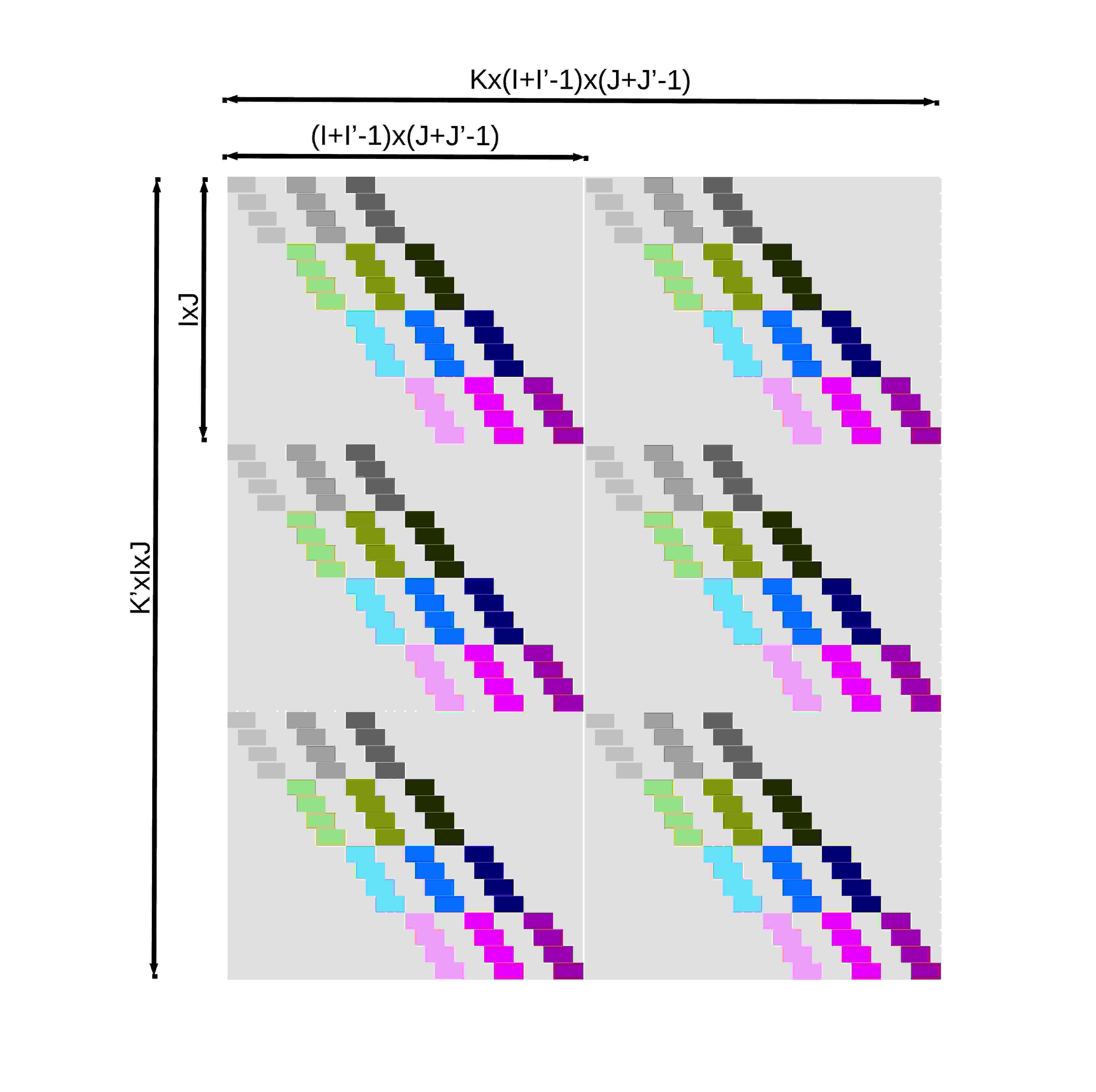}
        \label{fig:prob1_6_1}
\caption{(a) Depiction of one convolution matrix $\psi_{k,l}$. 
        (b) Depiction of the super convolution matrix $\bC$.}       
        \label{fig:w1}
\end{figure}

To highlight this fact, we first remind the reader of the multi-channel convolution operation performed on the unflatenned input $z^{(\ell-1)}(x)$ of shape $(C^{(\ell-1)},I^{(\ell-1)},J^{(\ell-1)})$ given a filter bank we denoted here as $\psi^{(\ell)}$ composed of $C^{(\ell)}$ filters, each of which is a $3D$ tensor of shape $(C^{(\ell-1)},I_{\bC}^{(\ell)},J_{\bC}^{(\ell)})$.  Hence $C^{(\ell-1)}$ represents the filters depth which equals to the number of channels of the input, and $(I_{\bC}^{(\ell)},J_{\bC}^{(\ell)})$ the spatial size of the filters.
The application of the linear filters $\psi^{(\ell)}$ on the signal forms another multi-channel signal as
\begin{align}
[\psi^{(\ell)} \star z^{(\ell-1)}(x)]_{c,i,j} ~=&~\sum_{k=1}^{C^{(\ell-1)}} [\psi^{(\ell)}]_{c,k,\bigcdot,\bigcdot}\star [z^{(\ell-1)}(x)]_{k,\bigcdot,\bigcdot}\nonumber \\
=&~\sum_{k=1}^{C^{(\ell-1)}} \sum_{m=1}^{I_{\bC}^{(\ell)}} \sum_{n=1}^{J_{\bC}^{(\ell)}} [\psi^{(\ell)}]_{c,k,m,n}  [z^{(\ell-1)}(\bx)]_{k,i-m,j-n},
\end{align}
where the output of this convolution contains $C^{(\ell)}$ channels, the number of filters in $\psi^{(\ell)}$. Then a bias term is added to each output channel, shared across spatial positions. We denote this bias term as $\xi \in \mathbb{R}^{C^{(\ell)}}$.
As a result, to create channel $c$ of the output, we first perform a $2D$ convolution of each channel $k=1,\dots,C^{(\ell-1)}$ of the input with the impulse response $[\psi^{(\ell)}]_{c,k,\bigcdot,\bigcdot}$, then sum those outputs element-wise over $k$ and finally add the bias, leading to $z^{(\ell)}(x)$ as
\begin{align}
[z^{(\ell)}(x)]_{c,\bigcdot,\bigcdot} ~=~ \sum_{k=1}^{C^{(\ell-1)}} \left([\psi^{(\ell)}]_{c,k,\bigcdot,\bigcdot}\star [z^{(\ell-1)}(x)]_{k,\bigcdot,\bigcdot}\right)+\xi_c.
\label{eq:conv1}
\end{align}
In general, the input is first transformed in order to apply some boundary conditions such as zero-padding, symmetric or mirror. Those are standard padding techniques in signal processing \cite{mallat1999wavelet}.
We now describe how to obtain the matrix $\bC^{(\ell)}$ and vector $b_{\bC}^{(\ell)}$ corresponding to the operations of Eq. \ref{eq:conv1} but applied on the flattened input $\bz^{(\ell-1)}(\bx)$ to produce the output vector $\bz^{(\ell)}(\bx)$.
The matrix $\bC^{(\ell)}$ is obtained by replicating the filter weights $[\psi^{(\ell)}]_{c,k,\bigcdot,\bigcdot}$ into the circulent-block-circulent matrices $\Psi^{(\ell)}_{c,k},c=1,\dots,C^{(\ell)},k=1,\dots,C^{(\ell-1)}$ \cite{jayaraman2009digital}
and stacking them into the super-matrix $\bC^{(\ell)}$
\begin{align}
\bC^{(\ell)} ~=~ 
\begin{bmatrix}
    \Psi^{(\ell)}_{1,1} & \Psi^{(\ell)}_{1,2} & \dots & \Psi^{(\ell)}_{1,C^{(\ell-1)}} \\
    \Psi^{(\ell)}_{2,1} & \Psi^{(\ell)}_{2,2} & \dots & \Psi^{(\ell)}_{2,C^{(\ell-1)}} \\
    \vdots & \vdots & \ddots & \vdots \\
    \Psi^{(\ell)}_{C^{(\ell)},1} & \Psi^{(\ell)}_{C^{(\ell)},2} & \dots & \Psi^{(\ell)}_{C^{(\ell)},C^{(\ell-1)}}
\end{bmatrix}.
\label{eq:bigC}
\end{align}
We provide an example in Figure \ref{fig:w1} for $\Psi^{(\ell)}_{c,k}$ and $\bC^{(\ell)}$.

By sharing the bias across spatial positions, the bias term $b_{\bC}^{(\ell)}$ inherits a specific structure. It is defined by replicating $\xi^{(\ell)}_c$ on all spatial position of each output channel $c$.

\subsection{Pooling operator}
\label{sm:poolingop}

\begin{figure}[!t]
\centering
\begin{minipage}{0.3\linewidth}
  \centering
  a)\includegraphics[width=1\textwidth]{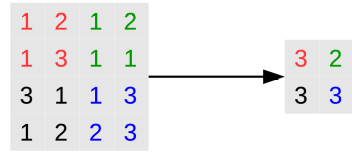}
  \end{minipage}
  \begin{minipage}{0.3\linewidth}
  \centering
  b)\includegraphics[width=1\textwidth]{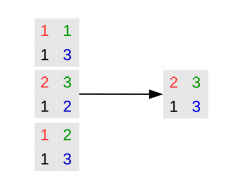}
  \end{minipage}
\qq
\caption{
(a) Depiction of $(2,2)$ spatial max-pooling 
applied to an input of shape $(1,4,4)$
with distinct regions per color; the max-pooling is applied on each.
(b) Depiction of $(3,1,1)$ channel pooling across 3 channels applied to an input of shape $(3,2,2)$; the channel-pooling is applied on each.
}
\label{fig:pooling}
\end{figure}

A {\em pooling} operator is a sub-sampling operation on its input according to a sub-sampling policy $\rho$ and a collection of regions on which $\rho$ is applied. We denote each region to be sub-sampled by $\mathcal{R}_d,d=1,\dots,D^{(\ell)}$ where $D^{(\ell)}$ is the total number of pooling regions. Each region contains the set of indices on which the pooling policy is applied, leading to
\begin{equation}
    \left[f^{(\ell)}_\rho \! \left(\bz^{(\ell-1)}(\bx)\right)\right]_k=\rho \!\left([\bz^{(\ell-1)}(\bx)]_{\mathcal{R}_k}\right),\quad k=1,\dots,D^{(\ell)}
\end{equation}
where $\rho$ is the pooling operator and  $[\bz^{(\ell-1)}(\bx)]_{\mathcal{R}_k}=\{[\bz^{(\ell-1)}(\bx)]_d,d\in \mathcal{R}_k\}$. Usually one uses mean or max pooling defined as
\begin{itemize}
    \item max-pooling: $\rho_{max}([\bz^{(\ell-1)}(\bx)]_{\mathcal{R}_k})=\max_{i\in \mathcal{R}_k} [\bz^{(\ell-1)}(x)]_i$,
    
    \item mean-pooling: $\rho_{mean}([\bz^{(\ell-1)}(\bx)]_{\mathcal{R}_k})=\frac{1}{Card(\mathcal{R}_k)}\sum_{i\in \mathcal{R}_d} [\bz^{(\ell-1)}(\bx)]_i$.
\end{itemize}

The regions $\mathcal{R}_k$ can be of different cardinality $\exists d_1,d_2 | {\rm Card}(\mathcal{R}_{d_1})\not = {\rm Card}(\mathcal{R}_{d_2}) $ and can be overlapping $\exists d_1,d_2 | {\rm Card}(\mathcal{R}_{d_1})\cap  {\rm Card}(\mathcal{R}_{d_2})\not = \emptyset $. However, in order to treat all input dimension, it is natural to require that each input dimension belongs to at least one region: $\forall k \in \{1,\dots, D^{(\ell-1)}\}, \exists d \in \{1,\dots, D^{(\ell)}\}| k\in \mathcal{R}_d$.
The benefits of a pooling operator are three-fold. First, by reducing the output dimension, it allows for faster computation and less memory requirement. Second, it allows to greatly reduce the redundancy of information present in the input $\bz^{(\ell-1)}(\bx)$. In fact, sub-sampling, even though linear, is common in signal processing after filter convolutions. Finally, in case of max-pooling, it allows to only backpropagate gradients through the pooled coefficient enforcing specialization of the neurons. The latter is the core of the winner-take-all strategy stating that each neuron specializes in what is performs best.


\subsection{Skip connection operator} 

A {\em skip-connection} operator can be considered as a bypass connection added between the input of an operator and its output. Hence, it allows for the input of an operator such as a convolutional operator or FC-operator to be linearly combined with its own output. The added connections lead to better training stability and overall performances because there always exists a direct linear link from the input to all inner operators. Simply written, given an operator $f^{(\ell)}_{\theta^{(\ell)}}$ and its input $\bz^{(\ell-1)}(\bx)$, the skip-connection operator is defined as
\begin{equation}
    f^{(\ell)}_{\rm skip} \!\left(\bz^{(\ell-1)}(\bx);f^{(\ell)}_{\theta^{(\ell)}}\right) =\bW^{(\ell)}_{\rm skip}\bz^{(\ell-1)}(x)+f^{(\ell)}_{\theta^{(\ell)}} \!\left(\bz^{(\ell-1)}(\bx)\right)+b_{\rm skip}^{(\ell)}
\end{equation}
where the linear connection can be cast into a convolutional one by replacing the variables.

\subsection{Recurrent operator}
\begin{figure}[t]
  \centering
  \includegraphics[width=4in]{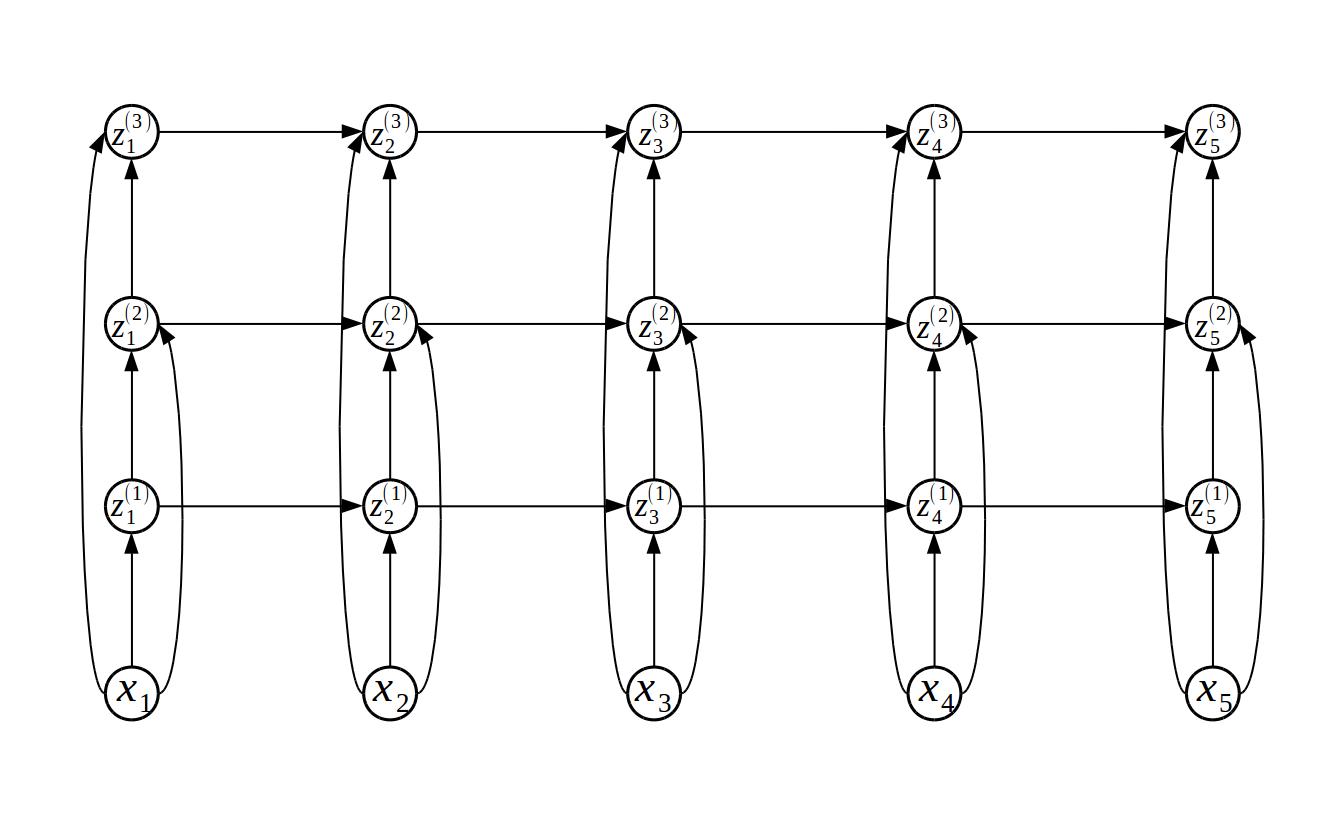}
  \caption{Depiction of a simple recursive neural network (RNN) with $3$ layers. Connections highlight input-output dependencies.}
\label{fig_rnn}
\end{figure}

A {\em recurrent operator} which aims to act on time-series. It is defined as a recursive application along time $t=1,\dots,T$ by transforming the input as well as using its previous output. The most simple form of this operator is a fully recurrent operator defined as
\begin{align}
    \bz^{(1,t)}(x)&=\sigma\left( W^{(in,h_1)}x^t+W^{(h_1,h_1)}\bz^{(1,t-1)}(x)+b^{(1)}  \right), \\
    \bz^{(\ell ,t)}(x)&=\sigma\left( W^{(in,h_\ell)}x^t+W^{(h_{\ell-1},h_\ell)}\bz^{(\ell-1),t}(x)+W^{(h_{\ell},h_\ell)}\bz^{(\ell,t-1)}(x)+b^{(\ell)} \right),
\end{align}
while some applications use recurrent operators on images by considering the series of ordered local patches as a time series, the main application resides in sequence generation and analysis especially with more complex topologies such as LSTM \cite{graves2005framewise} and GRU \cite{chung2014empirical} networks.
We depict the example topology in Figure \ref{fig_rnn}.

%% file: APPENDIX/affine.tex
\section{Details on Spline Operators}
\label{sec:layers}

\subsection{Spline functions}

\begin{definition}
Let $\Omega=\{\omega_1,\dots,\omega_R\}$ be a partition of $\mathbb{R}^D$ and let $\Phi=\{\phi_{1},\dots,\phi_{R}\}$ be a collection of local mappings with $\phi_{\omega_r}:\mathbb{R}^D\rightarrow \mathbb{R}$, Then we define multivariate spline with partition $\Omega$ and local mappings $\Phi$ the following
\begin{align}
    s[\Phi,\Omega](x)=\sum_{r=1}^R \phi_{r}(x)\Indic_{\{x \in \omega_r\}}
    =\phi[x](x),
\end{align}
where the input dependent selection is abbreviated via
\begin{equation}
    \phi[x]:=\phi_{r} \text{ s.t. } x \in \omega_r.
\end{equation}
If the local mappings $\phi_{r}$ are affine we have $\phi_{r}(x)=\langle [\a]_{r,\bigcdot},x\rangle+[\b]_{r}$ with $\a \in \mathbb{R}^{R\times D}$ and $\b \in \mathbb{R}$. We denote this functional as a \textbf{multivariate affine spline}:
\begin{align}
    s[\a,\b,\Omega](x)=\sum_{r=1}^R \left(\langle [\a]_{r,\bigcdot },x\rangle + [\b]_r \right)\Indic_{\{x \in \omega_r\}}
    =\langle \a[x],x\rangle +\b[x].
\end{align}
\end{definition}

\subsection{Spline operators}
\label{sec:sm_spline_op}

A natural extension of spline functions is the \textbf{{\em spline operator}} (SO) that we denote $S:\mathbb{R}^D \rightarrow \mathbb{R}^K,K>1$. We present here a general definition and propose in the next section an intuitive way to construct spline operators via a collection of multivariate splines, the special case of current DNs.
\begin{definition}
A spline operator is a mapping $S:\mathbb{R}^D \rightarrow \mathbb{R}^K$ defined by a collection of local mappings $\Phi^S=\{\phi^S_r:\mathbb{R}^D \rightarrow \mathbb{R}^K,r=1,\dots,R\}$ associated with a partition of $\mathbb{R}^D$ denoted as $\Omega^S=\{\omega^S_r,r=1,\dots,R\}$ such that
\begin{align}
    S[\Phi^S,\Omega^S](x)=\sum_{r=1}^R\phi^S_r(x)\Indic_{\{x \in \omega^S_r\}}
    =\phi^S[x](x),
\end{align}
where we denoted the region specific mapping associated to the input $x$ by $\phi^S[x]$.

\end{definition}
\begin{definition}
A special case occurs when the mappings $\phi^S_r$ are affine. We thus define in this case the \textbf{\textit{affine spline operator}} (ASO) which will play an important role for DN analysis. In this case, $\phi^S_r(x)=[A]_{.,r,.}x+[B]_{.,r}$, with $A\in \mathbb{R}^{K \times R \times D},B \in \mathbb{R}^{K\times R}$. As a result, a ASO can be rewritten as
\begin{align}
    S[A,B,\Omega^S](x)=\sum_{r=1}^R([A]_{\bigcdot,r,\bigcdot}x+[B]_{\bigcdot,r})\Indic_{\{x \in \omega^S_r\}}
    =A[x]x+b[x].
\end{align}
\end{definition}

Such operators can be defined via a collection of multivariate splines.
Given $K$  multivariate spline functions $s[\phi(k),\Omega(k)]:\mathbb{R}^D \rightarrow \mathbb{R},k=1,\dots,K$, their respective output is ''stacked'' to produce an output vector of dimension $K$. The internal parameters of each of the $K$ multivariate spline are $\Omega(k)$, a partition of $\mathbb{R}^D$ with $Card(\Omega(k))=R_k$ and $\Phi(k)=\{\phi(k)_{1},\dots,\phi(k)_{R_k}\}$. Stacking their respective output to form an output vector leads to the induced spline operator $S\left[\left(s[\Phi(k),\Omega(k)]\right)_{k=1}^K\right]$.

\begin{definition}
The spline operator $S:\mathbb{R}^D \rightarrow \mathbb{R}^K$ defined with $K$ 
multivariate splines $\left(s[\Phi(k),\Omega(k)]\right)_{k=1}^K$ with $s[\Phi(k),\Omega(k)]:\mathbb{R}^D \rightarrow \mathbb{R}$ is defined as 
\begin{align}
    S\left[\left(s[\Phi(k),\Omega(k)]\right)_{k=1}^K\right](x)=\left[ 
    \begin{matrix}
    s[\Phi(1),\Omega(1)](x)\\
    \vdots \\
    s[\Phi(K),\Omega(K)](x)
    \end{matrix}
    \right].
\end{align}
\end{definition}

The use of $K$ multivariate splines to construct a SO does not provide directly the explicit collection of mappings and regions $\Phi^S,\Omega^S$. Yet, it is clear that the SO is jointly governed by all the individual multivariate splines.
Let us first present some intuitions on this fact.
The spline operator output is computed with each of the $K$ splines having activated a region specific functional depending on their own input space partitioning. In particular, each of the region $\omega^S_r$ of the input space leading to a specific joint configuration $\phi^S_r$ is the one of interest, leading to $\Omega^S$ and $\Phi^S$.
We can thus write explicitly the new regions of the spline operator based on the ensemble of partition of all the involved multivariate splines as
\begin{align}\label{omegaS}
    \Omega^S=\left(\bigcup_{(\omega(1),\dots,\omega(K)) \in \Omega(1)\times \dots \times \Omega(K)}\left\{\bigcap_{k\in\{1,\dots,K\}}\omega(k)\right\}\right)\setminus\{\emptyset\}.
\end{align}
We can denote the number of region associated to this SO as $R^S=Card(\Omega^S)$.
From this, the local mappings of the SO $\phi^S_r$ correspond to the joint mappings of the splines being activated on $\omega^S_r$:
\begin{align}
\phi^S[\omega^S_r](x)=\left[ 
    \begin{matrix}
     \phi(1)[\omega^S_r](x)\\
    \vdots \\
     \phi(K)[\omega^S_r](x)
    \end{matrix}
    \right],
\end{align}
with $\phi(k)[\omega^S_r]=\phi(k)_q \in \Phi_k$ s.t. $\omega^S_r \subset \omega(k)_{q}$. In fact, for each region $\omega^S_r$ of the SO there is a unique region $\omega(k)_{q}$ for each of the splines $k=1,\dots,K$, such that it is a subset as $\omega^S_r\subset \omega(k)_{q}$ and it is disjoint to all others $\omega^S_r \cap \omega(k)_{l}=\emptyset, \forall l \not = q$. In other words, we have the following property:
\begin{align}
    \forall (r,k) \in \{1,\dots,R^S\}\times \{1,\dots,K\}, ~\exists ! ~q_k \in \{1,\dots,R_k\} \text{ s.t. } \omega^S_r \cap \omega(k)_{l}=\left\{\begin{array}{l}
        \omega^S_r,\;\;l = q_k\\
        \emptyset,\;\;l\not = q_k
    \end{array} \right.
\end{align}
where we recall that $\omega^S_r \cap \omega(k)_{l}=\omega^S_r\iff \omega^S_r \subset \omega(k)_{q_k}$.

This leads to the following SO formulation
\begin{align}
    S\left[\left(s[\Phi(k),\Omega(k)]\right)_{k=1}^K\right]\!(x)
    =&\sum_{r=1}^{R^S} \left[ 
    \begin{matrix}
    s[\Phi(1),\Omega(1)](x)\\
    \vdots \\
    s[\Phi(K),\Omega(K)](x)
    \end{matrix}
    \right]\Indic_{\{x\in \omega^S_r\}} 
    =\sum_{r=1}^{R^S} \left[ 
    \begin{matrix}
     \phi(1)[\omega^S_r](x)\\
    \vdots \\
     \phi(K)[\omega^S_r](x)
    \end{matrix}
    \right]\Indic_{\{x\in \omega^S_r\}} \nonumber\\
    =&\sum_{r=1}^{R^S} \phi^S[\omega^S_r](x)\Indic_{\{x\in\omega^S_r\}}
    =\phi^S[x](x).
\end{align}

We can now study the case of affine splines leading to ASOs. The affine property enables many notation simplifications.
It is defined as
\begin{align}
    S\left[\left(s[\a(k),\b(k),\Omega(k)]\right)_{k=1}^K\right](x)=&\sum_{r=1}^{R^S} \left[ 
    \begin{matrix}
    s[\a(1),\b(1),\Omega(1)](x)\\
    \vdots \\
    s[\a(K),\b(K),\Omega(K)](x)
    \end{matrix}
    \right]\Indic_{\{x\in \omega^S_r\}}\nonumber\\
    =&\sum_{r=1}^{R^S} \left[ 
    \begin{matrix}
    \langle \a(1)[\omega^S_r], x\rangle+b(K)[\omega^S_r]\\
    \vdots \\
    \langle \a(1)[\omega^S_r], x\rangle+b(K)[\omega^S_r]
    \end{matrix}
    \right]\Indic_{\{x\in \omega^S_r\}}\nonumber\\
    =&\sum_{r=1}^{R^S} \left(\left[ 
    \begin{matrix}
    \a(1)[\omega^S_r]^T\\
    \vdots \\
    \a(K)[\omega^S_r]^T
    \end{matrix}
    \right]x+
    \left[ 
    \begin{matrix}
    \b(1)[\omega^S_r]\\
    \vdots \\
    \b(K)[\omega^S_r]
    \end{matrix}
    \right]
    \right)\Indic_{\{x\in \omega^S_r\}}\nonumber\\
    =&\sum_{r=1}^{R^S} (A[\omega^S_r] x+B[\omega^S_r])\Indic_{\{x\in\omega^S_r\}}
    = A[x]x+B[x].
\end{align}
The collection of matrices and biases and the partitions completely define an ASO.

\subsection{Simplified Max-Affine Splines}
\label{app:SMAS}
Leveraging a unique bias vector for a MASO as opposed to a collection $K\times R$ scalars reduces the representation power of a MASO. However, standard activation functions in DNs can be thought of as ''degenerate'' as they are scalar nonlinearities applied after application of the same affine transform. Due to this, taking the ReLU as an illustrative example, the different states of the ReLU correspond to different scaling of the affine parameter used to project the input as opposed to using independent affine transformation parameters. As a result, for those nonlinearities, one can recover the exact MASO formulation from the simplified version by finding $\b'$ s.t. for each output dimension $\langle [A]_{k,r,\bigcdot},\b'\rangle=[B]_{k,r}$ (solving the linear system). The sufficient condition to recover the MASOs used in DN is to have linearly independent filters (that is, for DNs, the filters $[A^{(\ell)}]_{k,2,\bigcdot}$ for all $k$).


%% file: APPENDIX/proofs.tex
\section{Proofs}\label{sec:proofs}

The proofs of any propositions and theorems not provided below are trivial.

\subsection{Proofs of Propositions~\ref{thm:masoequiv}--\ref{thm:pool}}

Those propositions are trivial by definition and uniqueness of a piecewise affine convex mapping and their link to the rewritten operators. Also, see Sec.~\ref{sec:proof_rewritting} as well as \cite{boyd2004convex} for the convex properties and conditions.

\subsection{Proof of Proposition~\ref{thm:layerMASO} (DN layers are MASOs)}
\label{sec:proof_rewritting}

We now provide examples to better illustrate the DN layer reparametrization as a composition of DN operators.


\paragraph{Affine Transform+Activation:}
Any MASO following an affine transformation can be computed through the definition of a new reparametrized MASO with
\begin{align}
    S\!\left[A^{(\ell)}_\sigma,b^{(\ell)}_\sigma\right](W^{(\ell)} \bx+b^{(\ell)}_\bW)
    =&\left[ \begin{array}{c} \max_{r=1,\dots,R_1} [A^{(\ell)}_\sigma]_{1,r,\bigcdot}^T(W^{(\ell)}\bx+b^{(\ell)}_\bW)+[b^{(\ell)}_\sigma]_{1,r} \\
\dots \\
\max_{r=1,\dots,R_K} [A^{(\ell)}_\sigma]_{K,r,\bigcdot}^T(W^{(\ell)}\bx+b^{(\ell)}_\bW)+[b^{(\ell)}_\sigma]_{K,r}  
\end{array} \right]
\nonumber\\
    =&\left[ \begin{array}{c} \max_{r=1,\dots,R_1} (W^{(\ell)T}[A^{(\ell)}_\sigma]_{1,r})^T\bx+[A^{(\ell)}_\sigma]_{1,r}^T b^{(\ell)}_\bW+[b^{(\ell)}_\sigma]_{1,r} \\
\dots \\
\max_{r=1,\dots,R_K} (W^{(\ell)T}[A^{(\ell)}_\sigma]_{K,r})^T\bx+[A^{(\ell)}_\sigma]_{K,r}^T b^{(\ell)}_\bW+[b^{(\ell)}_\sigma]_{K,r} \end{array} \right]
\nonumber\\
=& ~ S\!\left[A,b\right](\bx),
\end{align}
with $[A^{(\ell)}]_{k,r,\bigcdot}= W^{(\ell)T}[A^{(\ell)}_\sigma]_{k,r,\bigcdot}$ and $[b^{(\ell)}]_{k,r}=[b^{(\ell)}_\sigma]_{k,r}+\langle b_\bW^{(\ell)},[A^{(\ell)}_\sigma]_{k,r,\bigcdot}\rangle$ 

\paragraph{Linear skip-connection:}
Any MASO with a skip-connection can be computed via definition of a new reparametrized MASO with
\begin{align}
    S\!\left[A^{(\ell)},b^{(\ell)}\right](\bx)+\bx
    =&\left[ \begin{array}{c}
     \max_{r=1,\dots,R_1} [A^{(\ell)}]_{1,r,\bigcdot}^T \bx+[b^{(\ell)}]_{1,r} \\
    \dots \\
      \max_{r=1,\dots,R_K} [A^{(\ell)}]_{K,r,\bigcdot}^T\bx+[b^{(\ell)}]_{K,r}  \end{array} \right]+\bx\\
    =&\left[ \begin{array}{c} \max_{r=1,\dots,R_1} ([A^{(\ell)}]_{1,r,\bigcdot}+\be_1)^T\bx+[b^{(\ell)}]_{1,r} \\
\dots \\
\max_{r=1,\dots,R_K} ([A^{(\ell)}]_{K,r,\bigcdot}+\be_D)^T\bx+[b^{(\ell)}]_{K,r} \end{array} \right]\\
=&  S\!\left[A',b'\right](\bx)\\
\end{align}
with $[A']_{k,r,\bigcdot}=[A^{(\ell)}]_{k,r,\bigcdot}+\be_r$ and $[b']_{k,r}=[b^{(\ell)}]_{k,r}$ 

\paragraph{ResNet layer:}
Any ResNet layer can be computed via definition of a new reparametrized MASO with
\begin{align}
    S\!\left[A^{(\ell)}_\sigma,b^{(\ell)}_\sigma\right]&(\bC \bx+b_{\bC})+\bC_{\rm skip}\bx+b_{\rm skip}
    \nonumber\\
    =&\left[ \begin{array}{c}  [\bC_{\rm skip}]_{1,\bigcdot}^T\bx+ [b_{\rm skip}]_{1}+ \max_{r=1,\dots,R_1} [A^{(\ell)}_\sigma]_{1,r,\bigcdot}^T(\bC \bx+b_{\bC})+[b^{(\ell)}_\sigma]_{1,r} \\
\dots \\
 \;[\bC_{\rm skip}]_{K,\bigcdot}^T\bx+ [b_{\rm skip}]_{K}+ \max_{r=1,\dots,R_K} [A^{(\ell)}_\sigma]_{K,r,\bigcdot}^T(\bC \bx+b_{\bC})+[b^{(\ell)}_\sigma]_{K,r} 
\end{array} \right]
\nonumber\\[2mm]
    =&\left[ \begin{array}{c} \max_{r=1,\dots,R_1} (\bC^T[A^{(\ell)}_{\sigma}]_{1,r,\bigcdot}+[\bC_{\rm skip}]_{1,\bigcdot})^T\bx+[A^{(\ell)}_{\sigma}]_{1,r,\bigcdot}^Tb_{\bC}+[b^{(\ell)}_{\sigma}]_{1,r}+[b_{\rm skip}]_{1} \\
\dots \\
\max_{r=1,\dots,R_K} (\bC^T[A^{(\ell)}_{\sigma}]_{K,r,\bigcdot}+[\bC_{\rm skip}]_{K,\bigcdot})^T\bx+[A^{(\ell)}_{\sigma}]_{K,r,\bigcdot}^Tb_{\bC}+[b^{(\ell)}_{\sigma}]_{K,r}+[b_{\rm skip}]_{K}
\end{array} \right]
\nonumber\\
=& ~ S\!\left[A^{(\ell)},b^{(\ell)}\right](\bx)
\end{align}
with $[A^{(\ell)}]_{k,r,\bigcdot}=\bC^T[A^{(\ell)}_{\sigma}]_{k,r,\bigcdot}+[\bC_{\rm skip}]_{k,\bigcdot}$ and $[b^{(\ell)}]_{k,r}=[A^{(\ell)}_{\sigma}]_{k,r,\bigcdot}^Tb_{\bC}+[b^{(\ell)}_{\sigma}]_{k,r}+[b_{\rm skip}]_{k}$.

\subsection{Proof of Proposition~\ref{thm:saliency}}

This result is direct from the fact that all the mappings are piecewise affine. Hence the derivative of the prediction or any neuron output with respect to the output gives the slope of the projection, itself corresponding to the definition of saliency maps. This is exact for piecewise affine mappings and will be the same for any input in the same input space region. However for other DNs (e.g., sigmoid), the derivative is only exact at the point at which it is computed and corresponds to a Taylor approximation of the DN mapping.

\subsection{Proof of Proposition~\ref{thm:lipschi}}
\label{proof_lipschi}

We now derive the Lipschitz constant of the softmax nonlinearity. We now that for any differentiable mapping $f$ we have the following upper bound 
\begin{equation}
    ||f(x)-f(y)||^2_2\leq \max_x ||Df(x)||_F^2||x-y||^2_2,
\end{equation}
with $Df$ the total derivative. Thus we now analyze $\max_x ||Df(x)||_F^2$ to find the softmax Lipschitz constant. We thus obtain by taking $f$ the softmax nonlinearity with input $x:=\bz^{(L)}$
\begin{align}
    \max_{p \in \bigtriangleup_D} ||Df(p)||_F^2=&\max_{p \in \bigtriangleup_D}\sum_{i=1}^D\sum_{j=1,j\not = i}^Dp_i^2p_j^2+\sum_{i=1}^Dp_i^2(1-p_i)^2,
\end{align}
where $\bigtriangleup_D$ denotes the simplex of dimension $D$ 
\begin{equation}
    \bigtriangleup_D=\{x\in \mathbb{R}^{D+}|\sum_i x_i=1\}.
\end{equation}
We now aim to maximize this functional, but instead of maximizing with respect to the input$\bz^{(L)}$ we directly maximize with respect to the induced density. The value of the maximum is the same. Note that we have a bijection between $\bz^{(L)}$ and $p$. The Lagrangian is the augmented loss function with the constraint
\begin{align}
    \mathcal{L}(p)=\sum_{i=1}^D\sum_{j=1,j\not = i}^Dp_i^2p_j^2+\sum_{i=1}^Dp_i^2(1-p_i)^2+\lambda(\sum_{i=1}^Dp_i-1),
\end{align}
and with $p_i=\frac{e^{[\bz^{(L)}]_i}}{\sum_i e^{[\bz^{(L)}]_i}}$.
We now seek the stationary points:
\begin{align}
    \frac{\partial \mathcal{L}}{\partial p_k} &= 4p_k\sum_{j=1,j\not = k}^Dp_j^2+2p_k(1-p_k)^2-2(1-p_k)p_k^2+\lambda
    \nonumber\\
    &= 4p_k\sum_{j=1,j\not = k}^Dp_j^2+2p_k-4p_k^2+2p_k^3-2p_k^2+2p_k^3+\lambda
    \nonumber\\
    &= 4p_k\sum_{j=1,j\not = k}^Dp_j^2+2p_k-6p_k^2+4p_k^3+\lambda
    \nonumber\\
    &= 4p_k\sum_{j=1}^Dp_j^2+2p_k-6p_k^2+\lambda\;\;\forall k,
    \nonumber\\
    \frac{\partial \mathcal{L}}{\partial \lambda} &= \sum_{i=1}^Dp_i-1.
\end{align}
We now solve the system $\nabla \mathcal{L}=0$. Since we have $\frac{\partial \mathcal{L}}{\partial p_k}=0~\forall k$, it is clear that $\sum_{k=1}^D\frac{\partial \mathcal{L}}{\partial p_k}=0$, leading to
\begin{align}
\sum_{k=1}^D\frac{\partial \mathcal{L}}{\partial p_k}&=0\\
\implies \sum_{k=1}^D\left( 4p_k\sum_{j=1}^Dp_j^2+2p_k-6p_k^2+\lambda\right) &=0\\
\implies 4\sum_{j=1}^Dp_j^2+2-6\sum_{j=1}^Dp_j^2+D\lambda &=0\\
\implies D\lambda&=2\sum_{j=1}^Dp_j^2-2\\
\implies \lambda = \frac{2}{D}(\sum_{j=1}^Dp_j^2-1). 
\end{align}
Plugging this into the gradient of $\mathcal{L}$, we arrive at the updated system of equations
\begin{align}
    4p_k\sum_{j=1}^Dp_j^2+2p_k-6p_k^2+\frac{2}{D}\left(\sum_{j=1}^Dp_j^2-1\right)
    & = 0\;\;\forall k
    \nonumber\\
    \sum_{i=1}^Dp_i & = 1
\end{align}
which can be written in vector form as
\begin{align}
    \textbf{p}(2||\textbf{p}||^2_2+1)-3 \textbf{p} \circ \textbf{p}=(1-||\textbf{p}||^2_2)
    {\rm vec}(1/D),
\end{align}
where $\circ$ denotes the Hadamard product.
It is clear that $\textbf{p}$ must be constant across its dimensions; this constraint leads to $p={\rm vec}(1/D)$. At the point, we have that the value of the maximum equals 
\begin{align}
    \mathcal{L}({\rm vec}(1/D))
    &=\sum_{i=1}^D\sum_{j=1,j\not = i}^D \frac{1}{D^4}+\sum_{i=1}^D\frac{1}{D^2}(1-\frac{1}{D})^2
    \nonumber\\
    &=\frac{D-1}{D^3}+\frac{1}{D}(1-\frac{1}{D})^2
    \nonumber\\
    &=\frac{D-1}{D^3}+\frac{1}{D}-\frac{2}{D^2}+\frac{1}{D^3}
    \nonumber\\
    &=\frac{D-1+D^2-2D+1}{D^3}
    \nonumber\\
    &=\frac{D(D-1)}{D^3}
    \nonumber\\
    &=\frac{D-1}{D^2}.
\end{align}
Note also that the softmax generalizes the sigmoid. In the case of the sigmoid (softmax with $D^{(L)}=2$ or 2 classes), we directly obtain that the maximum of the gradient occurs at the point $p_1=p_2=0.5$.
\hfill$\Box$

\subsection{Proof of Proposition~\ref{thm:colinear} }
\label{sec:proof_colinear}

 We aim to minimize the cross-entropy loss function for a given input $\bx_n$ belonging to class $y_n$. We abbeviate $[A[\bx_n]]_{c,.}$ as $A[\bx_n]_c$.We also have the constraint $\sum_{c=1}^C||A[\bx_n]_c||^2\leq K$. The loss function is thus convex on a convex set. It is thus sufficient to find a extremum point. We denote the augmented loss function with the Lagrange multiplier as
 \begin{align}
     l(A[\bx_n]_1,\dots,A[\bx_n]_C,\lambda)=-\langle A[\bx_n]_{y_n},\bx_n\rangle+\log \left(\sum_{c=1}^Ce^{\langle A[\bx_n]_c,\bx_n\rangle}\right)-\lambda \left(\sum_{c=1}^C ||A[\bx_n]_c ||^2-K\right).
 \end{align}
 The sufficient KKT conditions are thus
 \begin{align}
     &\frac{d l}{d A[\bx_n]_1}=-1_{\{y_n=1\}}x+\frac{e^{\langle A[\bx_n]_1, \bx_n\rangle }}{\sum_{c=1}^C e^{\langle A[\bx_n]_c, \bx_n\rangle }}\bx_n-2\lambda A[\bx_n]_1= \underline{0}\\
     & \vdots \nonumber\\
     &\frac{d l}{d A[\bx_n]_C}=-1_{\{y_n=C\}}x+\frac{e^{\langle A[\bx_n]_C, \bx_n\rangle }}{\sum_{c=1}^C e^{\langle A[\bx_n]_c, \bx_n\rangle }}x-2\lambda A[\bx_n]_C=\underline{0}\\
     &\frac{\partial l}{\partial \lambda}=K-\sum_{c=1}^C ||A[\bx_n]_c ||^2=0.
 \end{align}
 We first proceed by identifying $\lambda$ as follows
 \begin{align}
 \left.
 \begin{array}{l}
     \frac{d l}{d A[\bx_n]_1}= \underline{0}\\
      \vdots \\
     \frac{d l}{d A[\bx_n]_C}=\underline{0}\\
 \end{array}\right\} &\implies \sum_{c=1}^CA[\bx_n]_c^T\frac{d l}{d A[\bx_n]_c}=0\\
 &\implies -\langle A[\bx_n]_{y_n},x\rangle +\sum_{c=1}^C\frac{e^{\langle A[\bx_n]_c, \bx_n\rangle }}{\sum_{c=1}^C e^{\langle A[\bx_n]_c, \bx_n\rangle }}\langle A[\bx_n]_c,\bx_n\rangle-2\lambda \sum_{c=1}^C||A[\bx_n]_c||^2=0\\
 &\implies \lambda = \frac{1}{2K}\left(\sum_{c=1}^C\frac{e^{\langle A[\bx_n]_c, \bx_n\rangle }}{\sum_{c=1}^C e^{\langle A[\bx_n]_c, \bx_n\rangle }}\langle A[\bx_n]_c,\bx_n\rangle-\langle A[\bx_n]_{y_n},\bx_n\rangle \right). 
 \end{align}
 Now we plug $\lambda$ into $\frac{d l}{d A[\bx_n]_k}~\forall k=1,\dots,C$ to obtain 
 \begin{align}
     \frac{d l}{d A[\bx_n]_k}=&-1_{\{y_n=k\}}x+\frac{e^{\langle A[\bx_n]_k, \bx_n\rangle }}{\sum_{c=1}^C e^{\langle A[\bx_n]_c, \bx_n\rangle }}\bx_n-2\lambda A[\bx_n]_k
     \nonumber\\
     =&-1_{\{y_n=k\}}\bx_n+\frac{e^{\langle A[\bx_n]_k, \bx_n\rangle }}{\sum_{c=1}^C e^{\langle A[\bx_n]_c, \bx_n\rangle }}\bx_n-\frac{1}{K}\sum_{c=1}^C\frac{e^{\langle A[\bx_n]_c, x\rangle }}{\sum_{c=1}^C e^{\langle A[\bx_n]_c, \bx_n\rangle }}\langle A[\bx_n]_c,\bx_n\rangle A[\bx_n]_k
     \nonumber\\
     &+\frac{1}{K}\langle A[\bx_n]_{y_n},\bx_n\rangle A[\bx_n]_k. 
 \end{align}
We now leverage the fact that $A[\bx_n]_i=A[\bx_n]_j~\forall i,j \not = y_n$ to simplify the notation to 
 \begin{align}
    \frac{d l}{d A[\bx_n]_k}=&\left(\frac{e^{\langle A[\bx_n]_k, \bx_n\rangle }}{\sum_{c=1}^C e^{\langle A[\bx_n]_c, x\rangle }}-1_{\{k=y_n\}}\right)\bx_n-\frac{C-1}{K}\frac{e^{\langle A[\bx_n]_i, \bx_n\rangle }}{\sum_{c=1}^C e^{\langle A[\bx_n]_c, \bx_n\rangle }}\langle A[\bx_n]_i,\bx_n\rangle A[\bx_n]_k
    \nonumber\\
     &+\frac{1}{K}\left(1-\frac{e^{\langle A[\bx_n]_{y_n}, \bx_n\rangle }}{\sum_{c=1}^C e^{\langle A[\bx_n]_c, \bx_n\rangle }}\right)\langle A[\bx_n]_{y_n},\bx_n\rangle A[\bx_n]_k
     \nonumber\\
     =&\left(\frac{e^{\langle A[\bx_n]_k, \bx_n\rangle }}{\sum_{c=1}^C e^{\langle A[\bx_n]_c, \bx_n\rangle }}-1_{\{k=y_n\}}\right)\bx_n-\frac{C-1}{K}\frac{e^{\langle A[\bx_n]_i, \bx_n\rangle }}{\sum_{c=1}^C e^{\langle A[\bx_n]_c, x\rangle }}\langle A[\bx_n]_i,x\rangle A[\bx_n]_k
     \nonumber\\
     &+\frac{C-1}{K}\frac{e^{\langle A[\bx_n]_i, \bx_n\rangle }}{\sum_{c=1}^C e^{\langle A[\bx_n]_c, \bx_n\rangle }}\langle A[\bx_n]_{y_n},\bx_n\rangle A[\bx_n]_k
     \nonumber\\
     =&\left(\frac{e^{\langle A[\bx_n]_k, \bx_n\rangle }}{\sum_{c=1}^C e^{\langle A[\bx_n]_c, x\rangle }}-1_{\{k=y_n\}}\right)\bx_n+\frac{C-1}{K}\frac{e^{\langle A[\bx_n]_i, \bx_n\rangle }}{\sum_{c=1}^C e^{\langle A[\bx_n]_c, \bx_n\rangle }}\langle A[\bx_n]_{y_n}-A[\bx_n]_i,\bx_n\rangle A[\bx_n]_k. 
\end{align}
 
We now proceed by using the proposed optimal solutions $A^*[\bx_n]_c,c=1,\dots,C$ and demonstrate that they lead to an extremum point that, by nature of the problem, is the global optimum. We denote by $i$ any index different from $y_n$.
We first consider the case $k=y_n$:
 \begin{align}
    \frac{d l}{d A[\bx_n]_{y_n}}&=-\left(1-\frac{e^{\langle A[\bx_n]_{y_n}, \bx_n\rangle }}{\sum_{c=1}^C e^{\langle A[\bx_n]_c, x\rangle }}\right)x+\frac{C-1}{K}\frac{e^{\langle A[\bx_n]_i, \bx_n\rangle }}{\sum_{c=1}^C e^{\langle A[\bx_n]_c, x\rangle }}\langle A[\bx_n]_{y_n}-A[\bx_n]_i,\bx_n\rangle A[\bx_n]_{y_n}
    \nonumber\\
    &=-(C-1)\frac{e^{\langle A[\bx_n]_i, \bx_n\rangle }}{\sum_{c=1}^C e^{\langle A[\bx_n]_c, \bx_n\rangle }}x+\frac{C-1}{K}\frac{e^{\langle A[\bx_n]_i, \bx_n\rangle }}{\sum_{c=1}^C e^{\langle A[\bx_n]_c, x\rangle }}\langle A[\bx_n]_{y_n}-A[\bx_n]_i,\bx_n\rangle A[\bx_n]_{y_n}
    \nonumber\\
    &=\frac{C-1}{K}\frac{e^{\langle A[\bx_n]_i, x\rangle }}{\sum_{c=1}^C e^{\langle A[\bx_n]_c, \bx_n\rangle }}\left(-K\bx_n+\langle A[\bx_n]_{y_n}-A[\bx_n]_i,x\rangle A[\bx_n]_{y_n}\right)
    \nonumber\\
    &=\frac{C-1}{K}\frac{e^{\langle A[\bx_n]_i, x\rangle }}{\sum_{c=1}^C e^{\langle A[\bx_n]_c, \bx_n\rangle }}\left(-K\bx_n+\langle \sqrt{\frac{C-1}{C}K}\bx_n+\sqrt{\frac{K}{C(C-1)}}\bx_n,\bx_n\rangle \sqrt{\frac{C-1}{C}K}\bx_n\right)
    \nonumber\\
     &=\frac{(C-1)}{K}\frac{e^{\langle A[\bx_n]_i, \bx_n\rangle }}{\sum_{c=1}^C e^{\langle A[\bx_n]_c, \bx_n\rangle }}\left(-K\bx_n+K||\bx_n||^2\bx_n\right)
     \nonumber\\
     &=0. 
 \end{align}
Now, we consider the other cases $k\not =y_n$:
 \begin{align}
    \frac{d l}{d A[\bx_n]_{i}}=&\frac{e^{\langle A[\bx_n]_i, \bx_n\rangle }}{\sum_{c=1}^C e^{\langle A[\bx_n]_c, x\rangle }}\bx_n+\frac{C-1}{K}\frac{e^{\langle A[\bx_n]_i, \bx_n\rangle }}{\sum_{c=1}^C e^{\langle A[\bx_n]_c, \bx_n\rangle }}\langle A[\bx_n]_{y_n}-A[\bx_n]_i,\bx_n\rangle A[\bx_n]_i
    \nonumber\\
    &=~\frac{e^{\langle A[\bx_n]_i, \bx_n\rangle }}{\sum_{c=1}^C e^{\langle A[\bx_n]_c, \bx_n\rangle }}\left(\bx_n+\frac{C-1}{K}\langle A[\bx_n]_{y_n}-A[\bx_n]_i,\bx_n\rangle A[\bx_n]_i\right)
    \nonumber\\
        &=\frac{e^{\langle A[\bx_n]_i, x\rangle }}{\sum_{c=1}^C e^{\langle A[\bx_n]_c, x\rangle }}\left(x-\frac{C-1}{K}\langle \sqrt{\frac{C-1}{C}K}\bx_n+\sqrt{\frac{K}{C(C-1)}}\bx_n,x\rangle \sqrt{\frac{K}{C(C-1)}}\bx_n\right)
        \nonumber\\
        &=~\frac{e^{\langle A[\bx_n]_i, x\rangle }}{\sum_{c=1}^C e^{\langle A[\bx_n]_c, x\rangle }}\left(\bx_n-||\bx_n||^2\bx_n\right)
        \nonumber\\
     &=~ 0. 
 \end{align}
This analysis is per-example, but it is clear that minimizing the per-example loss minimizes the global (dataset) loss.
\hfill $\Box$

 \subsection{Proofs of Theorems~\ref{thm:big2} and \ref{thm:big3}}

 The proofs follow easily by leveraging the definition of the nonlinearity coupled with the results from Section~\ref{sec:proof_rewritting} and \cite{boyd2004convex}.

\subsection{Proof of Theorem \ref{thm:universality}}
\label{sec:proof_universality}

The proof leverages a result from \cite{breiman1993hinging} that is similar but targeted at a functional $f$. Thanks to the independence of the $K$ max-affine splines being used internally to construct the two MASOs, the same proof can be used here applied independently on each of the output $[f(\bx)]_k$ as
 \begin{align}
     \| f(\bx)-f_\Theta(\bx)\|^2_2 =~ \sum_k \left( [f(\bx)]_k-[f_\Theta(\bx)]_k \right)^2 
     \leq~  \frac{\sum_k \eta_k}{R}
 \end{align}
since the same number of regions is present for each of the MASO's max-affine spline functions.
\hfill $\Box$

\subsection{Proof that Activation Optimization Leads to MASO DNs}
\label{sec:proof_unser}


\begin{proof}
\cite{unser2018representer} models a DN as
\begin{align}
    \underline{\bz^{(\ell)}}(\bx)=
    \left(\underline{f}_\sigma^{(\ell)}\circ \underline{f}_W^{(\ell)} \circ \dots \circ \underline{f}_\sigma^{(\ell)}\circ \underline{f}_W^{(\ell)}\right)(\bx),
\end{align}
with optimal nonlinearities for each operator and dimension defined as 
\begin{align}
    \underline{\sigma}^{(\ell)}_k(u)=b_{1,k,\ell}+b_{2,k,\ell}u+\sum_{r=1}^{\underline{R}} a_{r,k,\ell}\max(u-\tau_{r,k,\ell},0).
\end{align}
leading to a formulation in which $\underline{\bz}^{(\ell)}(\bx)=\underline{f}_\sigma^{(\ell)}(\underline{f}_W^{(\ell)}(\underline{\bz}^{(\ell-1)}(\bx)))$, $\underline{f}^{(\ell)}_W:\mathbb{R}^{\underline{D_\sigma^{(\ell-1)}}} \rightarrow \mathbb{R}^{\underline{D_W^{(\ell)}}}$ and $\underline{f}^{(\ell)}_\sigma:\mathbb{R}^{\underline{D_W^{(\ell-1)}}} \rightarrow \mathbb{R}^{\underline{D_\sigma^{(\ell)}}}$. 
Such a composition of operators can be cast into a composition of MASO operators. First, due to the linear combinations of nonlinearities in the nonlinear layer, we first have to split this nonlinear operator $\text{f}_\sigma^{(\ell)}$ into $f_{W_\sigma}^{(\ell)}\circ f_\sigma^{(\ell)}$ as it is clear from the above that after the nonlinearity with the $\max$ application, the linear combination is performed by a linear operator. 
With this definition and letting $\text{f}_W^{(\ell)}=f_W^{(\ell)}$ we obtain in accordance with our layer definition from Section \ref{def:layer}, the same DN from above but defined as 
\begin{align}
    f_\Theta(\bx)=
    \left(f_{W_\sigma}^{(L)}\circ f_\sigma^{(L-1)}\circ f_{W}^{(L-1)}\circ f_{W_\sigma}^{(L-1)} \circ \dots \circ f_\sigma^{(2)}\circ f_{W}^{(2)}\circ f_{W_\sigma}^{(2)}   \circ  f_\sigma^{(1)}\circ f_{W}^{(1)} \right)(\bx)
\end{align}
each of the above DN operators can now be cast as a MASO as follows:
\begin{itemize}
    \item $f_W^{(\ell)}=S[A_W^{(\ell)},B_W^{(\ell)}]$ with $D^{(\ell)}_W=\underline{D^{(\ell)}_W}$, $R=1$ and $[A^{(\ell)}_W]_{k,1,\bigcdot}=[W]_{k,\bigcdot}$ and $[B^{(\ell)}]_{k,1}=0,\forall k$,
    
    \item $f_\sigma^{(\ell)}=S[A_\sigma^{(\ell)},B_\sigma^{(\ell)}]$ with $D^{(\ell)}_\sigma=\underline{R}\underline{D^{(\ell)}_W}+\underline{D^{(\ell)}_W}$, $R=2$ and $[A^{(\ell)}_\sigma]_{k,1,\bigcdot}=\be_{k\%\underline{D^{(\ell)}_W}},[A^{(\ell)}_\sigma]_{k,2,\bigcdot}=0, \forall k$ and $[B^{(\ell)}]_{k,1}=\tau_{k\% \underline{D^{(\ell)}_W},k/\underline{D^{(\ell)}_W},\ell},[B^{(\ell)}]_{k,2}=0,k=1,\dots\underline{R}\underline{D^{(\ell)}_W}$ and $[B^{(\ell)}]_{k,1}=[B^{(\ell)}]_{k,2}=0,k=\underline{R}\underline{D^{(\ell)}_W},\dots,\underline{R}\underline{D^{(\ell)}_W}+\underline{D^{(\ell)}_W}$,
    
    \item $f_{W_\sigma}^{(\ell)}=S[A_{W_\sigma}^{(\ell)},B_{W_\sigma}^{(\ell)}]$ with $D^{(\ell)}_{W_\sigma}=\underline{D^{(\ell-1)}_\sigma}$, $R=1$ and $[A^{(\ell)}_{W_\sigma}]_{k,1,\bigcdot}=\sum_{r=1}^{\underline{R}}\be_{r+(k-1)\underline{R}}a_{r,k,\ell}+\be_{\underline{R}\underline{D^{(\ell)}}+k}$ and $[B^{(\ell)}_{W_\sigma}]_{k,1}=b_{1,k,\ell},\forall k$.
    
\end{itemize}
leading to the entire DN being a composition of MASO operators. 
To simplify notation, one could collapse together $f_\sigma^{(\ell)}\circ f_W^{(\ell)}\circ f_{W_\sigma}^{(\ell)}$ into a single MASO layer.
\end{proof}

\section{Recursive Neural Networks as a Composition of MASOs}
\label{sec:sm_RNN}

\begin{figure}
  \centering
(a)  
\includegraphics[width=2.5in]{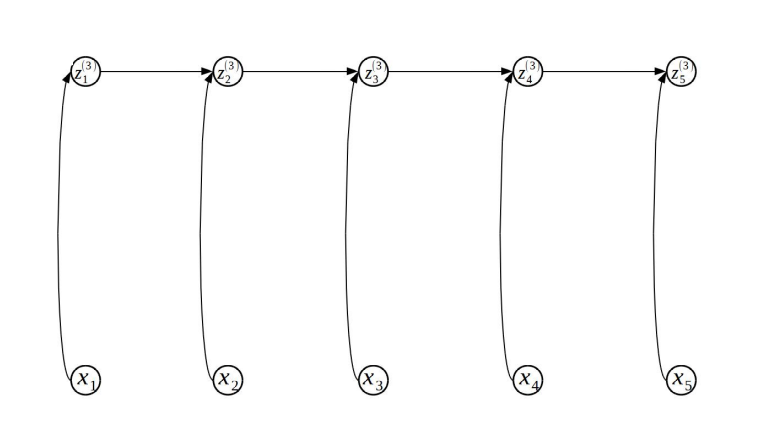}
(b) 
\includegraphics[width=2.50in]{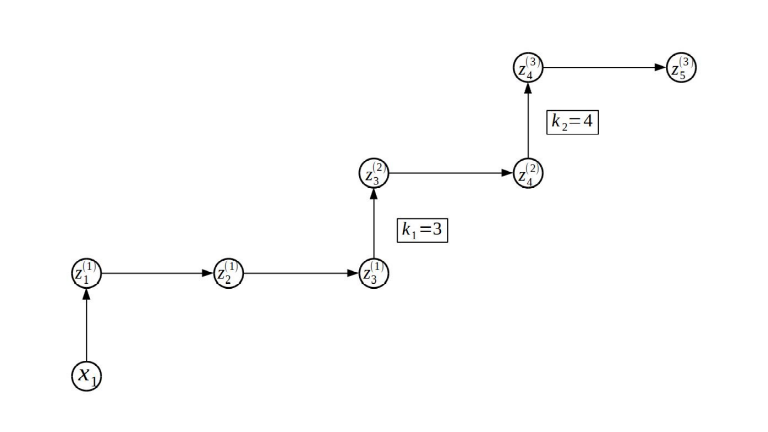}
  \caption{(a) At any given layer of a recursive neural network (RNN), there exists a direct input to a representation path and recursion. (b) In addition of all the possible input paths going through the hidden layers there are a combinatorially large number of paths, each one fully determined by the succession of forward in time or upward in layer successions.}
  \label{fig:blah1}
\end{figure}

We can derive one step of a standard fully recurrent neural network (RNN) \cite{graves2013generating} as
\begin{align}
    \bz_{\rm RNN}^{(1,t)} &=A^{(1,t)}_{\sigma}
    \left(W^{(1)}_{\rm  in}x^t+W^{(1)}_{\rm rec}\bz^{(1,t-1)}_{\rm RNN}+b^{(1)}
    \right)\left(W^{(1)}_{\rm in}x^t+W^{(1)}_{\rm rec}\bz^{(1,t-1)}_{\rm RNN}+b^{(1)}\right)+b_{\sigma}^{(1,t)},
    \\
    \bz_{\rm RNN}^{(\ell,t)} &=A^{(\ell,t)}_{\sigma}\left(W^{(\ell)}_{\rm in}x^t+W^{(\ell)}_{\rm rec}\bz^{(\ell,t-1)}_{\rm RNN}+W^{(\ell)}_{\rm up}\bz^{(\ell-1,t)}_{RNN}+b^{(\ell)}\right)
    \left(W^{(\ell)}_{\rm in}x^t+W^{(\ell)}_{\rm rec}\bz^{(\ell,t-1)}_{\rm RNN} \right.
    \nonumber\\
    &~~~~~~~ \left. +~W^{(\ell)}_{up}\bz^{(\ell-1,t)}_{RNN}+b^{(\ell)}\right)+b_{\sigma}^{(\ell,t)},\text{ $\ell>1$}.
\end{align}
By the double recursion of the formula (in time and in depth) we first proceed by writing the time unrolled RNN mapping as
\begin{align}
    \bz^{(1,T)}_{\rm RNN}(x) &= \sum_{t=T}^1\left(\prod_{k=T}^{t+1}A^{(1,k)}_{\sigma}W^{(1)}_{\rm rec}\right)A^{(1,t)}_{\sigma}
    \left(W^{(1)}_{\rm in}x^t+ b_{\sigma}^{(1,t)}+A^{(1,t)}_{\sigma} b^{(1)}\right) 
    \nonumber\\
    &= \sum_{t=T}^1\left(\prod_{k=T}^{t+1}A^{(1,k)}_{\sigma}W^{(1)}_{\rm rec}\right)A^{(1,t)}_{\sigma}W^{(1)}_{\rm in}x_t+\sum_{t=T}^1\left( \prod_{k=T}^{t+1}A^{(1,k)}_{\sigma} W_{\rm rec}^{(1)} \right)\left( b_{\sigma}^{(1,t)}+A^{(1,t)}_{\sigma} b^{(1)} \right)
    \\
    \bz^{(\ell,T)}_{\rm RNN}(x) 
    &=  \sum_{t=T}^1\left(\prod_{k=T}^{t+1}A^{(\ell,k)}_{\sigma}W^{(\ell)}_{\rm rec}\right)A^{(\ell,t)}_{\sigma}W^{(\ell)}_{\rm in}x^t
    \nonumber\\
    &~~~~~~~ +\sum_{t=T}^1\left( \prod_{k=T}^{t+1}A^{(\ell,k)}_{\sigma} W_{\rm rec}^{(\ell)} \right)\left( b_{\sigma}^{(\ell,t)}+A^{(\ell,t)}_{\sigma} b^{(\ell)}+ A^{(\ell,t)}_{\sigma}W^{(\ell)}_{\rm up}\bz^{(\ell-1,t)}_{\rm RNN}(x)\right),\text{ $\ell >1$.}
\end{align}
The formula unrolled in time is still recursive in depth. 
While the exact unrolled version would be cumbersome to compute for any layer $\ell$, we propose a simple way to find the analytical formula based on the possible paths an input can take till the final time representation of layer $\ell$. To do so, see Figure \ref{fig:blah1}. 

Hence we can decompose all of the paths by blocks of forward interleave with upward paths.
With this, we can see that the possible paths are all the paths from the input to the final nodes,and they can not got back in time nor down in layers. Hence, they are all the possible combinations for forward in time or upward in depth. We can thus find the exact output formula below for RNN where we focus on the template matching part for clarity (thus omitting the bias terms from the formula):
\begin{align}
    \bz^{(2,T)}_{\rm RNN}(x) 
    =&  \sum_{t=1}^{T}\sum_{k_1=t}^{T}\left(\prod_{q=T}^{k_1+1}A^{(2,q)}_{\sigma}W^{(2)}_{\rm rec}\right)A^{(2,k_1)}_{\sigma}W^{(2)}_{\rm up}\left( \prod_{q=t+1}^{k_1-1}A^{(1,q)}_{\sigma} W_{\rm rec}^{(1)} \right)A^{(1,t)}_{\sigma}W^{(1)}_{\rm in}x^t\\
    \bz^{(3,T)}_{\rm RNN}(x) 
    =&  
    \sum_{t=1}^{T}\sum_{k_1=t}^{T}\sum_{k_2\geq k_1}^{T}\left(\prod_{q=T}^{k_2+1}A^{(3,q)}_{\sigma}W^{(3)}_{\rm rec}\right)A^{(3,k_2)}_{\sigma}W^{(3)}_{\rm up}\left(\prod_{q=k_1}^{k_2-1}A^{(2,q)}_{\sigma}W^{(2)}_{\rm rec}\right)A^{(2,k_1)}_{\sigma}W^{(2)}_{\rm up}
    \nonumber\\
    & ~~~
    \times\left( \prod_{q=t+1}^{k_1-1}A^{(1,q)}_{\sigma} W_{\rm rec}^{(1)} \right)A^{(1,t)}_{\sigma}W^{(1)}_{\rm in}x^t 
    \\
    \bz^{(4,T)}_{RNN}(x) 
    =&  
    \sum_{t=1}^{T}\sum_{k_1=t}^{T}\sum_{k_2\geq k_1}^{T}\sum_{k_3\geq k_2}^{T}\left(\prod_{q=T}^{k_3+1}A^{(4,q)}_{\sigma}W^{(4)}_{\rm rec}\right)A^{(4,k_3)}_{\sigma}W^{(4)}_{\rm up}\left(\prod_{q=T}^{k_2+1}A^{(3,q)}_{\sigma}W^{(3)}_{\rm rec}\right)A^{(3,k_2)}_{\sigma}W^{(3)}_{\rm up}
    \nonumber \\
    & ~~~ \times\left(\prod_{q=k_1}^{k_2-1}A^{(2,q)}_{\sigma}W^{(2)}_{\rm rec}\right)A^{(2,k_1)}_{\sigma}W^{(2)}_{\rm up}\left( \prod_{q=t+1}^{k_1-1}A^{(1,q)}_{\sigma} W_{\rm rec}^{(1)} \right)A^{(1,t)}_{\sigma}W^{(1)}_{\rm in}x^t\\
    \vdots&\nonumber
\end{align}

%% file: APPENDIX/topology.tex
\section{Deep Network Topologies and Datasets}
\label{ap:dntopology}

The {\em smallCNN} and {\em largeCNN} architectures used in the experiments are detailed in Figure \ref{fig:topo}, where, for example,
{\tt Conv2DLayer(layers[-1],192,3,pad='valid')} denotes a standard 2D convolution with $192$ filters of spatial size $(3,3)$ and with valid padding (i.e., no padding). 
ResNetD-W denotes the standard wide ResNet topology with depth $D$ and width $W$.

\begin{figure}
    \centering
    \includegraphics[width=0.4\linewidth]{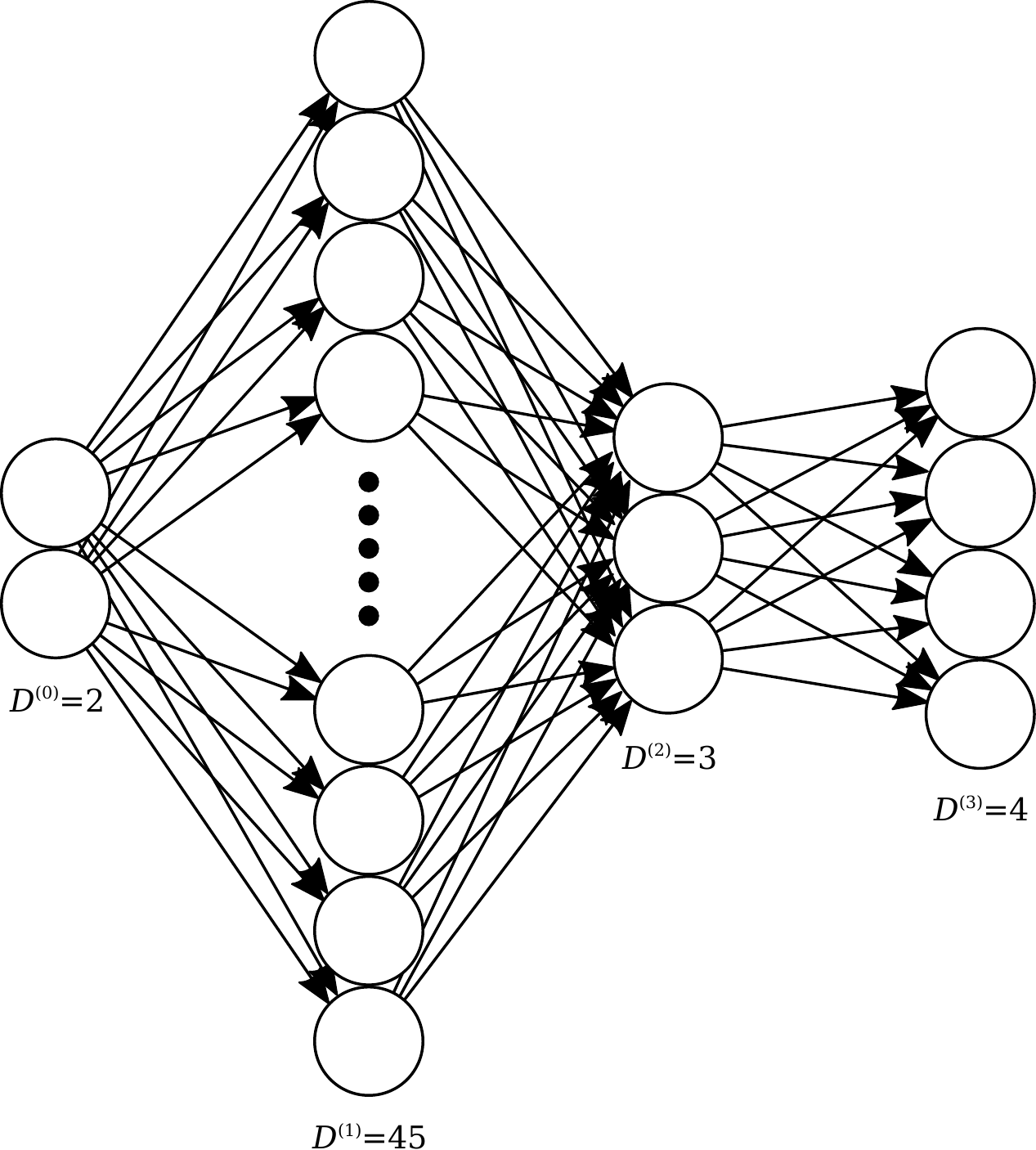}
    \caption{Block diagram of the toy DN used for the experiments in
Figure~\ref{fig:simplepartitions}.
}
\label{fig:toydn}
\end{figure}

Figure \ref{fig:toydn} contains the block diagram of the toy DN analyzed and visualized in Section~\ref{sec:simplestats}
(e.g., Figure~\ref{fig:simplepartitions}).

We used the standard datasets MNIST, CIFAR10, CIFAR100 and SVHN. 
We employed standard training regimes except for MNIST, where we drew the training and test set at random.

\begin{figure}[t]
\begin{minipage}{0.49\textwidth}
\begin{center}
    {\small\em smallCNN}
\end{center}
\footnotesize
\begin{verbatim}
Conv2DLayer(layers[-1],32,3,pad='valid')
Pool2DLayer(layers[-1],2)
Conv2DLayer(layers[-1],64,3,pad='valid')
Pool2DLayer(layers[-1],2)
Conv2DLayer(layers[-1],128,1,pad='valid')
Pool2DLayer(layers[-1],2)
\end{verbatim}
\end{minipage}
\begin{minipage}{0.49\textwidth}
\begin{center}
    {\small\em largeCNN}
\end{center}
\footnotesize
\begin{verbatim}
Conv2DLayer(layers[-1],96,3,pad='same')
Conv2DLayer(layers[-1],96,3,pad='full')
Conv2DLayer(layers[-1],96,3,pad='full')
Pool2DLayer(layers[-1],2)
Conv2DLayer(layers[-1],192,3,pad='valid')
Conv2DLayer(layers[-1],192,3,pad='full')
Conv2DLayer(layers[-1],192,3,pad='valid')
Pool2DLayer(layers[-1],2)
Conv2DLayer(layers[-1],192,3,pad='valid')
Conv2DLayer(layers[-1],192,1)
Pool2DLayer(layers[-1],2)
\end{verbatim}
\end{minipage}
\normalsize
\caption{Description of the {\em smallCNN} and {\em largeCNN} topologies.}
\label{fig:topo}
\end{figure}


\section{More on Orthogonal Templates} 
\label{ap:moreortho}

\begin{figure}[!ht]
\centering
{\em smallCNN} with SVHN \\
(a) \begin{minipage}{0.75\textwidth}
\centering
\includegraphics[width=1\textwidth]{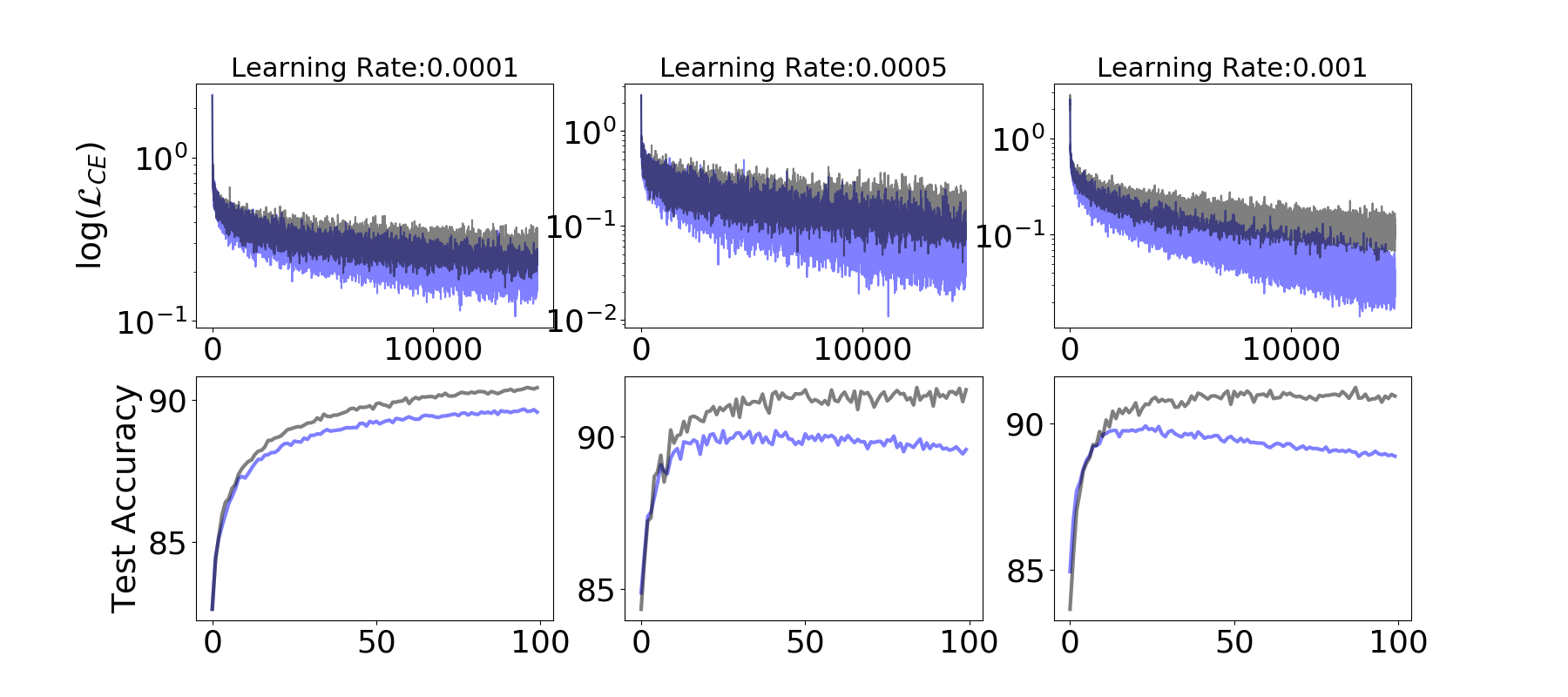}
\end{minipage}
\\
(b) \begin{minipage}{0.75\textwidth}
\centering
\includegraphics[width=1\textwidth]{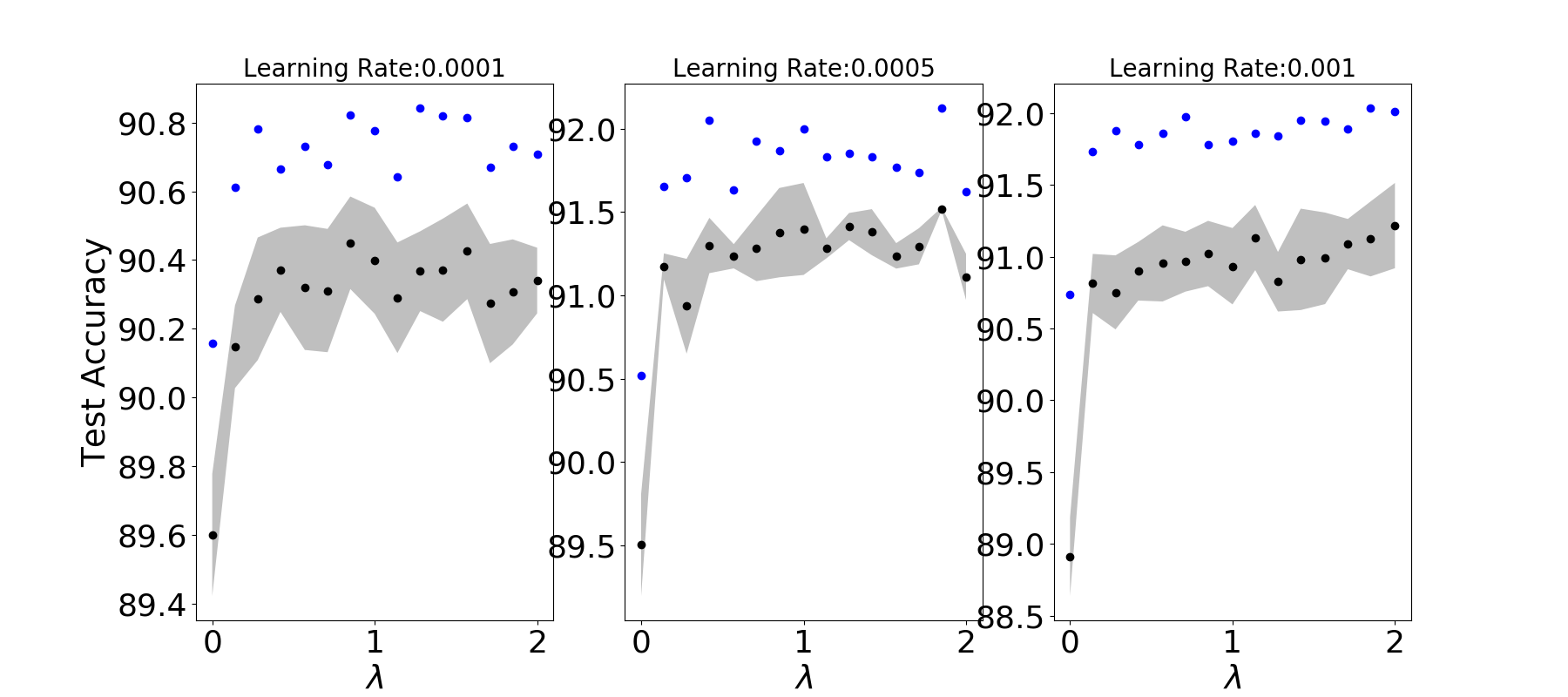}
\end{minipage}
\caption{Further experiments on orthogonal templates for {\em smallCNN} with the SVHN dataset.
(a) The presence of the orthogonal penalty loss shows greater accuracy as well as robustness to overfitting as opposed to standard DN setting. (b) Different values of the regularization parameter show the gain of having it greater than $0$ hence the impact of the proposed penalty term.}
\label{fig:extrahisto1}
\end{figure}

See Figure \ref{fig:extrahisto1} for additional results from our experiments with orthogonal template DNs. 

\section{Additional Partition Visualization and Statistics}
\label{sec:extratoy}

We provide additional experimental results for the toy DN studied in Section \ref{sec:simplestats} in Tables \ref{tab:null1}--\ref{tab:occupancy2}.

\begin{table}[t]
    \centering
    \small
    Number of Nonzero Volume Regions \\[2mm]
    \begin{tabular}{c|cccc} 
    Partition & \multicolumn{2}{c}{Initialization} & \multicolumn{2}{c}{Trained} \\
     & (\st{BN}) & (BN) & (\st{BN}) & (BN) \\ \hline
    $\omega_{\rm in}^{(1)}$    
    & 90 & 90 & 681 & 801 \\ \hline
    $\omega_{\rm in}^{(2)}$       
    & 6 & 8 & 8 & 8\\ \hline
    $\omega_{\rm g}^{(2)}$ &98&98&800&1068
    \end{tabular}
    \caption{
    Number of nontrivial partition regions having nonzero volume for the input signal space partition induced by 
    $\omega_{\rm in}^{(1)}$ from layer 1 and 
    $\omega_{\rm in}^{(2)}$ from  2 of the toy DN from Figure~\ref{fig:simplepartitions} for leaky RELU.}
    \label{tab:null1}
    \end{table}
    
    \begin{table}[t]
    \centering
    \small
    Number of Nonempty Regions \\[2mm]
    \begin{tabular}{c|cccc} 
     & \multicolumn{2}{c}{Initialization} & \multicolumn{2}{c}{Trained} \\
    Partition & (\st{BN}) & (BN) & (\st{BN}) & (BN) \\ \hline
    $\omega_{\rm in}^{(1)}$    
    & 90 & 90 & 288 & 332 \\ \hline
    $\omega_{\rm in}^{(2)}$
    & 6 & 8 & 6 & 8 \\ \hline
    $\omega_{\rm g}^{(2)}$ &98&98&311&370
    \end{tabular}
    \caption{
        Number of partition regions that are occupied by at least one training data point for the toy DN from Figure~\ref{fig:simplepartitions} for leaky RELU. 
    }
    \label{tab:occupancy1}
\end{table}

\begin{table}[t]
    \centering
    \small
    Number of Nonzero Volume Regions \\[2mm]
    \begin{tabular}{c|cccc} 
    Partition & \multicolumn{2}{c}{Initialization} & \multicolumn{2}{c}{Trained} \\
     & (\st{BN}) & (BN) & (\st{BN}) & (BN) \\ \hline
    $\omega_{\rm in}^{(1)}$    
    & 90 & 90 & 777 & 831 \\ \hline
    $\omega_{\rm in}^{(2)}$       
    & 3 & 8 & 8 & 8\\ \hline
    $\omega_{\rm g}^{(2)}$ &102&106&950&1115
    \end{tabular}
    \caption{
    Number of nontrivial partition regions having nonzero volume for the input signal space partition induced by 
    $\omega_{\rm in}^{(1)}$ from layer 1 and 
    $\omega_{\rm in}^{(2)}$ from  2 of the toy DN from Figure~\ref{fig:simplepartitions} for the absolute value activation function.}
    \label{tab:null2}
    \end{table}
    
    \begin{table}[t]
    \centering
    \small
    \vspace*{2mm}
    Number of Nonempty Regions \\[2mm]
    \begin{tabular}{c|cccc} 
     & \multicolumn{2}{c}{Initialization} & \multicolumn{2}{c}{Trained} \\
    Partition & (\st{BN}) & (BN) & (\st{BN}) & (BN) \\ \hline
    $\omega_{\rm in}^{(1)}$    
    & 90 & 90 & 294 & 341 \\ \hline
    $\omega_{\rm in}^{(2)}$
    & 3 & 8 & 8 & 8 \\ \hline
    $\omega_{\rm g}^{(2)}$ &102&106&376&445
    \end{tabular}
    \caption{
        Number of partition regions that are occupied by at least one training data point for the toy DN from Figure~\ref{fig:simplepartitions} for the absolute value activation function. 
    }
    \label{tab:occupancy2}
\end{table}